\pdfoutput=1

\documentclass[11pt]{article}

\usepackage{ACL2023}

\usepackage{times}
\usepackage{latexsym}
\usepackage{graphicx}
\usepackage{subfig}
\usepackage{verbatim}

\usepackage[T1]{fontenc}

\usepackage[utf8]{inputenc}

\usepackage{microtype}
\usepackage{amsmath}
\usepackage{amssymb}
\usepackage{amsfonts}
\usepackage{algorithm}
\usepackage{algpseudocode}

\usepackage{inconsolata}

\title{Rethinking the Role of Scale for In-Context Learning: An Interpretability-based Case Study at 66 Billion Scale}

\author{Hritik Bansal$^{1}$\thanks{~ Work done as an intern at AWS AI Labs.} \quad Karthik Gopalakrishnan$^{2}$\thanks{~ Corresponding author.\newline Code: \href{https://github.com/amazon-science/llm-interpret}{github.com/amazon-science/llm-interpret}} \quad Saket Dingliwal$^{2}$ \quad Sravan Bodapati$^{2}$\\
{\bf Katrin Kirchhoff$^{2}$ \quad Dan Roth$^{2}$}\\
$^1$University of California, Los Angeles \;\;\;\;\; $^2$AWS AI Labs\\
{\tt\small hbansal@cs.ucla.edu}\\
{\tt\small \{karthgop, skdin, sravanb, katrinki, drot\}@amazon.com}}

\begin{document}
\maketitle

\begin{abstract}
    Language models have been shown to perform better with an increase in scale on a wide variety of tasks via the in-context learning paradigm. In this paper, we investigate the hypothesis that the ability of a large language model to in-context learn-perform a task is not uniformly spread across all of its underlying components. Using a 66 billion parameter language model (OPT-66B) across a diverse set of 14 downstream tasks, we find this is indeed the case: $\sim$70\% of attention heads and $\sim$20\% of feed forward networks can be removed with minimal decline in task performance. We find substantial overlap in the set of attention heads (un)important for in-context learning across tasks and number of in-context examples. We also address our hypothesis through a task-agnostic lens, finding that a small set of attention heads in OPT-66B score highly on their ability to perform primitive \textit{induction} operations associated with in-context learning, namely, prefix matching and copying. These \textit{induction} heads overlap with task-specific important heads, reinforcing arguments by \citet{olsson2022context} regarding induction head generality to more sophisticated behaviors associated with in-context learning. Overall, our study provides several insights that indicate large language models may be under-trained for in-context learning and opens up questions on how to pre-train language models to more effectively perform in-context learning.
\end{abstract}

\section{Introduction}

\begin{figure}[h]
    \centering
    \includegraphics[scale=0.26]{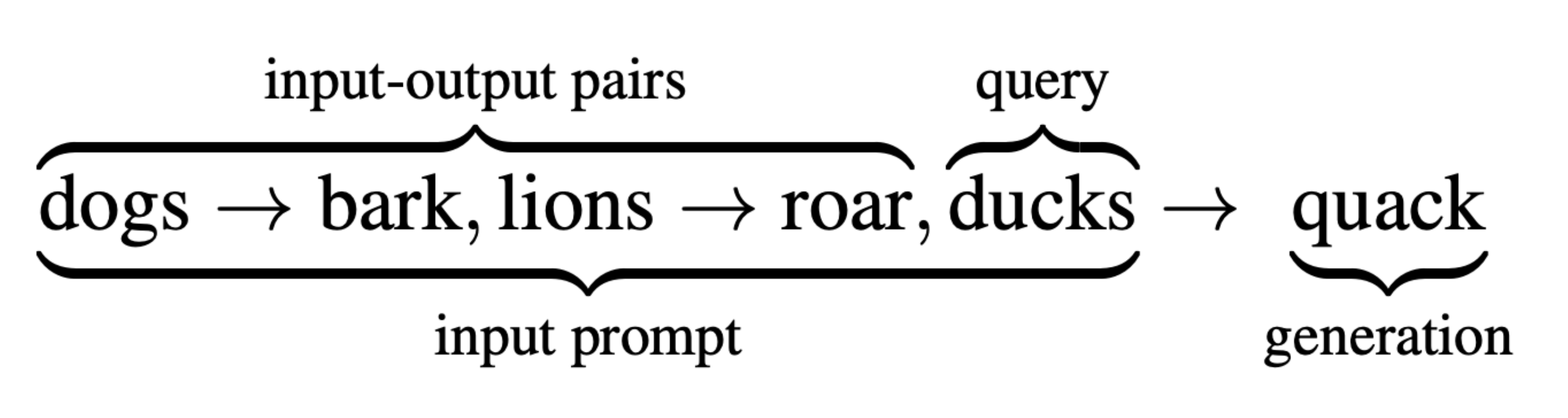}
    \caption{A sample input prompt for in-context learning and the model output.}
    \label{fig:example}
\end{figure}

In recent years, large language models (LLMs) \cite{brown2020language,rae2021scaling,lieber2021jurassic,black2022gpt,zhang2022opt,chowdhery2022palm,hoffmann2022training,smith2022using} based on the Transformer architecture \cite{vaswani2017attention} pre-trained using self-supervision on web-scale textual corpora have revolutionized the field of natural language processing (NLP). At larger scales, these models demonstrate remarkable \textit{emergent} \cite{wei2022emergent} prowess in performing a wide variety of tasks without any form of fine-tuning, via the zero/few-shot in-context learning paradigm \cite{brown2020language}. In this paradigm (depicted in Figure \ref{fig:example}), LLMs are prompted to generate output text conditioned on a few (or zero) in-context examples that form solved "input-output" pairs along with a query input.

How in-context learning works has been an open question since its advent and recent studies \cite{xie2021explanation,garg2022can,olsson2022context, min2022rethinking} have begun scratching the surface toward better understanding the paradigm. In this paper, we empirically address the following key question:
\begin{quote}
\textit{Are all LLM components really needed to perform in-context learning?}
\end{quote}

The first way we address the aforementioned question is through the lens of task-specific importance scores and structured pruning \cite{li2016pruning,molchanov2016pruning,anwar2017structured} of components underlying modern LLMs, which are primarily stacks composed of multiple high-dimensional self-attention blocks that form \textit{multi-headed attention} and densely activated \textit{feed forward networks} (FFNs). We pick the Open Pre-trained Transformer (OPT) \cite{zhang2022opt} model with 66B parameters for our analyses, which yield several surprising observations. We find that important attention heads are primarily clustered in the intermediate layers and important FFNs are primarily in later (31+) layers (\S \ref{exp:individual_components}). We find that the ability to perform zero/few-shot in-context learning on almost all of a variety of 14 NLP datasets/tasks stays nearly intact when up to 70\% ($\sim$15.7B parameters in OPT-66B) of the attention heads are removed (\S \ref{exp:iterative attention heads}). The attention heads that are (un)important for in-context learning also seem to overlap across tasks (\S \ref{exp:cross_task}) and shots (\S \ref{exp:cross_shot}), and pruning of attention heads based on a "universal" importance order computed using all 14 datasets generalizes to varying degrees on out-of-distribution datasets (\S \ref{exp:generalization_trends}). These observations indicate that a common task-agnostic subset of the attention heads are responsible for in-context learning.
We also find that only up to 20\% of the FFNs ($\sim$8.5B parameters) can be removed with minimal decline in zero/few-shot in-context learning performance (\S \ref{exp:iterative fc subnetworks}), indicating the importance of FFNs toward in-context learning. In addition, when accounting for the interplay between attention heads and FFNs by jointly pruning both based on their individual importance orders, the newly obtained inflection points to in-context learning performance do not deviate much from the inflection points obtained via standalone pruning (\S \ref{exp:combined iterative pruning}).

The second way we address the aforementioned question is by quantifying the capacity of all attention heads in OPT-66B to perform a subset of task-agnostic primitive operations associated with in-context learning, namely, \textit{prefix matching} and \textit{copying}: explicitly searching for a prior occurrence of the current token in-context and copying over its suffix. \citet{elhage2021mathematical} and \citet{olsson2022context} developed a mathematical framework to reverse-engineer a Transformer and also find such heads, termed \textit{induction heads}, and explored the hypothesis that such heads drive in-context learning with model sizes up to 13B parameters in a mostly task-agnostic fashion. Using this framework, we compute task-agnostic scores for prefix matching and copying for each attention head and find that a small set of heads in OPT-66B have non-trivial scores for both primitives (\S \ref{exp:ind_heads}). Qualitative inspection and quantitative analyses show that these heads overlap (to varying degrees) with the ones identified earlier to be important for in-context learning via our set of 14 NLP datasets/tasks, suggesting that induction heads are capable of more sophisticated behaviors associated with in-context learning such as latent concept matching but are not the only heads with such capabilities (\S \ref{exp:prune_prefixmatch_copy}).

Overall, our study provides several insights about in-context learning at massive scale using both task-specific and task-agnostic settings. In a world of ever increasing language model sizes, we believe these insights serve as a strong foundation for researchers and practitioners in language modeling to build and leverage compact language models that can also demonstrate emergent abilities.

\section{Background \& Methods}
\label{background}

In this section, we briefly describe the Open Pre-trained Transformer (OPT) \cite{zhang2022opt} model used for our study, provide background on in-context learning and the mathematical formulation of induction heads by \citet{olsson2022context} that we build on, and describe our adaptation of oracle and gradient-based importance score formulations for in-context learning.

\subsection{Open Pre-trained Transformer (OPT)}
\label{opt architecture}

OPT is a suite of language models of varying sizes that are aimed at serving as open replicas of GPT-3. The largest openly accessible model from this suite is OPT-66B with 66 billion parameters, which was also the largest publicly available dense decoder-only language model at the time of our experiments. Hence, we chose OPT-66B for our study.

\textbf{Architecture}: Consider a tokenized input sentence to OPT,  $\mathbf{X} \in \mathbb{R}^{N \times d_e}$, where $N$ is the number of tokens in the sentence and $d_e$ is the embedding dimension. The input is processed by multiple decoder layers consisting of \textit{multi-headed attention} (MHA) blocks, \textit{layer norm} (LN) and \textit{feed forward networks} (FFN), followed by a final FFN to produce logits over the output vocabulary. The decoder layers can be formally expressed as follows:
\begin{align*}
    \mathbf{t}^{(\ell+1)} &= \mathbf{z}^\ell + \text{MHA}^{\ell}(\text{LN}^{\ell}(\mathbf{z}^\ell))\tag{1}\label{eq:1}\\
    \mathbf{z}^{(\ell+1)} &= \mathbf{t}^{(\ell+1)} + \text{FFN}^{\ell}(\mathbf{t}^{(\ell+1)})\tag{2}\label{eq:2}
\end{align*}

\noindent where $\mathbf{z}^{1} = \mathbf{X}$, and \eqref{eq:1} \& \eqref{eq:2} are the residual connections corresponding to the MHA and FFN in layer $\ell >= 1$ respectively. OPT-66B was pre-trained with a maximum sequence length of 2048 and embedding dimension $d_e = 9216$.\\

\textbf{MHA}: In an MHA block, $H$ attention heads are applied in parallel to the input and their outputs are concatenated. In OPT-66B, there are $H = 72$ attention heads of dimension $d_h = 128$ in every layer $\ell$. An individual attention head $h$ in layer $\ell$ consists of three learnable matrices, $\mathbf{W}_k^{h}, \mathbf{W}_q^{h}, \mathbf{W}_v^{h} \in \mathbb{R}^{d_e \times d_h}$, all unique to the head, such that it applies self-attention $A^{h}(.)$ on the input, where $d_h = d_e/H$. Formally, for input $\mathbf{M}$ in layer $\ell$:
\begin{align*}
    \text{MHA}^\ell(\mathbf{M}) &= [A^{1}(\mathbf{M}); \cdots; A^{H}(\mathbf{M})]\mathbf{W}_o^{\ell}\tag{3}\label{eq:3}\\
    A^{h}(\mathbf{M}) &= s^{h}(\mathbf{M})\mathbf{M}\mathbf{W}_v^{h}\tag{4}\label{eq:4}\\
    s^{h}(\mathbf{M}) &= \sigma\Big(\frac{\mathbf{M}\mathbf{W}_q^{h}(\mathbf{W}_k^{h})^T\mathbf{M}^T}{\sqrt{d_h}}\Big)\tag{5}\label{eq:5}
\end{align*}

\noindent where $\sigma$ is the softmax function and $\mathbf{W}_o^{\ell} \in \mathbb{R}^{d_e \times d_e}$ is a learnable output matrix unique to the MHA block in layer $\ell$. To ensure OPT is auto-regressive, the output of $s^{h}(.)$ is masked to prevent the dependence of the hidden state of the token $i$, $z^\ell_i \in \mathbb{R}^{d_e}$, on the tokens that lie to the right of it in indices $\{i+1,\dots,N\}$.

To remove a head $h$ in layer $\ell$ in practice, we set $A^{h}(\mathbf{M})$ to be the zero matrix in Equation \eqref{eq:3}. This implies that $\mathbf{W}_k^{h}, \mathbf{W}_q^{h}, \mathbf{W}_v^{h}$ can be entirely removed, and the corresponding $d_h$ rows in $\mathbf{W}_o^{\ell}$ can also be removed. In total, there are $4608$ attention heads across $64$ layers in OPT-66B that constitute 21.7B of the total 66B parameters.\\

\textbf{FFN}: Each layer $\ell$ consists of a feed forward network (FFN) parameterized by a high-dimensional projection matrix, $\mathbf{W}^\ell_1 \in \mathbb{R}^{d_e \times d}$ followed by a low-dimensional projection matrix, $\mathbf{W}^\ell_2 \in \mathbb{R}^{d \times d_e}$ where $d = 36864$ for OPT-66B. Formally, for input $\mathbf{M}$ in layer $\ell$:
\begin{align*}
    \text{FFN}^\ell(\mathbf{M}) = \text{ReLU}(\text{LN}^\ell(\mathbf{M})\mathbf{W}^\ell_1)\mathbf{W}^\ell_2\tag{6}\label{eq:6}
\end{align*}

\noindent where ReLU is the rectified linear unit activation function and LN is the layer norm.

To remove an FFN in layer $\ell$ in practice, we set FFN$^{\ell}(\mathbf{M})$ to be the zero matrix in Equation \eqref{eq:6}. This implies $\mathbf{W}^\ell_1$, $\mathbf{W}^\ell_2$ and the layer norm LN$^{\ell}(.)$ for the FFN can be entirely removed. In total, FFNs constitute $43.4$B parameters in OPT-66B.

\subsection{In-Context Learning \& Induction Heads}
\label{bg:in-context}

With increasingly larger language models being trained in recent years, a new paradigm of learning termed \textit{in-context learning} \cite{brown2020language} has become popular. In this paradigm, language models perform tasks by being prompted to generate output text conditioned on a few (or zero) in-context training examples that form solved "input-output" pairs for the task along with a query input. There are no gradient-based model updates involved as this is a purely inference-time paradigm. Figure \ref{fig:example} illustrates the paradigm for the task of identifying the sound that an animal makes. In some cases, tasks can also be accompanied by task descriptions/templates to help prime the language model better, e.g., zero-shot translating from English to German using the prompt:
\begin{quote}
\textit{English phrase: I like dogs.\\ German phrase: }
\end{quote}

\begin{figure}[h]
    \centering
    \includegraphics[scale=0.25]{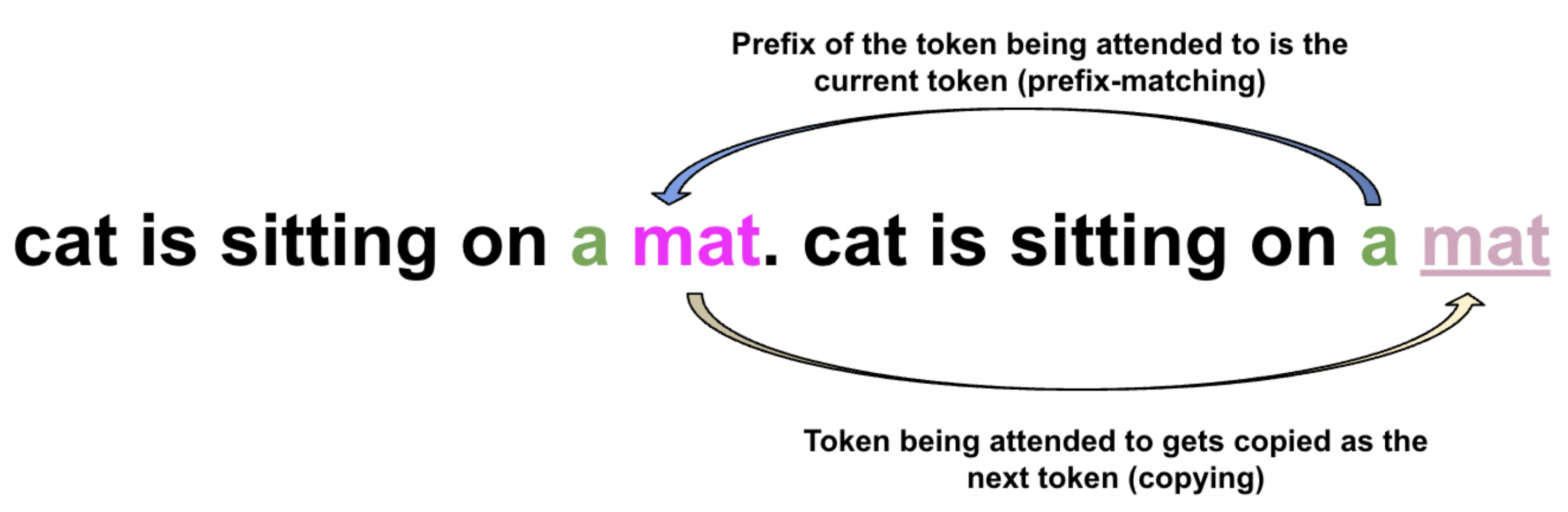}
    \caption{Prefix matching and copying depicted at a given time-step for a repeated sequence of tokens.}
    \label{fig:copying_prefix_matching}
\end{figure}

While these examples involve learning and relying on latent concepts during inference, few-shot in-context learning can additionally involve explicit primitive interactions between the in-context examples. For example, with the prompt:
\begin{quote}
\textit{English phrase: I like dogs.\\ German phrase: ich mag Hunde.\\ English phrase: I like ducks.\\ German phrase: }
\end{quote}

\noindent the model may rely on prior in-context translations of the tokens \textit{I} and \textit{like} when performing the task for the query input. \citet{olsson2022context} develop a mathematical framework toward better understanding such mechanics, starting off with a task-agnostic formulation of in-context learning as the ability of a model to better predict tokens later in the context than the tokens earlier \cite{kaplan2020scaling} and defining a heuristic for it. They define a set of task-agnostic primitive operations that reflect the kind of interactions we refer to in the above example, namely, \textit{prefix matching} and \textit{copying}. These operations are defined in a simplistic fashion on a repeated sequence of randomly generated tokens: explicitly searching for a prior occurrence of the current token in-context and copying over its suffix. The heads that are capable of performing these operations are termed \textit{induction heads}. Figure \ref{fig:copying_prefix_matching} depicts these operations for a repeated sequence of tokens. While these operations are intertwined in practice, the capacity of attention heads to \textit{independently} perform them is computed with the scoring algorithms described in detail in Appendix \ref{appen:pm_cs}.

\subsection{Importance Scores}

Consider a model $\mathcal{M}$ and a dataset $\mathcal{D}=\{\mathcal{X},\mathcal{Y}\}$, where $\mathcal{X}=\{\mathbf{x}_1, \cdots,\mathbf{x}_L\}$ and $\mathcal{Y}=\{\mathbf{y}_1, \cdots, \mathbf{y}_L\}$ such that $\mathbf{x}_i$ represents a prompt with few (or zero) in-context training examples along with a query input and $\mathbf{y}_i$ represents the corresponding target output sequence. We define and compute importance scores for model components using such datasets to quantify their relative contributions to the model's ability to perform in-context learning.

\subsubsection{Oracle}
\label{bg:oracle_pruning}
Let $\mathcal{P}_{\mathcal{M}}(\mathcal{D})$ denote a dataset/task-specific performance metric, e.g., accuracy. Given dataset $\mathcal{D}$, the oracle importance score of a component $\mathcal{C}$ in $\mathcal{M}$ is computed as follows:
\begin{equation}
    IS_{\mathcal{C}}(\mathcal{D}) = \mathcal{P}_{\mathcal{M}}(\mathcal{D}) - \mathcal{P}_{\mathcal{M}_{\backslash \mathcal{C}}}(\mathcal{D})\tag{7}\label{eq:7}
\end{equation}

\noindent where $\mathcal{M}_{\backslash \mathcal{C}}$ denotes the resultant model when $\mathcal{C}$ is pruned from $\mathcal{M}$. Clearly, if pruning a component leads to poor model performance on the task, it must be important for the task. Similarly, if there is no difference or an improvement in performance upon pruning a component, it must be unimportant.

Computing oracle importance scores for $K$ model components requires us to perform $\mathcal{O}(K)$ evaluations for each dataset $\mathcal{D}$. Given that OPT-66B only has 64 FFNs, this is computationally feasible. However, it becomes infeasible to score the 4608 attention heads in OPT-66B in this manner, so we adopt the second method described below to compute importance scores for attention heads.

\subsubsection{Gradient-based}
\label{bg:head_importance}
Given dataset $\mathcal{D}$, the gradient-based importance score \cite{molchanov2016pruning,michel2019sixteen} of an attention head $h$ captures the expected sensitivity of the model to $h$ and is computed as follows:

\begin{equation*}
    IS_{h}(\mathcal{D}) = \mathbb{E}_{(\mathbf{x},\mathbf{y})}\left|A^{h}([\mathbf{x};\mathbf{y}])^{T}\frac{\partial \mathcal{L}(\mathbf{y}|\mathbf{x})}{\partial A^{h}([\mathbf{x};\mathbf{y}])}\right|\tag{8}\label{eq:8}
\end{equation*}%

\noindent where ; is the concatenation operator, $(\mathbf{x},\mathbf{y}) \sim \mathcal{D}$ such that $\mathbf{x}$ is a sequence of $T_x$ tokens $x_{1:T_x}$, $\mathbf{y}$ is a sequence of $T_y$ tokens $y_{1:T_y}$, $A^{h}$ is defined in \eqref{eq:4} and the loss term in \eqref{eq:8} is computed using the auto-regressive decomposition of the log-likelihood:
\begin{align*}
\mathcal{L}(\mathbf{y}|\mathbf{x}) = -\frac{1}{T_y}\sum_{j=1}^{j=T_y}\text{log}(p(y_{j}|\mathbf{x}, y_{1:j-1}))\tag{9}\label{eq:9}
\end{align*}

These importance scores can be efficiently computed for all heads by simply performing a single forward and backward pass over the model with $\mathcal{D}$.

We also define the aggregated importance score of an attention head on a set of datasets $\mathbb{S}=\{\mathcal{D}_1,\cdots,\mathcal{D}_K\}$ as follows:

\begin{equation}
    IS_{h}(\mathbb{S}) = \mathbb{E}_{\mathcal{D}\sim \mathbb{S}}\left[IS_{h}(\mathcal{D})\right]\tag{10}\label{eq:10}
\end{equation}

\section{Experimental Setup}

We perform our experiments with OPT-66B on a variety of 14 NLP datasets/tasks. For consistency in the evaluation metric, we report accuracy on all tasks. Our choice of datasets and metric is in line with that of Meta AI in the OPT paper \cite{zhang2022opt}. The datasets include ARC Easy and Challenge \cite{clark2018think} and OpenBookQA \cite{mihaylov2018can} for advanced question-answering, HellaSwag \cite{zellers2019hellaswag}, PIQA \cite{bisk2020piqa} and Winogrande \cite{sakaguchi2021winogrande} for various forms of commonsense reasoning, and the following datasets from the standard SuperGLUE benchmark \cite{wang2019superglue}: BoolQ, CB, COPA, MultiRC, ReCoRD, RTE, WiC, and WSC. For a subset of experiments involving evaluation for out-of-distribution generalization, we also use 2 additional datasets: MathQA \cite{amini2019mathqa} and LAMBADA \cite{paperno2016lambada}.

We use a modified version of Eleuther AI's \textit{lm-evaluation-harness} framework \cite{eval-harness} for our experiments. The default framework samples in-context examples at random, which we use as-is without modification.

\section{Importance Scores for OPT-66B}
\label{exp:individual_components}

In this section, we present importance scores for all attention heads and feed forward networks in OPT-66B toward performing zero-shot and few-shot (1-shot and 5-shot) in-context learning.

\subsection{Attention Heads}
\label{exp:attention_importance}

\begin{figure*}[h]
    \centering
    \subfloat[\centering\label{exp_fig:heatmap_a} Zero-shot]{{\includegraphics[width=0.36\linewidth]{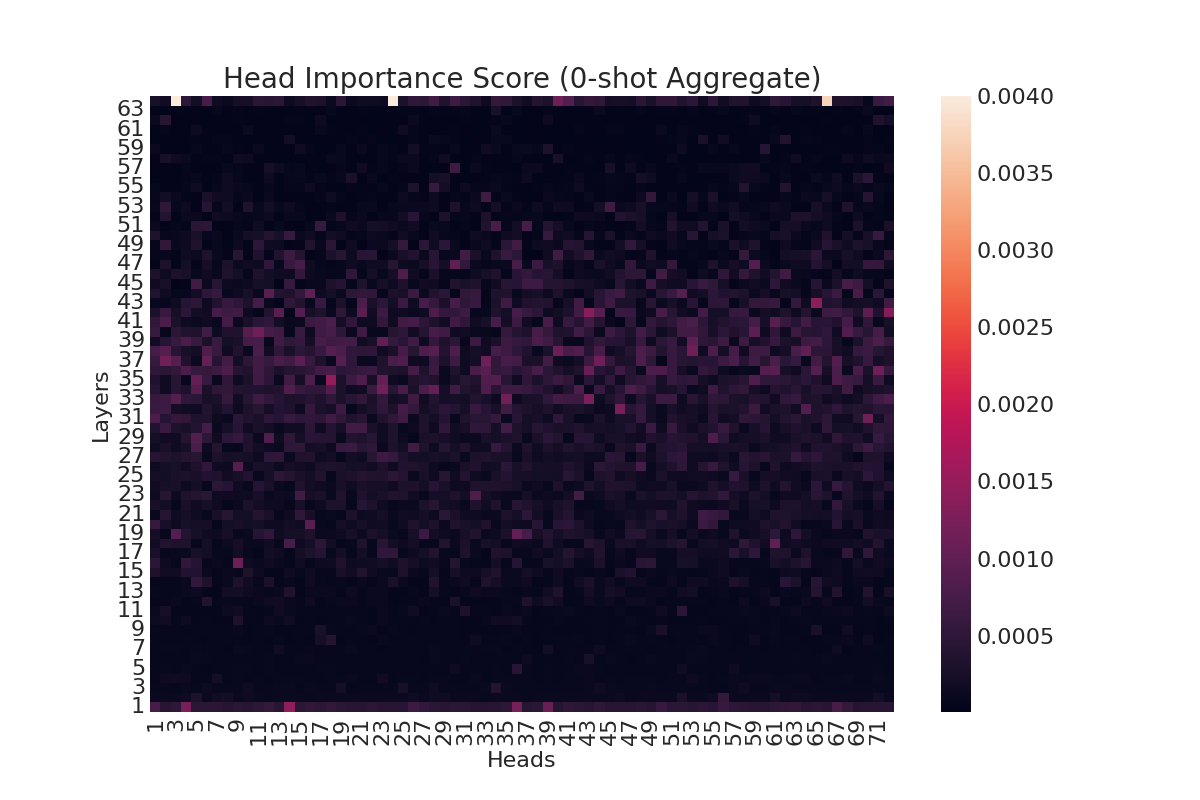}}}
\subfloat[\centering\label{exp_fig:heatmap_b} One-shot]{{\includegraphics[width=0.36\linewidth]{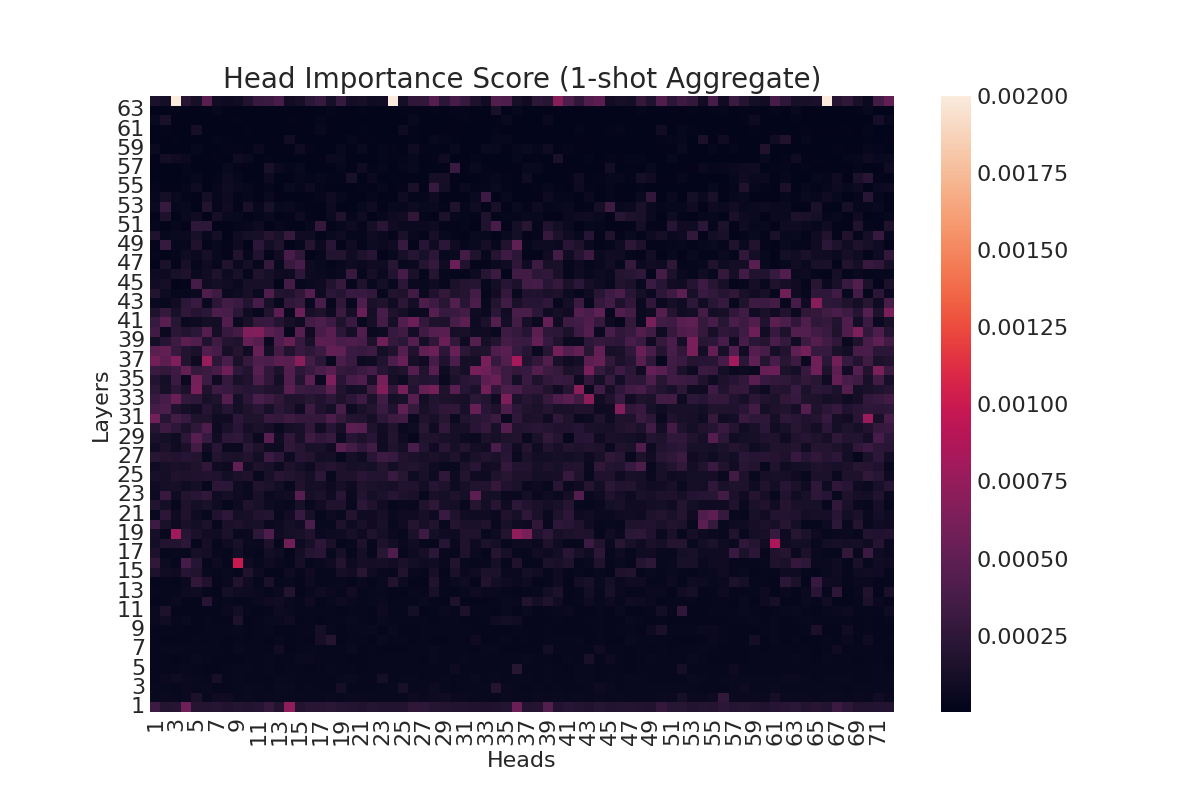}}}
\subfloat[\centering\label{exp_fig:heatmap_c} Five-shot]{{\includegraphics[width=0.36\linewidth]{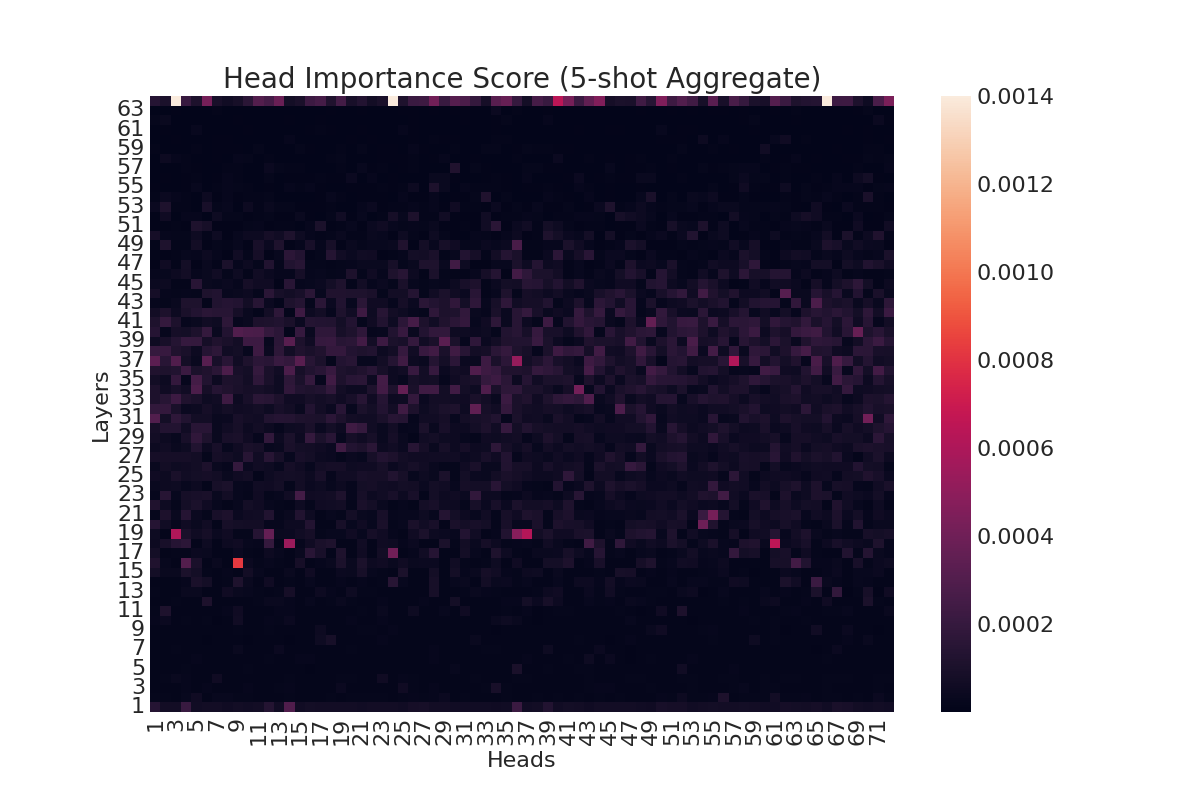}}}
    \caption{Attention head aggregate importance score heatmaps for in-context learning with OPT-66B.}
    \label{exp_fig:heatmap}
\end{figure*}

Figure \ref{exp_fig:heatmap} depicts heatmaps of the head importance scores averaged across all tasks (as described in \S \ref{bg:head_importance}) in the zero-shot, one-shot and five-shot settings. Task-specific heatmaps are provided in Appendix \ref{app:head_imp_pertask}. We observe that the important attention heads are primarily clustered in the intermediate layers of OPT-66B in both the task-averaged and task-specific cases. We also observe some overlap in the most important attention heads across the different zero/few-shot settings. This is further confirmed in follow-up analysis in \S \ref{exp:cross_shot}.

\subsection{Feed Forward Networks}
\label{exp:fc_score}

\begin{figure*}[h]
    \centering
    \subfloat[\centering\label{exp_fig:fc_imp_a} Zero-shot]{{\includegraphics[width=0.36\linewidth]{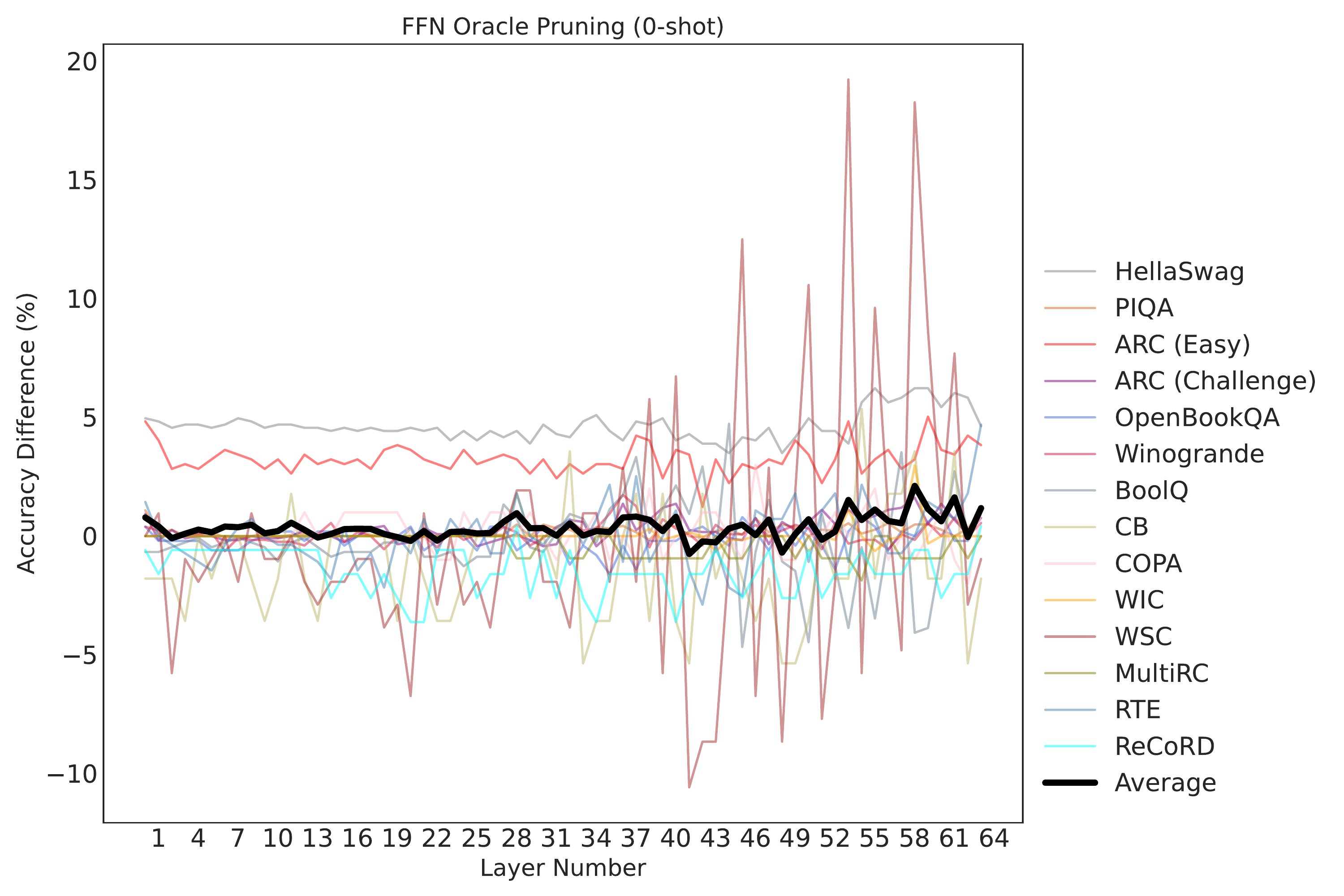}}}
    \subfloat[\centering\label{exp_fig:fc_imp_b} One-shot]{{\includegraphics[width=0.36\linewidth]{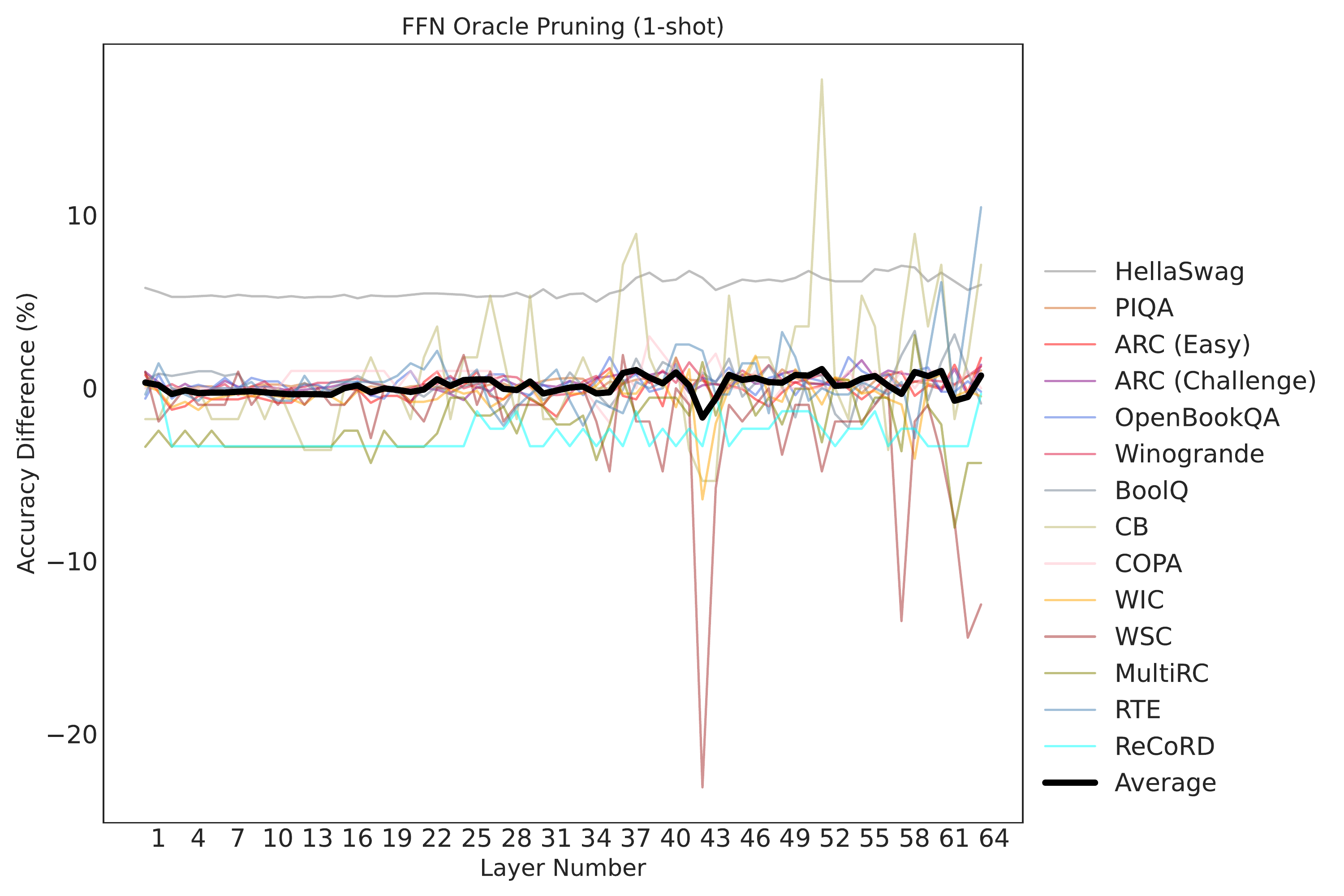}}}
    \subfloat[\centering\label{exp_fig:fc_imp_c} Five-shot]{{\includegraphics[width=0.36\linewidth]{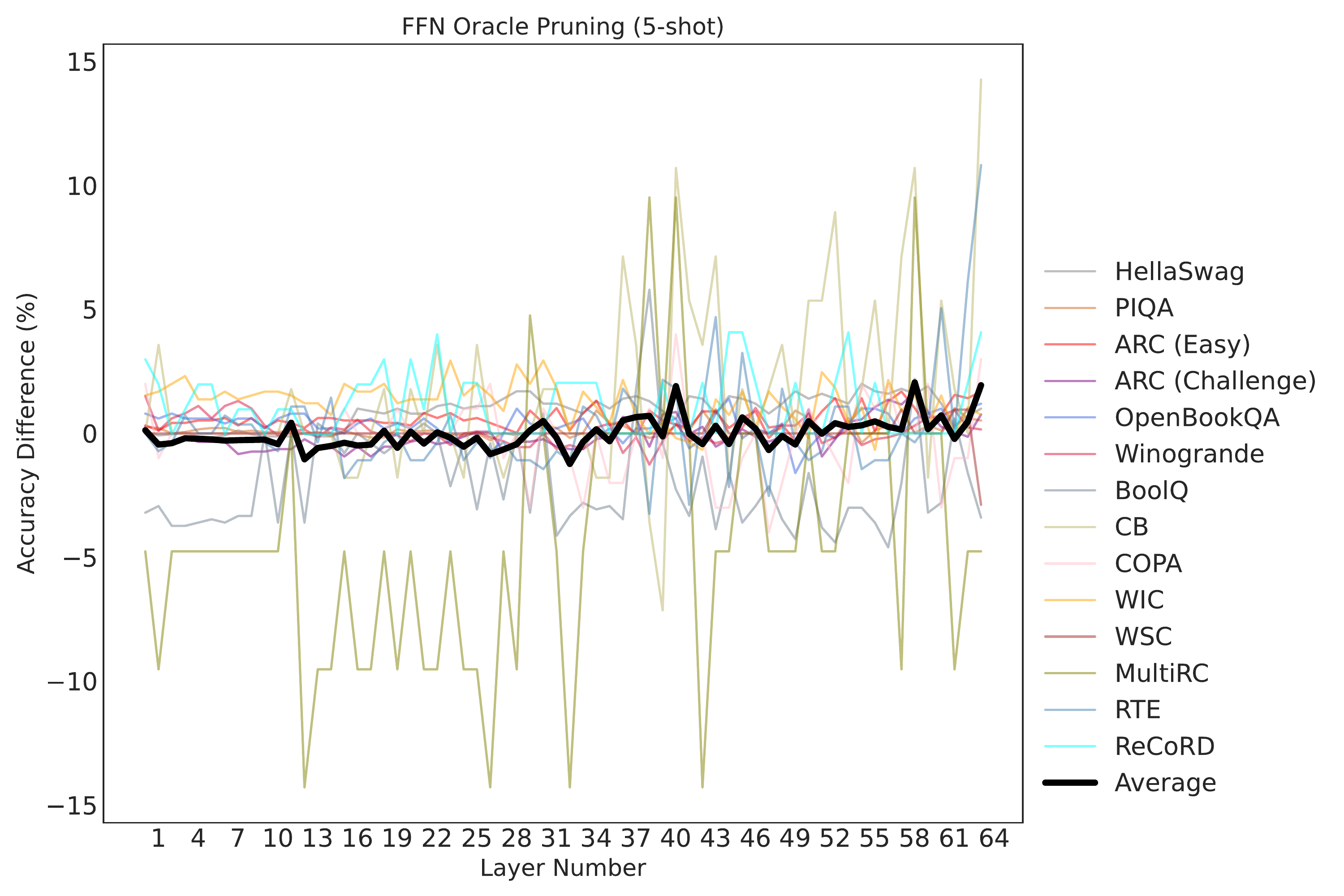}}}
    \caption{Feed forward network (FFN) oracle importance scores for in-context learning with OPT-66B. Each FFN is knocked off independently to compute these scores, i.e., the curves are discrete and \textbf{not} cumulative.}
    \label{exp_fig:fc_imp}
\end{figure*}

We compute oracle importance scores (both task-specific and averaged across tasks) for each of the 64 FFNs in OPT-66B as described in \S \ref{bg:oracle_pruning} in the zero-shot and few-shot settings and present them in Figure \ref{exp_fig:fc_imp}. Note that if the importance score, i.e., accuracy difference, for an FFN $f$ is near zero, $f$ is redundant. If it is negative, the model is better off without $f$, and if it is positive, $f$ is important.

We observe that in the zero-shot and one-shot settings, the removal of any FFN in the early (1-30) layers of OPT-66B either gives comparable or better performance for a vast majority of tasks. In the five-shot setting however, both the early and later layers seem to have important FFNs for most tasks. We also generally observe high variance in FFN importance scores in later layers. We particularly note high variance for WSC and MultiRC, observing that removal of some individual FFNs can lead to absolute accuracy improvements/degradation of up to 20\%! One hypothesis for the cause for such variance could be the more challenging nature of some of these tasks such as MultiRC, which has fairly large sequence lengths, and such FFNs might be relied upon to be able to process challenging task instances. We leave further investigation into the cause for this variance for future work.

\section{Iterative Pruning}
\label{exp:iterative}

The observations in the previous section (\S \ref{exp:individual_components}) indicated that the ability to in-context learn-perform a task is not uniformly spread across all of OPT-66B's underlying components. In this section, we build on these observations and assess to what extent we can remove \textit{multiple} attention heads and FFNs with minimal decline in task performance.

\subsection{Removing Attention Heads}
\label{exp:iterative attention heads}

\begin{figure*}[h]
    \centering
    \subfloat[\centering\label{exp_fig:ip_head_a} Zero-shot]{{\includegraphics[width=0.36\linewidth]{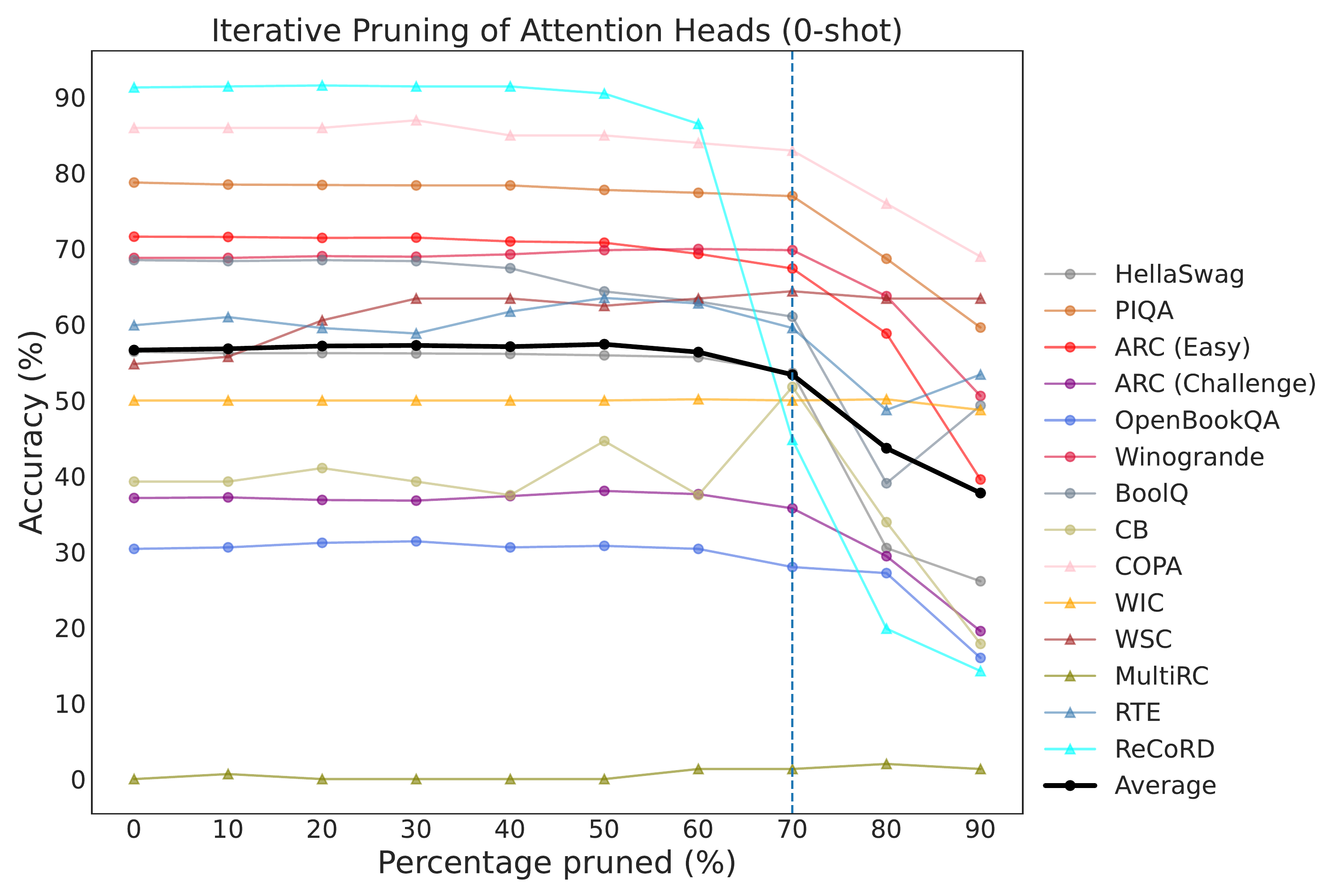}}}
    \subfloat[\centering\label{exp_fig:ip_head_b} One-shot]{{\includegraphics[width=0.36\linewidth]{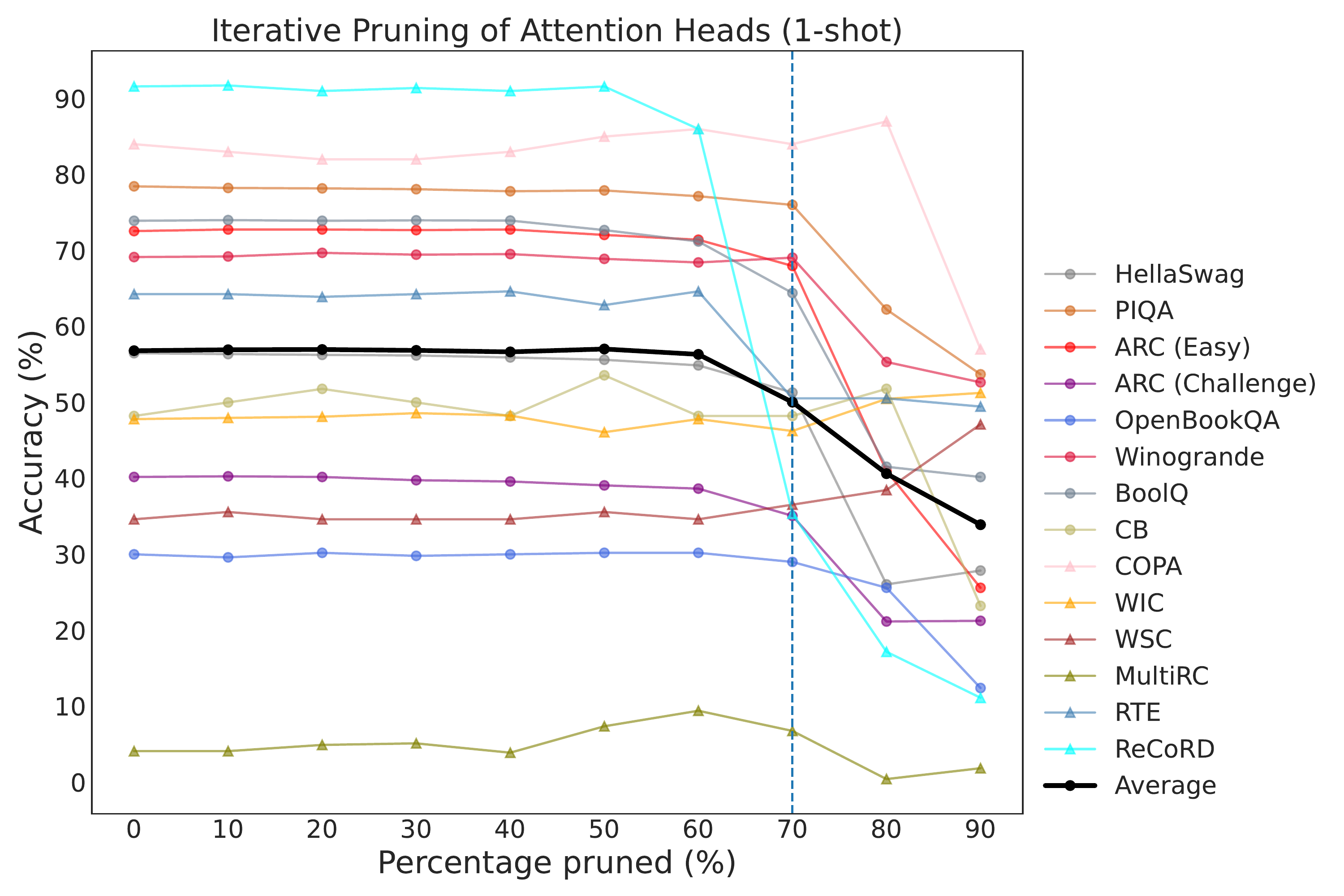}}}
    \subfloat[\centering\label{exp_fig:ip_head_c} Five-shot]{{\includegraphics[width=0.36\linewidth]{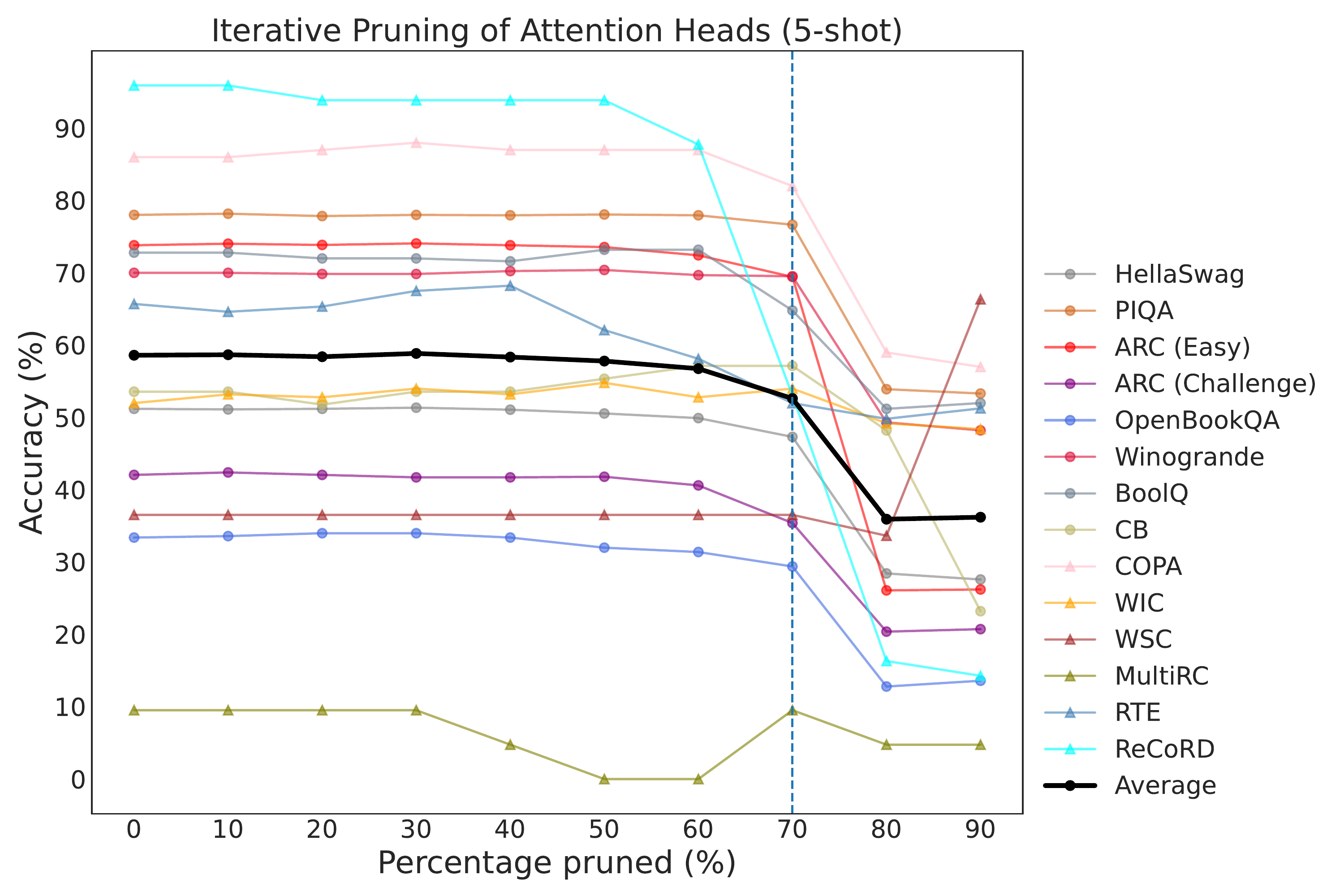}}}
    \caption{Effect on in-context learning accuracy when removing attention heads in OPT-66B in an iterative manner based on task-specific and shot-specific importance scores.}
    \label{exp_fig:ip_head}
\end{figure*}

For each task in each (zero-shot, one-shot and five-shot) in-context learning setting, we sort all attention heads in OPT-66B in ascending order by importance score (\S \ref{exp:attention_importance}). We then remove attention heads in an iterative fashion from the start of this order, 10\% at a time, and re-evaluate task performance after each removal.\footnote{We do not remove attention heads one at a time and re-evaluate given the number of heads and evaluation cost.} Figure \ref{exp_fig:ip_head} depicts the resulting accuracy trends for each task and the trends when averaged across all tasks.

We observe that the average accuracy across tasks does not change much up until $\sim$70\% of the attention heads are removed. A fine-grained look at the individual tasks also mostly shows similar trends, with accuracy staying fairly intact until a large proportion of the attention heads are removed. Some oddities include tasks such as WSC and CB, wherein we observe that the zero-shot accuracy actually increases after the removal of 70\% of the attention heads. As noted by \citet{zhang2022opt}, we also observe low absolute accuracy values on MultiRC for the full model and upon pruning.

\subsection{Removing FFNs}
\label{exp:iterative fc subnetworks}

\begin{figure*}[h]
    \centering
    \subfloat[\centering\label{exp_fig:ip_fc_a} Zero-shot]{{\includegraphics[width=0.36\linewidth]{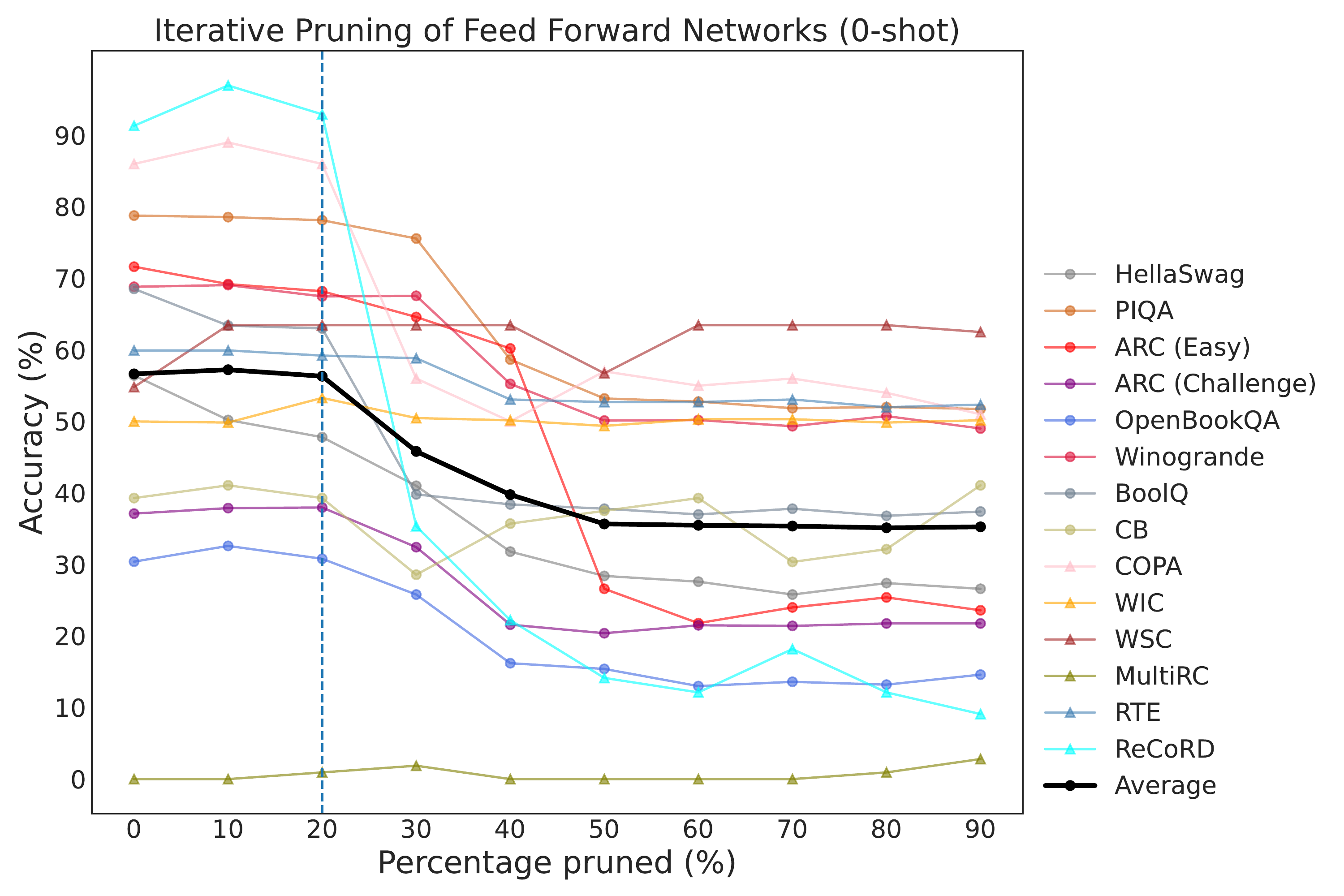}}}
    \subfloat[\centering\label{exp_fig:ip_fc_b} One-shot]{{\includegraphics[width=0.36\linewidth]{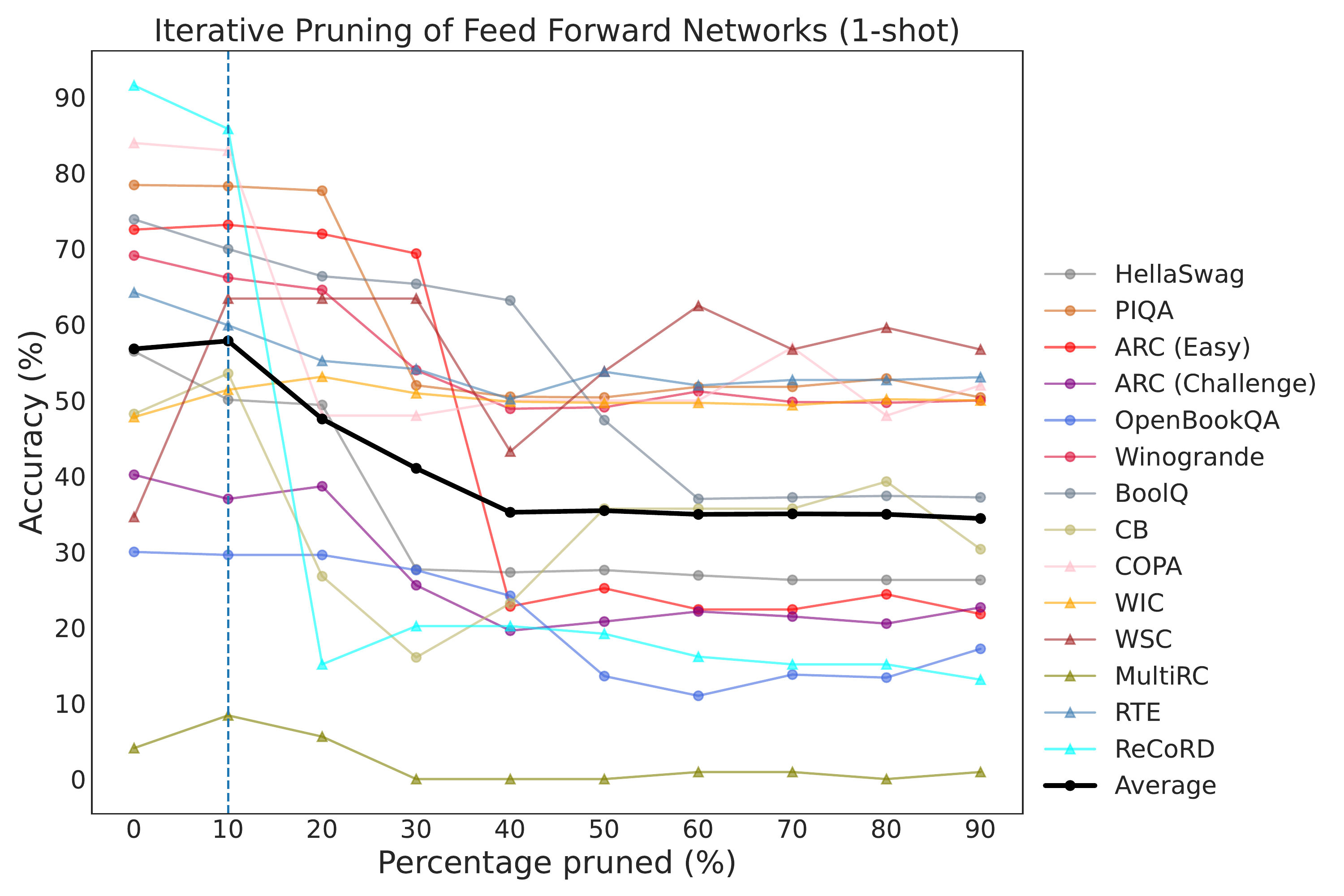}}}
     \subfloat[\centering\label{exp_fig:ip_fc_c} Five-shot]{{\includegraphics[width=0.36\linewidth]{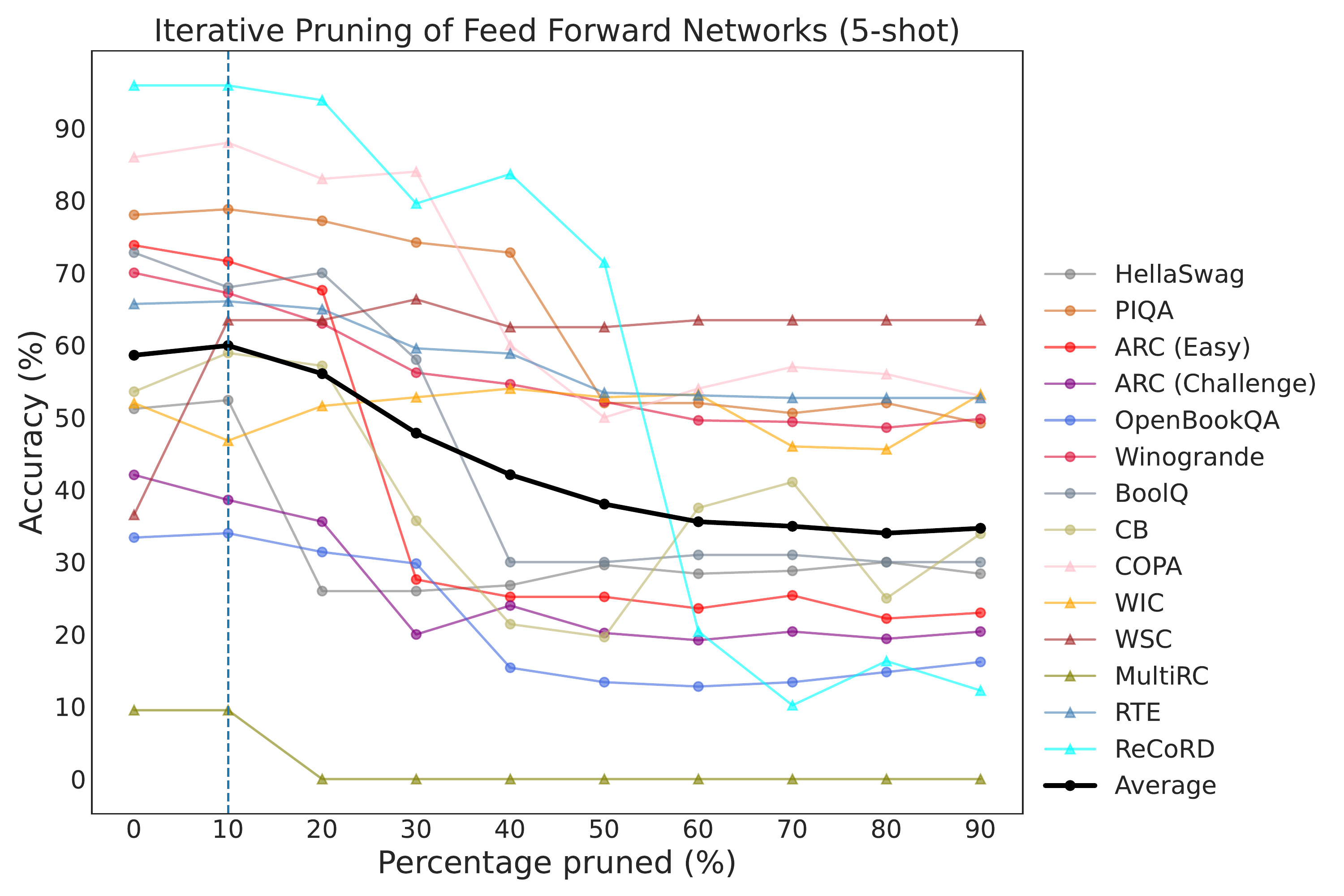}}}
    \caption{Effect on in-context learning accuracy when removing feed forward networks (FFNs) in OPT-66B in an iterative manner based on task-specific and shot-specific importance scores.}
    \label{exp_fig:ip_fc}
\end{figure*}

For each task in each (zero-shot, one-shot and five-shot) in-context learning setting, we sort all FFNs in OPT-66B in ascending order by importance score (\S \ref{exp:fc_score}). We then remove FFNs in an iterative fashion from the start of this order, 10\% at a time, and re-evaluate task performance after each removal. Figure \ref{exp_fig:ip_fc} depicts the resulting accuracy trends for each task and the trends when averaged across all tasks. We observe that in the zero-shot setting, the average accuracy across tasks does not change up until $\sim$20\% of the FFNs are removed. For some tasks such as PIQA, Winogrande and RTE, the accuracy does not change even if 30\% of the FFNs ($\sim$13B of the 66B parameters) are removed. We also observe that the inflection point after which we observe a sharp decline in accuracy changes to 10\% for the few-shot settings. Overall, these observations indicate that FFNs play a critical role toward in-context learning.

\subsection{Combined Removal of Heads \& FFNs}
\label{exp:combined iterative pruning}

\begin{figure*}[h]
    \centering
    \subfloat[\centering\label{exp_fig:cp_a} Zero-shot]{{\includegraphics[width=0.36\linewidth]{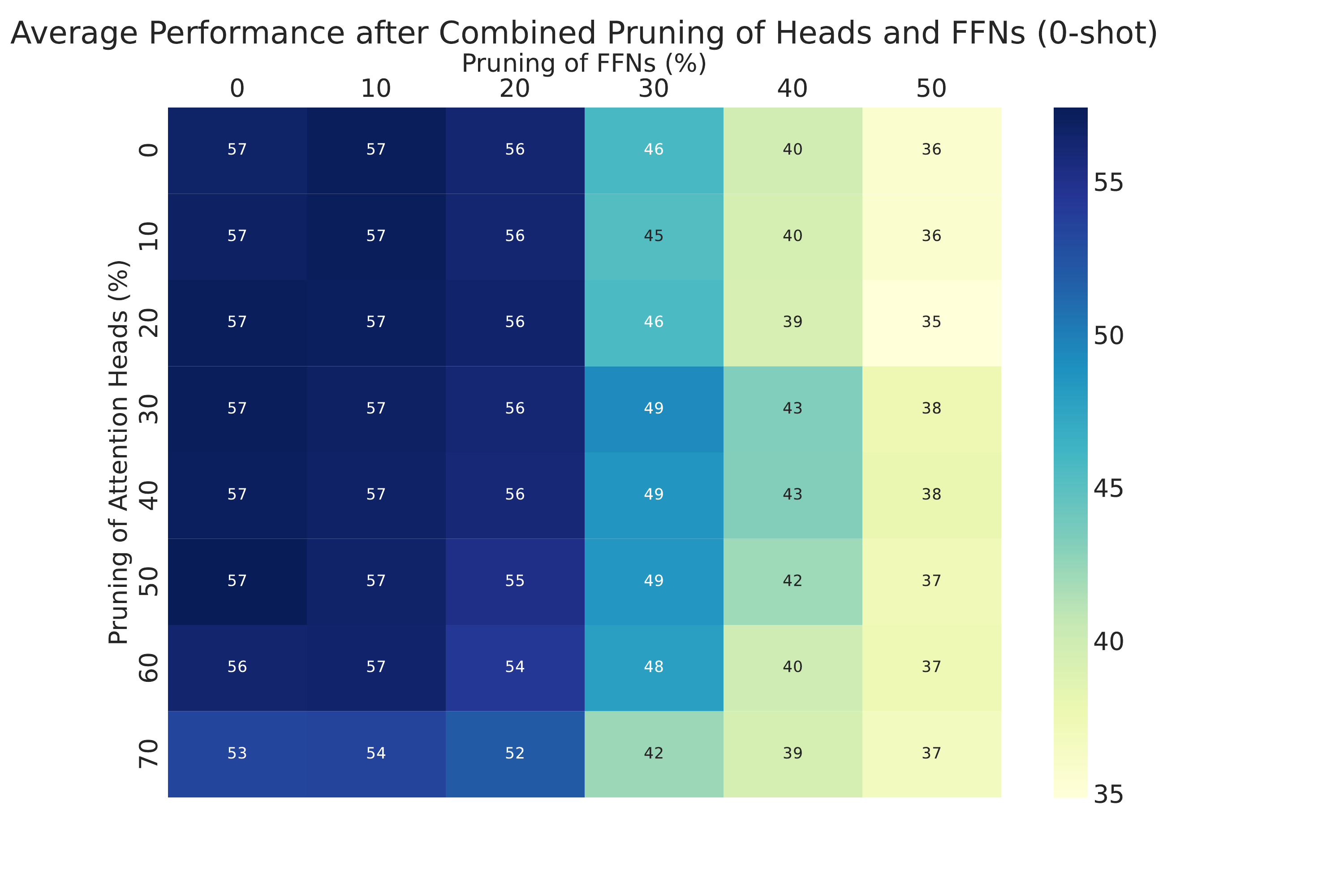}}}
    \subfloat[\centering\label{exp_fig:cp_b} One-shot]{{\includegraphics[width=0.36\linewidth]{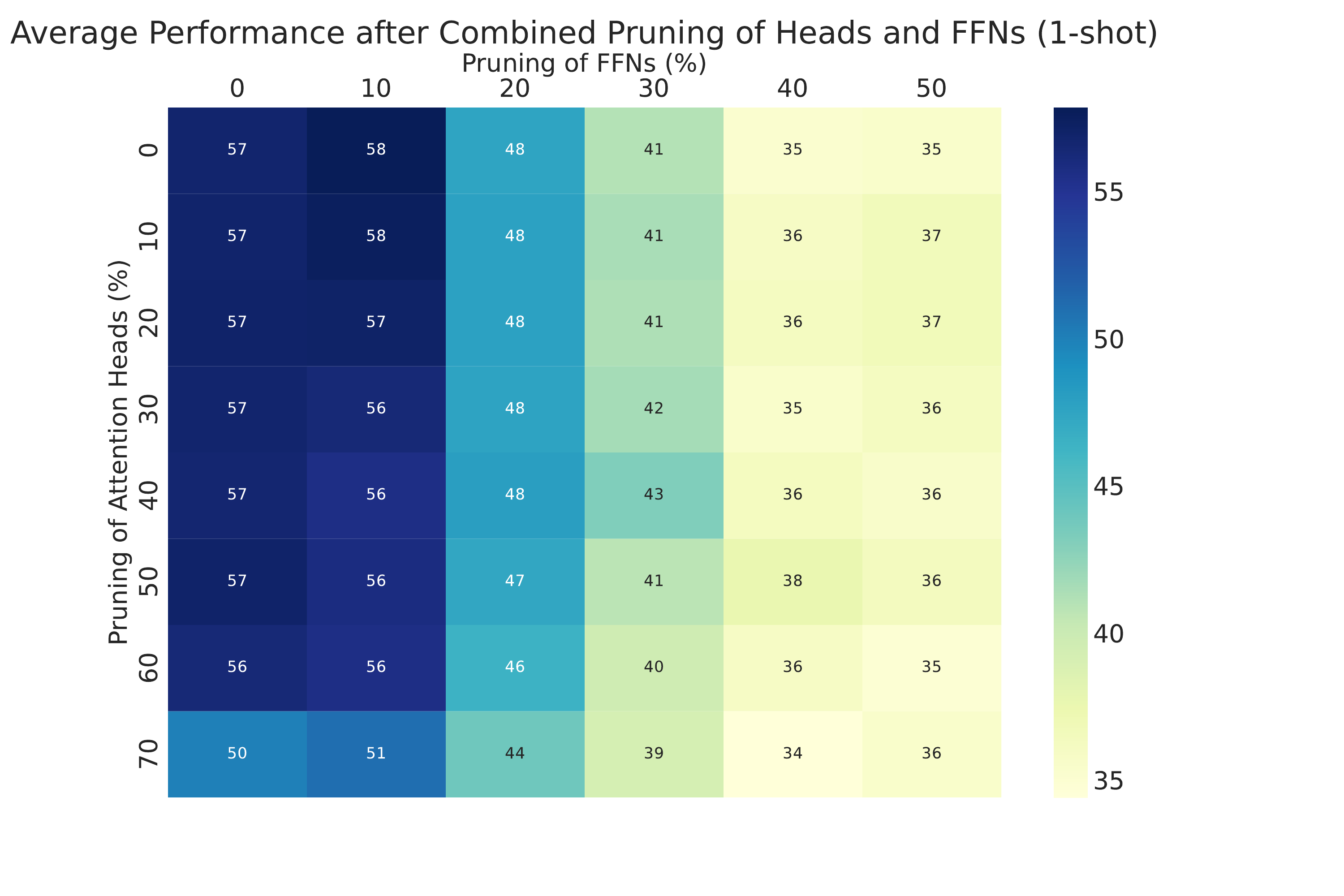}}}
    \subfloat[\centering\label{exp_fig:cp_c} Five-shot]{{\includegraphics[width=0.36\linewidth]{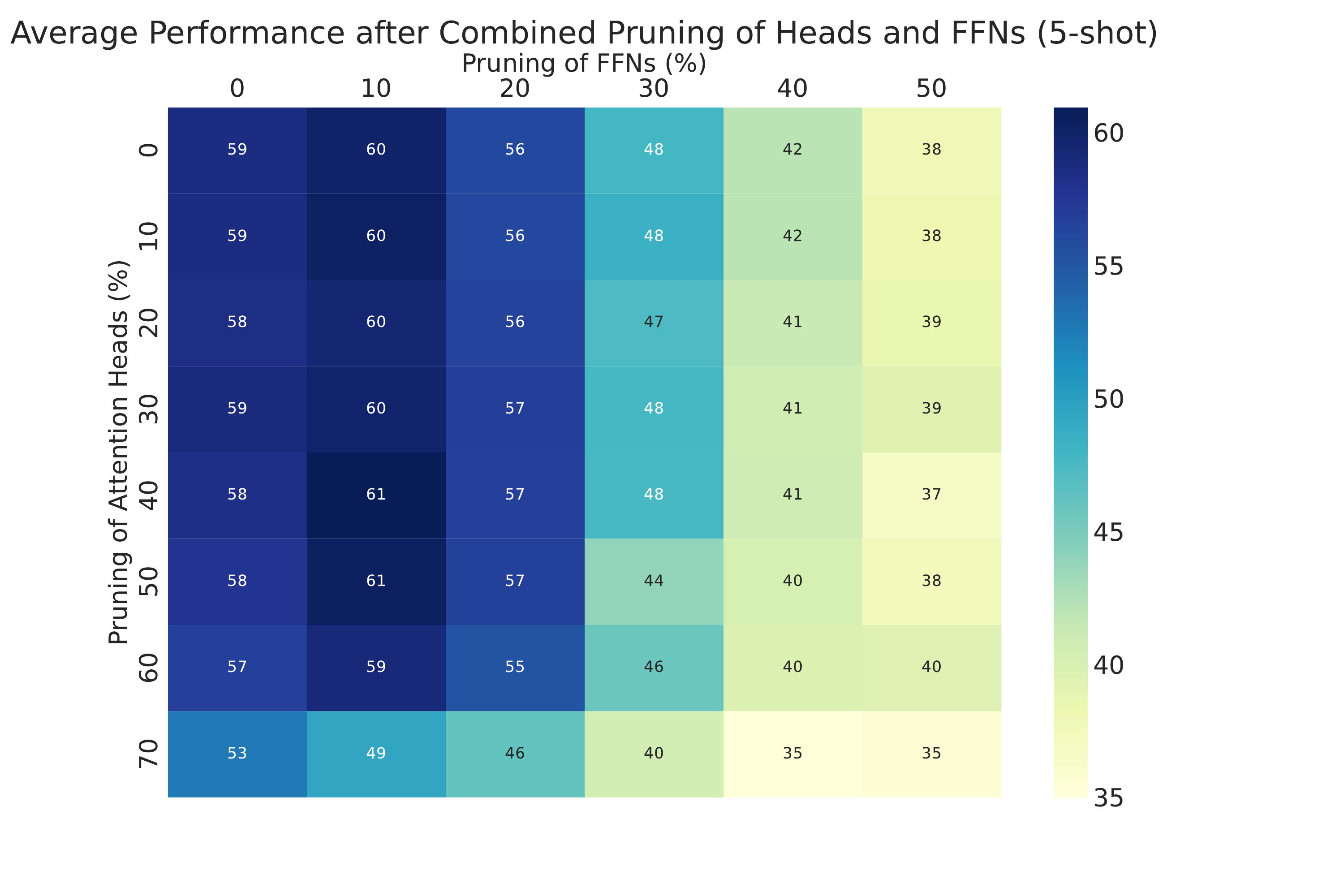}}}
    \caption{Effect on average in-context learning accuracy when removing \textbf{both} attention heads and feed forward networks (FFNs) in OPT-66B in an iterative manner based on shot-specific task-aggregate importance scores.}
    \label{exp_fig:combined_pruning}
\end{figure*}

We now investigate whether the inflection points to in-context learning performance when removing either attention heads or FFNs in an iterative fashion still hold when removing them in \textit{tandem}. Figure \ref{exp_fig:combined_pruning} depicts the average accuracy of all tasks on joint iterative removal of attention heads and FFNs, 10\% at a time, based on their respective importance scores in the zero and few-shot settings.

We observe that the removal of $70\%$ of the attention heads ($\sim$15.7B parameters) and $20\%$ of the FFNs ($\sim$8.5B parameters) leads to a mere $5\%$ absolute drop in the average zero-shot accuracy. In the one-shot setting, the absolute drop in accuracy is $6\%$ on removing $70\%$ of the attention heads and $10\%$ of the FFNs. In the five-shot setting, the absolute drop in accuracy is $4\%$ on removing $60\%$ of the attention heads and $20\%$ of the FFNs. Overall, these new inflection points have deviated by at most a factor of 10\% absolute, which may be attributed to the interplay between heads and FFNs.

\section{Detailed Analysis of Attention Heads}
In this section, we perform a detailed analysis of the attention heads in OPT-66B, given that in-context learning is auto-regressive in nature and attention heads explicitly encode cross-token interactions. We perform cross-task analyses to understand whether various tasks share (un)important attention heads. We also perform cross-shot analyses to study whether the (un)important attention heads for a task are shared across the zero-shot and few-shot settings. We finally quantify the capacity of the attention heads to perform task-agnostic induction operations (defined in \S \ref{bg:in-context}) and study correlations with task-specific importance scores.

\subsection{Cross-Task Analysis}
\label{exp:cross_task}
\citet{michel2019sixteen} found preliminary empirical evidence of the existence of "universally" important attention heads in trained task-specific Transformer and BERT models via evaluating on out-of-domain test sets for machine translation and natural language inference respectively. With similar motivation, we study if the (un)important attention heads identified in various in-context learning settings for OPT-66B are shared across tasks.

\subsubsection{Spearman's Rank Correlation}
\label{exp:spearman_rank}

We assess overlap in (un)important attention heads across tasks by sorting task-specific head importance scores to get head importance rankings and computing the Spearman's rank correlation coefficient (SRCC) between the rankings for every pair of tasks in the zero-shot and few-shot settings. We also sort the task-aggregate head importance scores to get the aggregate ranking and compute the SRCC against the ranking for every constituent task. All pairwise correlations for the zero and one-shot settings are depicted in Figure \ref{exp_fig:spearman}, and depicted for the five-shot setting in Appendix \ref{appendix:spearman rank}.

In both zero and few-shot settings, we observe statistically significant ($p<0.01$) positive correlations in the head importance rankings for every pair of tasks, as well as between every task's ranking and the aggregate ranking. This indicates that the set of (un)important attention heads are clustered together across tasks. We also observe seemingly lower magnitude SRCC values between every task and ReCoRD, a long reading comprehension task which requires commonsense reasoning, indicating the amount of head overlap is proportionally lower.

\begin{figure*}[h]
    \centering
    \subfloat[\centering\label{exp_fig:spearman_a} Zero-shot]{{\includegraphics[width=0.5\linewidth]{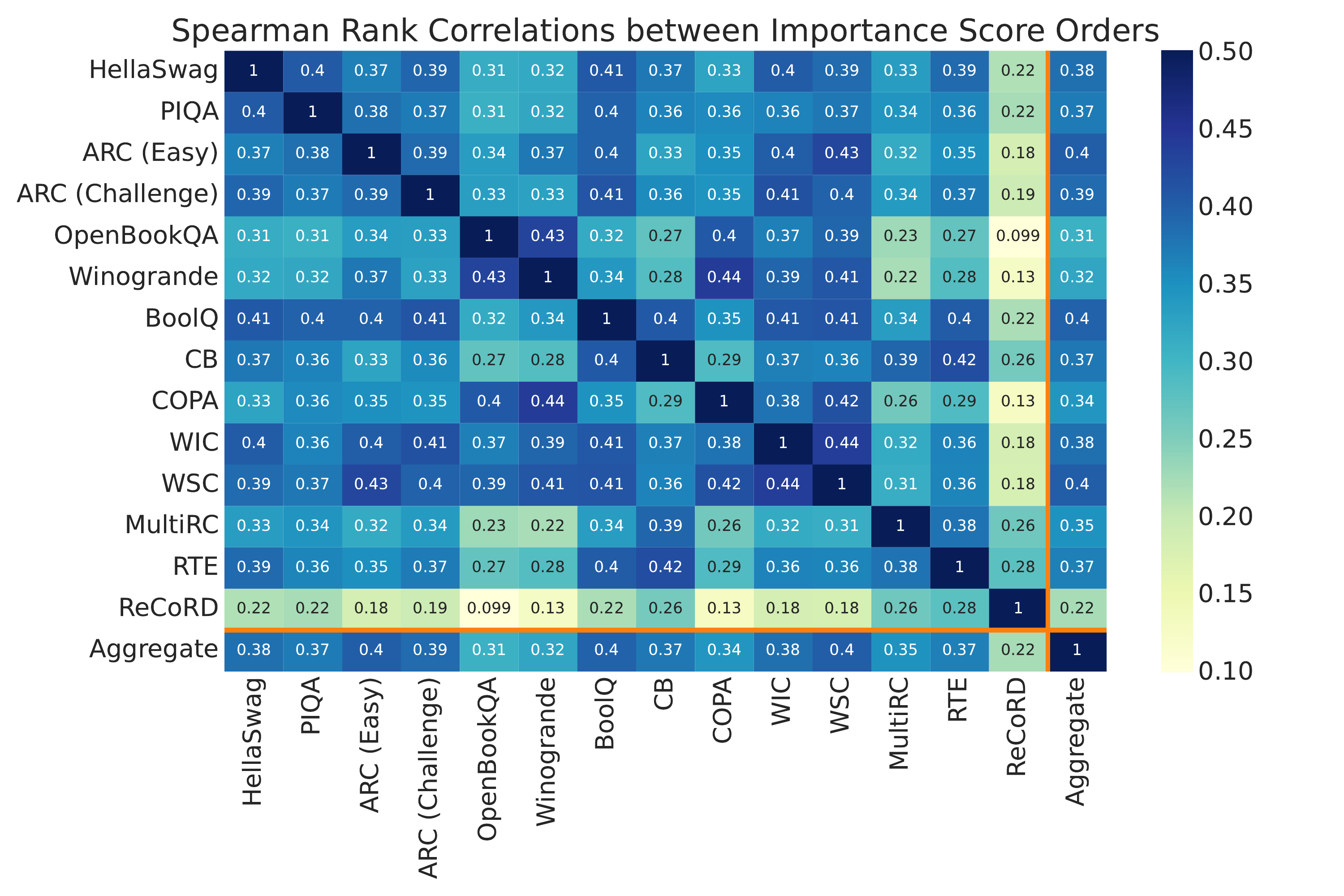}}}
    \subfloat[\centering\label{exp_fig:spearman_b} One-shot]{{\includegraphics[width=0.5\linewidth]{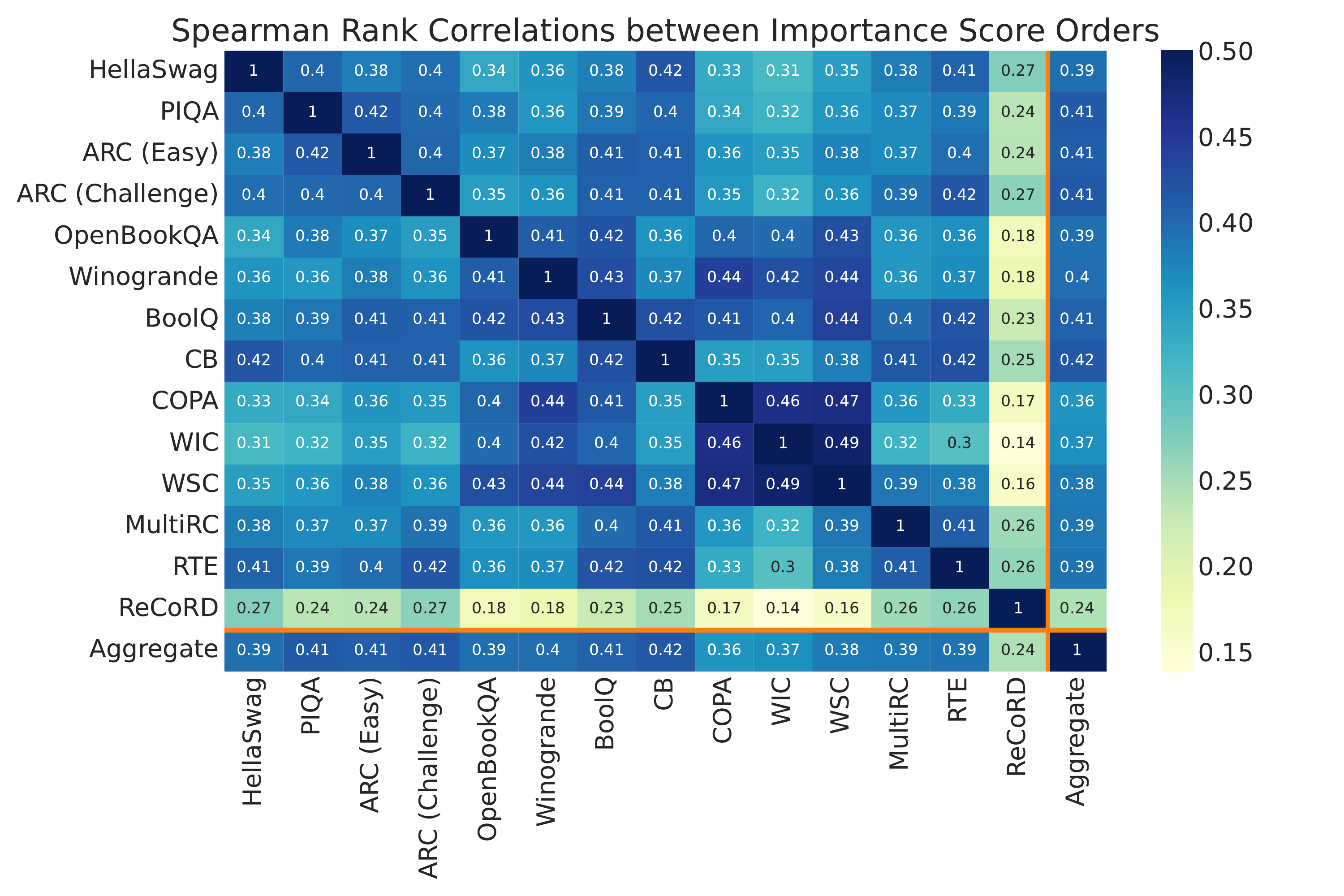}}}
    \caption{Spearman's rank correlation coefficients between the attention head importance rankings for various tasks. All $p$-values $< 0.01$. Coefficients in the five-shot setting are provided in Appendix \ref{appendix:spearman rank}.}
    \label{exp_fig:spearman}
\end{figure*}

\subsubsection{Generalization Trends}
\label{exp:generalization_trends}

We now investigate how well head importance rankings generalize across tasks by studying accuracy trends for a task when pruning using its own head importance ranking as well as using different head importance rankings. We perform this investigation for two sets of tasks.

\begin{figure*}[h]
    \centering
    \subfloat[\centering\label{exp_fig:cross_task_a} COPA]{{\includegraphics[width=0.36\linewidth]{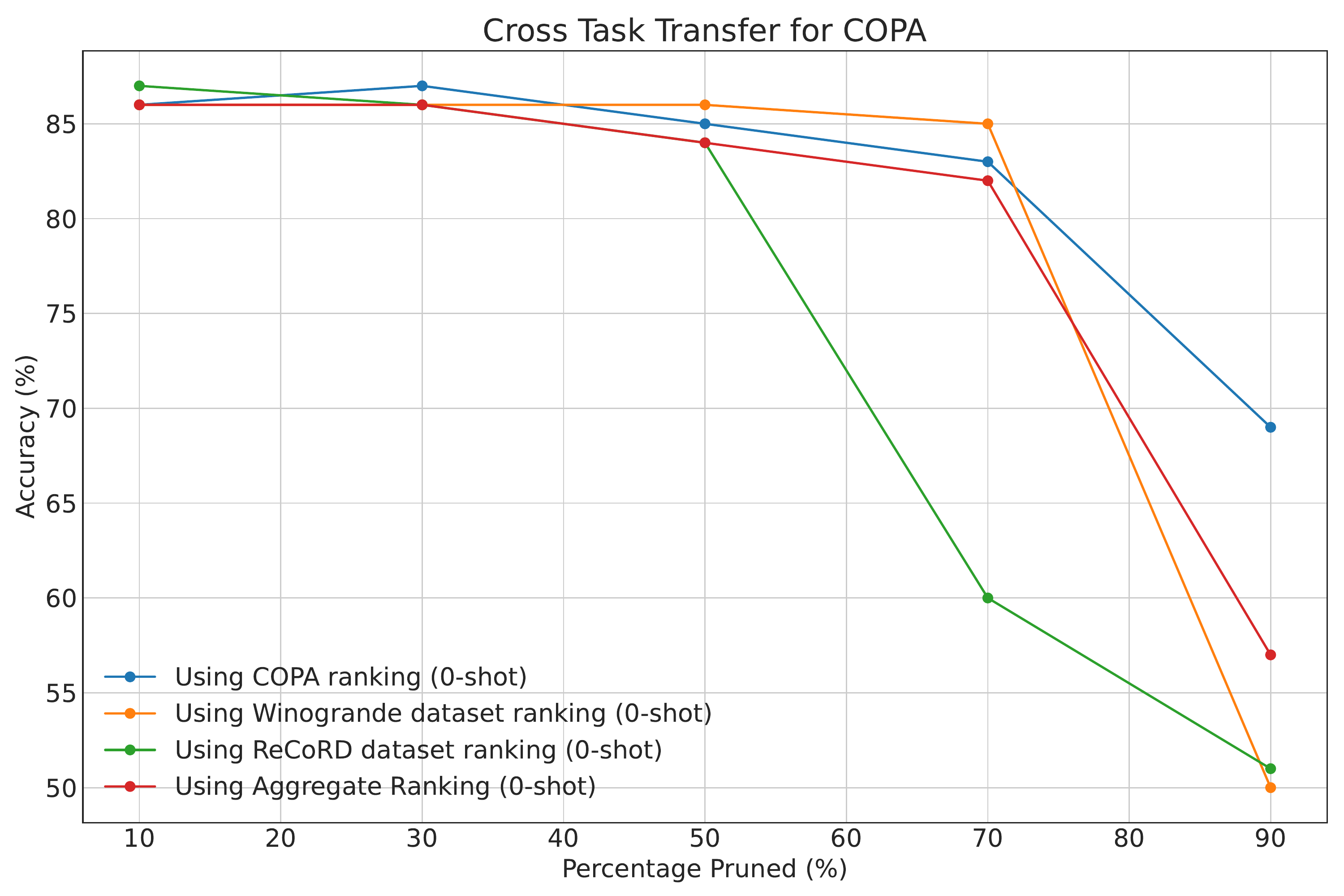}}}
    \subfloat[\centering\label{exp_fig:cross_task_b} Winogrande]{{\includegraphics[width=0.36\linewidth]{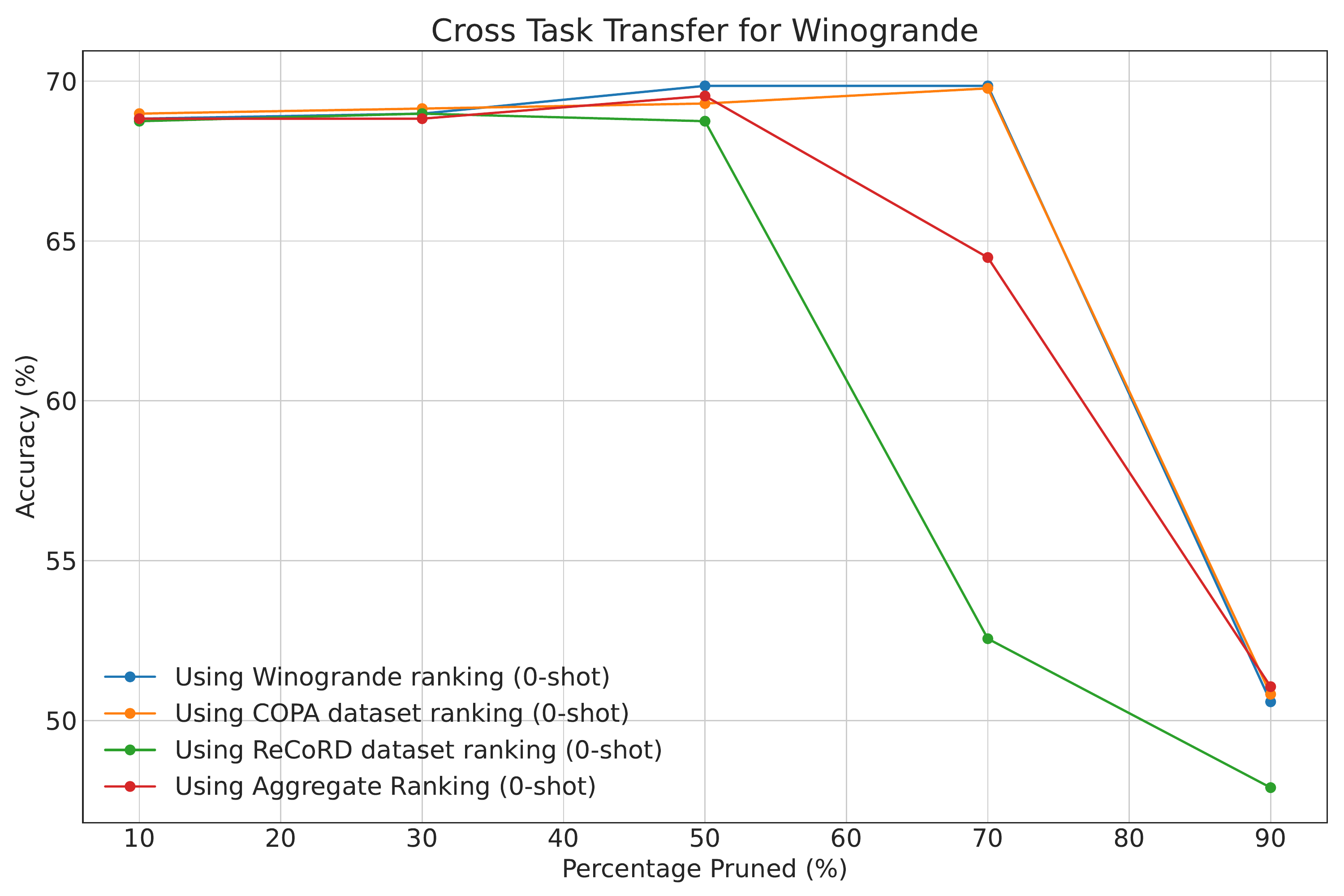}}}
    \subfloat[\centering\label{exp_fig:cross_task_c} ReCoRD]{{\includegraphics[width=0.36\linewidth]{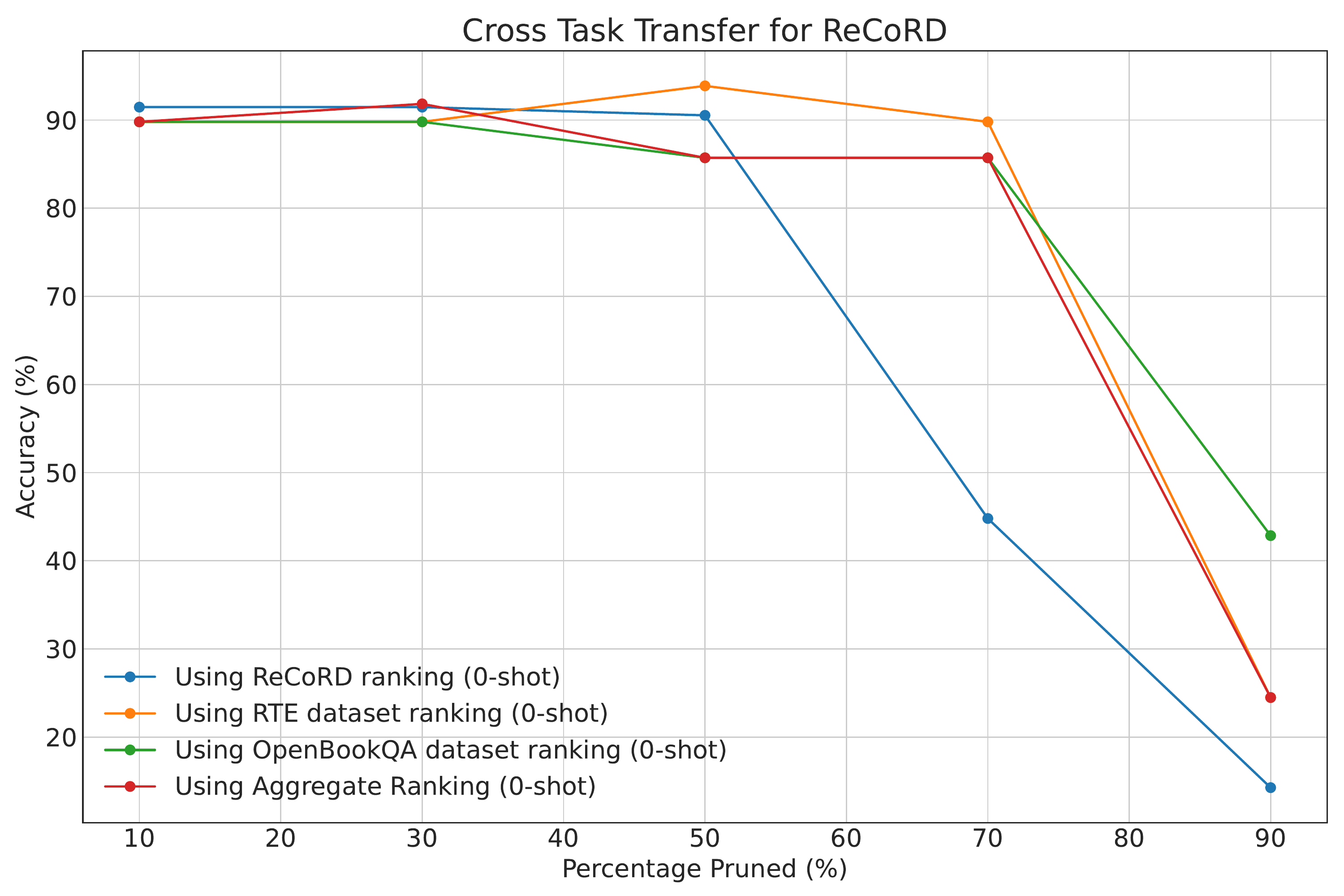}}}
    \qquad
    \subfloat[\centering\label{exp_fig:cross_task_d} MathQA]{{\includegraphics[width=0.4\linewidth]{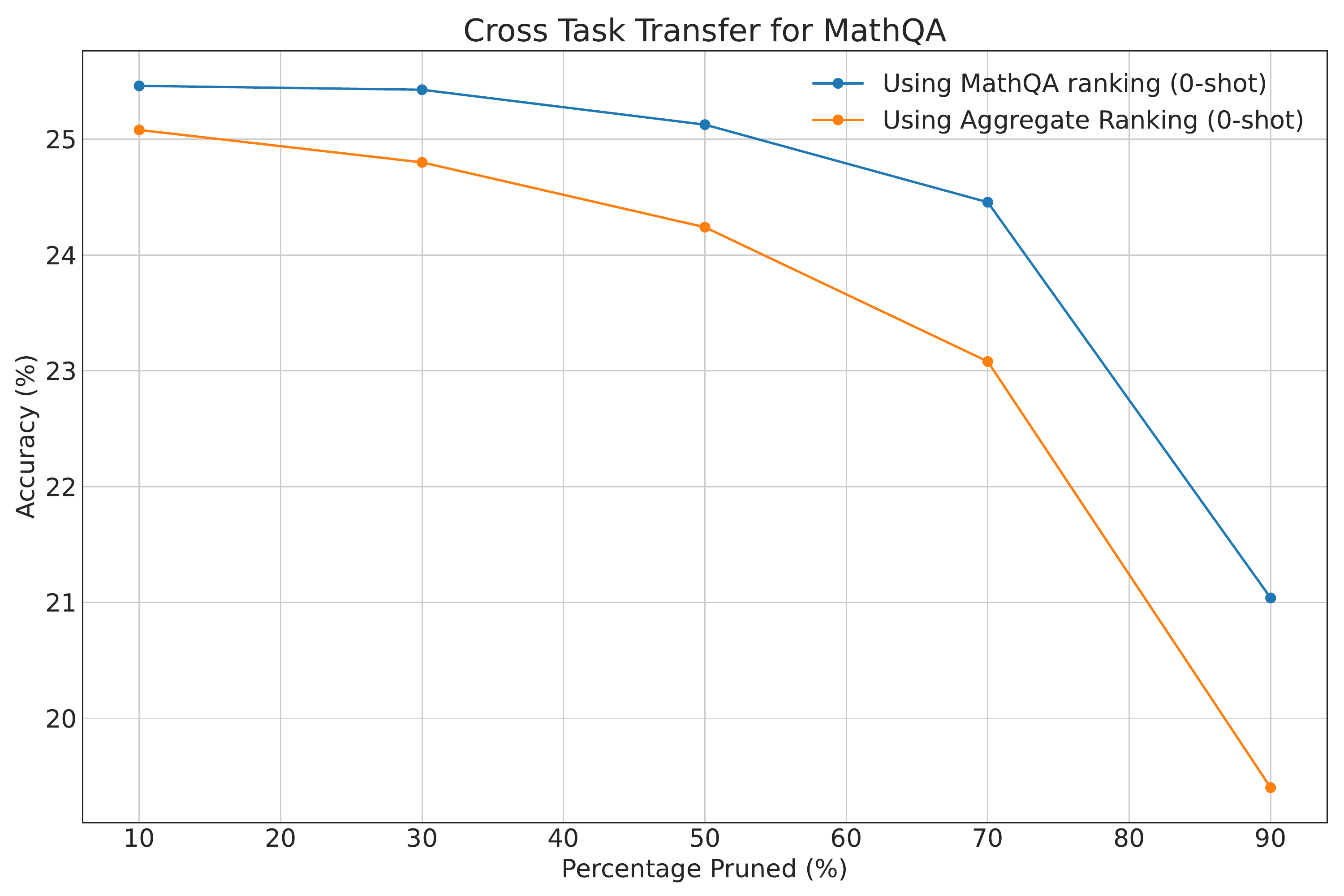}}}
    \subfloat[\centering\label{exp_fig:cross_task_e} LAMBADA]{{\includegraphics[width=0.4\linewidth]{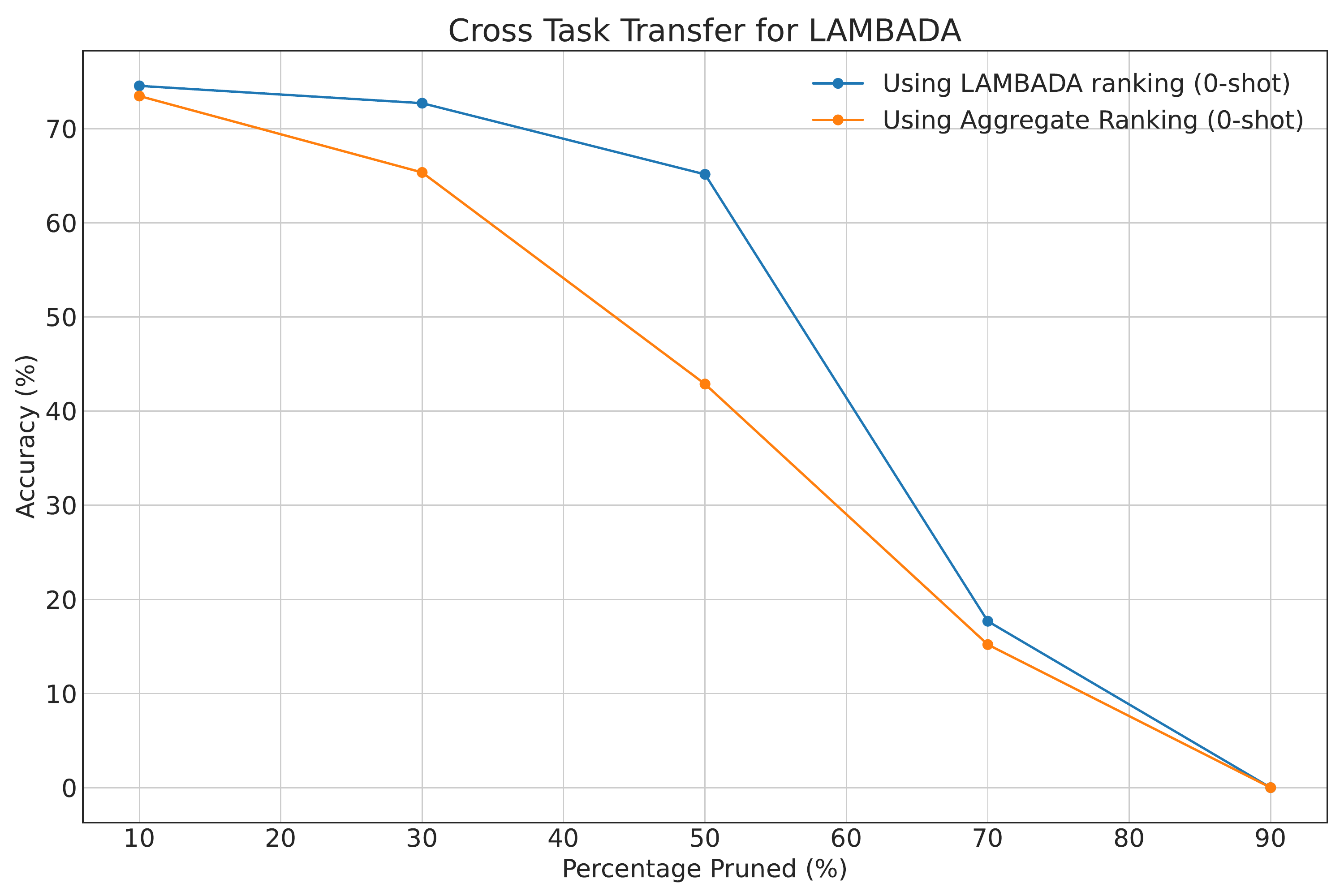}}}
    \caption{Cross-task transfer of attention head importance rankings as measured by impact of pruning on accuracy in the zero-shot setting.}
    \label{exp_fig:cross_task}
\end{figure*}

The first set of tasks we investigate were used to compute the aggregate ranking: COPA, Winogrande and ReCoRD. For each of these 3 tasks, we use the self-ranking, aggregate ranking and the rankings from the tasks which share the \textit{highest} and \textit{lowest} SRCC with them based on Figure \ref{exp_fig:spearman}. For instance, we see that ReCoRD has the highest and lowest SRCCs with RTE and OpenBookQA respectively in the zero-shot setting, so we consider the impact of pruning based on their head importance rankings on ReCoRD accuracy. Figures \ref{exp_fig:cross_task_a}, \ref{exp_fig:cross_task_b} and \ref{exp_fig:cross_task_c} depict the accuracy trends for these 3 tasks in the zero-shot setting. We observe that the accuracy on all 3 tasks when pruning using the rankings described is almost unaffected up to the 50\% mark. We then observe a sharp decline in accuracy on COPA and Winogrande when the model is pruned to the 70\% mark using the ranking identified via ReCoRD, the task with the lowest SRCC (0.13) with both COPA and Winogrande in Figure \ref{exp_fig:spearman_a}. This indicates that even if the rankings vary between ReCoRD and COPA/Winogrande (as reflected in the low magnitude of the SRCC score), the set of attention heads important for zero-shot learning with ReCoRD are important for COPA/Winogrande too. To further verify this, we calculated and found $71\%$ and $76\%$ overlap between the top $30\%$ important attention heads for ReCoRD-COPA and ReCoRD-Winogrande respectively. We also surprisingly observe that the accuracy on ReCoRD at the $70\%$ pruning mark is better using the aggregate ranking than using the ranking for ReCoRD itself. We observe similar trends in the few-shot settings as well, depicted in Appendix \ref{appen:cross_task}. A key distinction we observe relative to the zero-shot setting is that the decline/divergence in accuracy beyond the 50\% pruning mark using the ReCoRD ranking is less sharp for COPA and Winogrande in the one-shot setting and fades away in the five-shot setting, indicating a convergence of important attention heads across tasks.

The second set of tasks we investigate were unseen, i.e., not used to compute the aggregate ranking: MathQA and LAMBADA. For these tasks, we analyze accuracy trends when pruning using the self-ranking and aggregate ranking. Figures \ref{exp_fig:cross_task_d} and \ref{exp_fig:cross_task_e} depict their accuracy trends in the zero-shot setting. As expected, we observe that the self-ranking accuracy curves are somewhat higher than the aggregate ranking accuracy curves in general across both tasks. For MathQA, we also observe that the absolute difference in accuracy for both cases is within 1-2\%. These indicate that the aggregate rankings generalize well to MathQA but not as much to LAMBADA. We observe similar trends in the few-shot settings, depicted in Appendix \ref{appen:cross_task}.

\begin{table}[h]
\centering
\begin{tabular}{l|rrr}
\hline
          & \multicolumn{1}{l}{0-shot} & \multicolumn{1}{l}{1-shot} & \multicolumn{1}{l}{5-shot}  \\ 
\hline
0-shot & $1$                              & $0.39_{\small{\textbf{0.001}}}$                          & $0.37_{\small{\textbf{0.001}}}$                              \\ 
\hline
1-shot  & $0.39_{\small{\textbf{0.001}}}$                           & $1$                             & $0.41_{\small{\textbf{0.001}}}$                              \\ 
\hline
5-shot & $0.37_{\small{\textbf{0.001}}}$                             & $0.41_{\small{\textbf{0.001}}}$                            & $1$                               \\
\hline
\end{tabular}
\caption{Spearman's rank correlation coefficient (SRCC) between attention head importance rankings for different in-context learning settings. Each cell depicts the mean and variance (subscript) in SRCC across 14 tasks. $p$-value $< 0.01$ for every correlation pair.}
\label{exp_table:cross_shot}
\end{table}

\subsection{Cross-Shot Analysis}
\label{exp:cross_shot}

To see if the attention heads that are identified to be (un)important for a task are shared across the different zero and few-shot settings, we compute Spearman's rank correlation coefficient (SRCC) between the cross-shot head importance rankings for each task and compute the mean and variance across all 14 tasks. Table \ref{exp_table:cross_shot} depicts these metrics. We observe that the SRCC is higher for rankings \textit{within} the few-shot setting (0.41 between 1-shot and 5-shot) than for rankings \textit{across} the zero and few-shot settings. This matches the intuition that a similar set of heads must be important within the different few-shot settings than across the zero-shot and any of the few-shot settings. However, we also see that the SRCC magnitudes for the latter (0.39 and 0.37) are not very far off. In totality, these indicate non-trivial overlap in the (un)important attention heads for tasks across shots.

\subsection{Induction Heads in OPT-66B}
\label{exp:ind_heads}
We look for induction heads in OPT-66B by quantifying the capacity of all attention heads to perform prefix matching and copying using random input sequences in a task-agnostic fashion, following the definition and algorithms by \citet{olsson2022context} discussed in \S \ref{bg:in-context} and Appendix \ref{appen:pm_cs}.

\begin{figure}[h]
    \centering
    \subfloat[\centering\label{exp_fig:pm} Prefix Matching]{{\includegraphics[width=\linewidth]{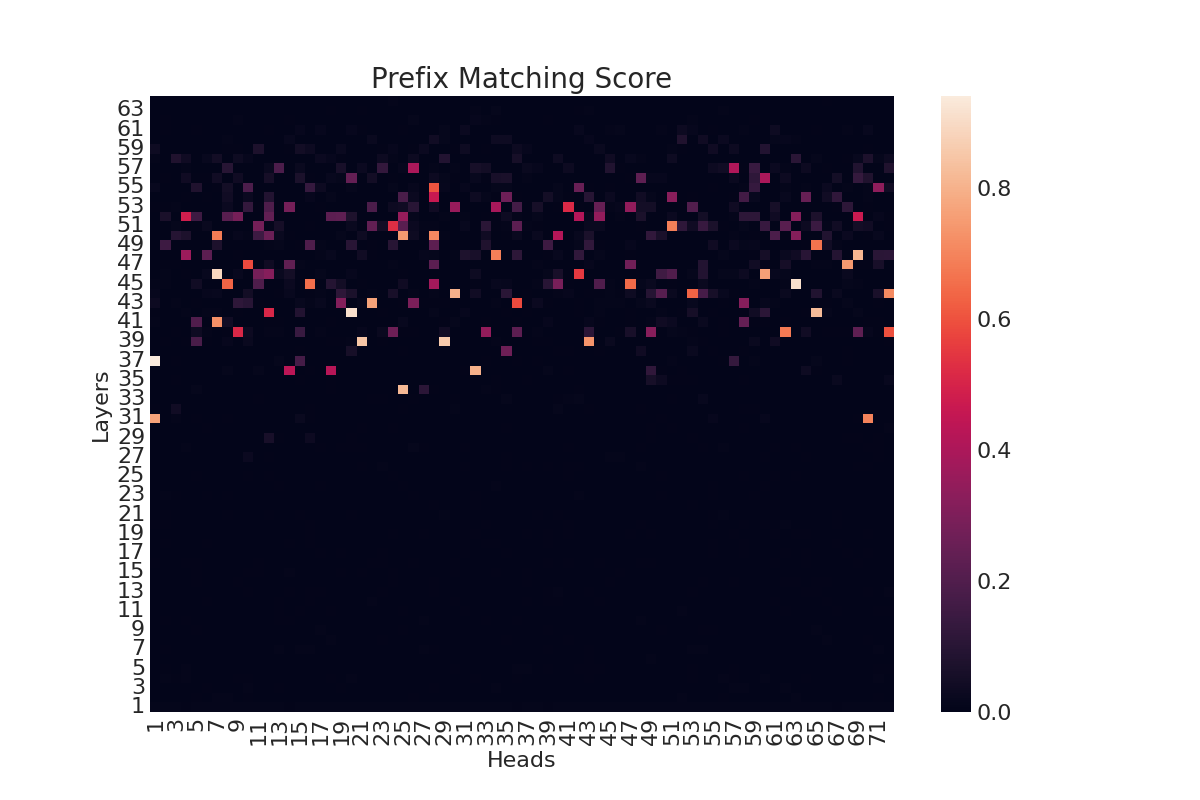}}}
    \qquad
    \subfloat[\centering\label{exp_fig:cs} Copying]{{\includegraphics[width=\linewidth]{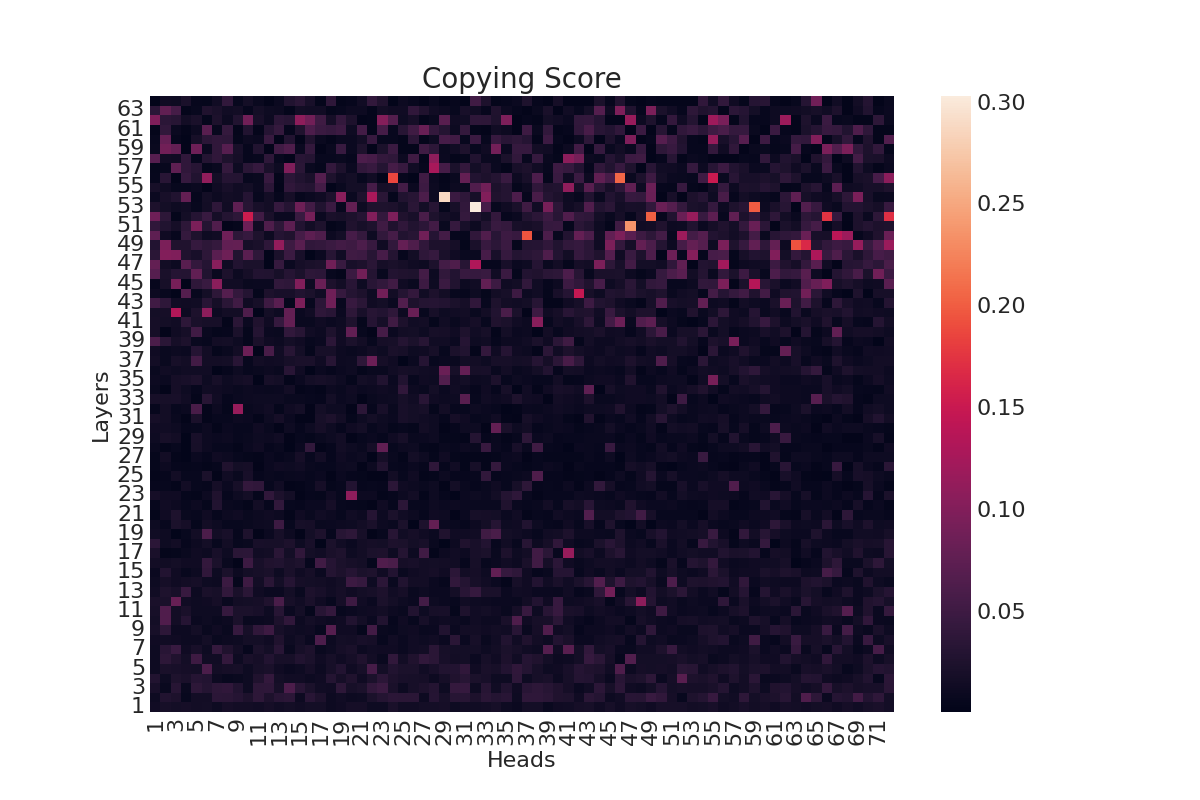}}}
    \caption{Attention head prefix matching and copying score heatmaps for OPT-66B.}
    \label{exp_fig:pm_cs}
\end{figure}

Figures \ref{exp_fig:pm} and \ref{exp_fig:cs} depict the prefix matching and copying score heatmaps respectively for OPT-66B. We observe that a small subset of attention heads in OPT-66B have high prefix matching scores, located in the upper layers (31+) of the model. On the other hand, there are a relatively larger number of attention heads with high copying scores, although the vast majority of these are also located in the upper layers (41+). When seen in conjunction, these observations indicate that there is a sparse set of attention heads that are capable of performing both primitive operations and thus can be deemed plausible induction heads.

\subsubsection{Are Induction Heads Important?}
\label{exp:prune_prefixmatch_copy}

We previously (\S \ref{exp:attention_importance}) computed task-specific and task-aggregated importance scores for attention heads in the zero-shot and few-shot in-context learning settings. By virtue of being important for the challenging downstream tasks we consider, these attention heads are capable of sophisticated and latent behaviors associated with in-context learning that go well beyond the basic primitives of explicit prefix matching and copying. We now study whether induction heads overlap with attention heads deemed important for our chosen tasks.

\begin{figure}[h]
    \centering
    \subfloat[\centering\label{exp_fig:attn_pm} Prefix Matching]{{\includegraphics[width=\linewidth]{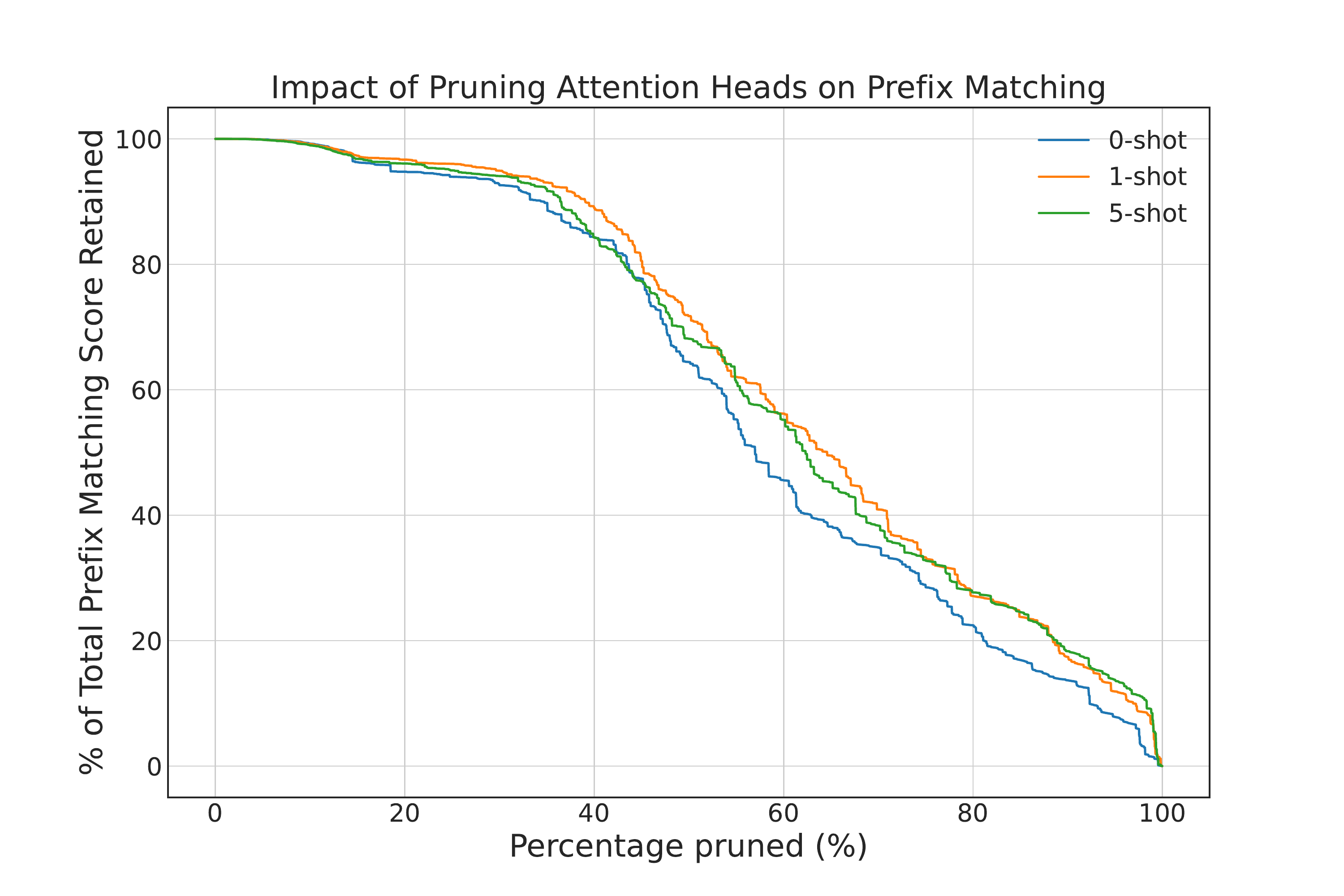}}}
    \qquad
    \subfloat[\centering\label{exp_fig:attn_cs} Copying]{{\includegraphics[width=\linewidth]{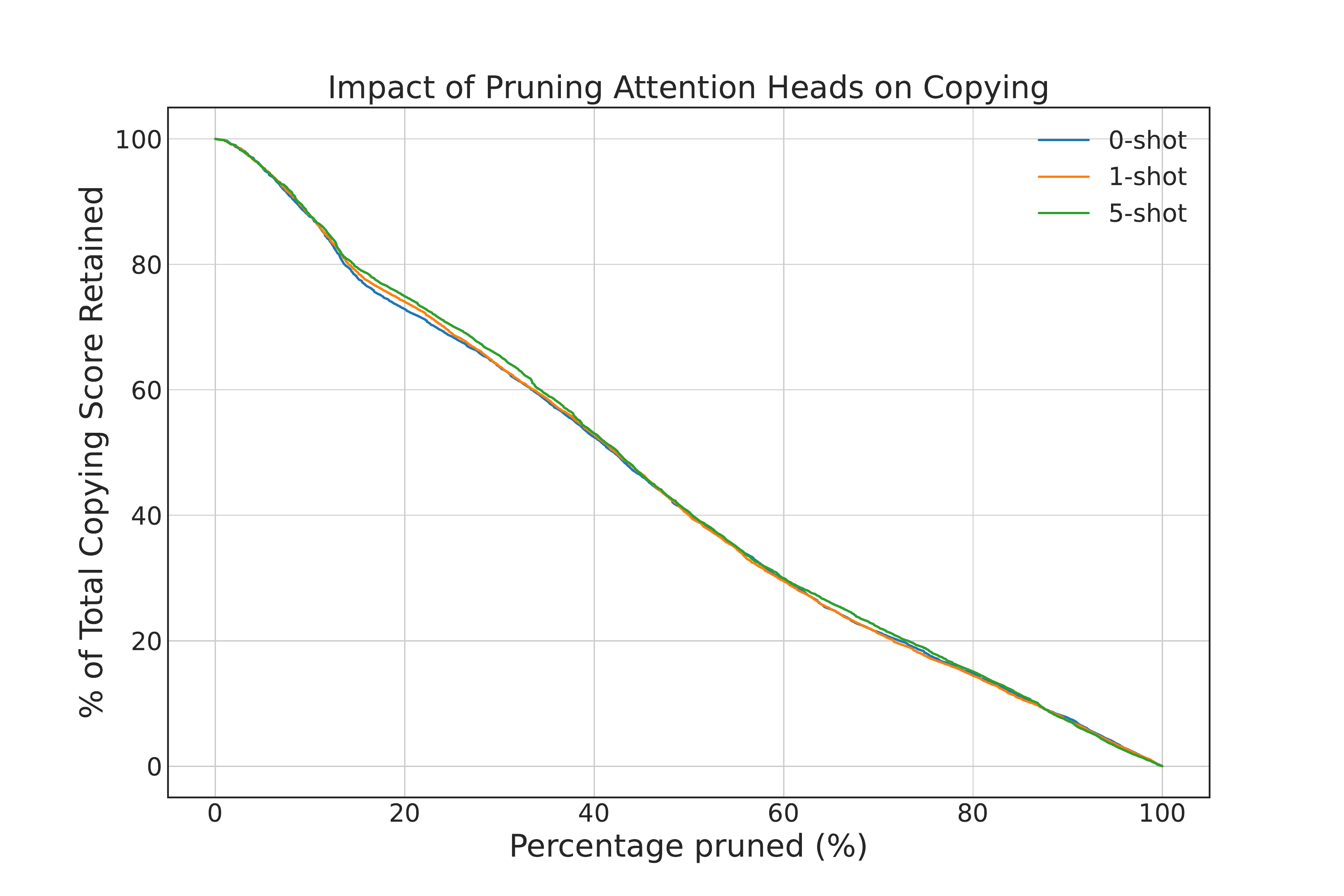}}}
    \caption{Total inductive (prefix matching / copying) capacity retained as a function of percentage of attention heads pruned, where heads are pruned based on task-aggregate importance score rankings in the order of least important first.}
    \label{exp_fig:attn_pm_cs}
\end{figure}

A qualitative comparison of the heatmaps in Figure \ref{exp_fig:pm_cs} against the heatmaps in Figure \ref{exp_fig:heatmap} indicate that induction heads do overlap with task-aggregated important attention heads. To better facilitate this comparison, we first formalize the total capacity of a model to perform prefix matching (or copying) to be the sum of the respective scores for individual attention heads in the model. We then investigate how much of this capacity is retained when attention heads are pruned in the order of least important heads first. Figure \ref{exp_fig:attn_pm_cs} depicts this comparison. We observe that much of the total prefix matching score is retained when 20\% of the least important heads are removed, with the slope of decline becoming sharp only after the 40\% pruning mark. This indicates that unimportant heads also have low prefix matching scores. We also observe that the prefix matching scores are generally higher for heads important for few-shot in-context learning than for heads important for zero-shot learning. On the other hand, we observe across the zero-shot and few-shot settings that the total copying score retained on pruning attention heads rapidly and consistently declines, indicating that even unimportant heads have a non-trivial capacity to perform copying. When seen in conjunction, these observations indicate that induction heads in OPT-66B are capable of sophisticated behaviors associated with in-context learning popular downstream NLP tasks and reinforce the induction head generality arguments \citet{olsson2022context} make in the context of smaller models with stylized and synthetic tasks. We provide further evidence of this in Appendix \ref{appen:pm_cs_task} with per-task plots which showcase that some tasks rely on induction heads more than other tasks.

\section{Related Work}

There has been an interest in effectively \textit{leveraging} the in-context learning paradigm \cite{zhao2021calibrate,holtzman-etal-2021-surface,min-etal-2022-noisy,liu-etal-2022-makes,lu-etal-2022-fantastically,rubin-etal-2022-learning,mishra-etal-2022-reframing} ever since its introduction by \citet{brown2020language}, but there have been relatively fewer studies toward better \textit{understanding} the paradigm itself. \citet{xie2021explanation} cast in-context learning as implicit Bayesian inference where the language model implicitly infers a shared concept among in-context examples when making a prediction, showing that this occurs when the pre-training distribution is a mixture of Hidden Markov Models (HMMs) over latent concepts despite a mismatch with the prompt distribution. \citet{min2022rethinking} study the role of the in-context examples themselves, finding that the ground-truth labels are not needed in the examples and that the more important drivers are provision of the label space, the distribution of the input text and the overall format of the sequence. \citet{garg2022can} showcase that Transformer models trained from scratch can in-context learn the class of linear functions with performance comparable to the optimal least squares estimator even under distribution shifts. \citet{razeghi2022impact} showcase that in-context learning performance is correlated strongly with term frequencies in the pre-training corpora used. \citet{olsson2022context} consider an alternate framing of in-context learning as the ability of a language model to better predict tokens later in the context than tokens earlier and hypothesize the existence of induction heads that are responsible for in-context learning. \citet{chan2022transformers} explore training Transformers separately on controlled stimuli and natural language to showcase that Transformers exhibit striking differences in their generalization from in-context vs. in-weights information.

Several works have also focused on analyzing and interpreting how attention works. \citet{vig2019analyzing} performed a study on GPT-2 \textit{small} with Wikipedia sentences, finding that attention targets different parts of speech at different layer depths and aligns with dependency relations most strongly in the middle layers. \citet{tenney-etal-2019-bert} showcase that BERT encodes the classical NLP pipeline in an interpretable way across layers. There are works relying on different formulations for head importance, such as layer-wise relevance propagation \cite{voita2019analyzing}, gradient-based importance and oracle knock-off importance \cite{michel2019sixteen}, with small task-specific trained models and report the existence of specialized heads. Given the recent trend of increasing model scale \cite{lieber2021jurassic,chowdhery2022palm,smith2022using,rae2021scaling,hoffmann2022training} toward tuning-free general-purpose language models that exhibit emergent in-context learning abilities, we draw and build on prior work to understand just how much scale is really needed and/or used for in-context learning \textit{downstream}, an aspect somewhat eclipsed by the focus on the pre-training loss curve in scaling laws \cite{hoffmann2022training}. It is also worth noting that some of our empirical observations rely on a simple greedy approach to training-free pruning since our focus was not to optimally prune a language model with respect to performing in-context learning. \citet{li2021differentiable} show the greedy approach is sub-optimal and produces under-estimates and \citet{halabi2022data} account for the need to re-compute importance scores after removal of each attention head or FFN by formulating pruning as weakly sub-modular maximization.

\section{Conclusion \& Future Work}
In this paper, we studied the efficacy of attention heads and feed forward networks (FFNs) in a large language model (OPT-66B) in performing in-context learning in both task-specific and task-agnostic settings. We observed that while in-context learning may have emerged via self-supervised pre-training at scale, only a core nucleus of attention heads and FFNs seem to be important for in-context learning across a wide variety of downstream tasks. We observed that a small set of attention heads have the capacity to perform task-agnostic primitive induction operations associated with in-context learning, namely, prefix matching and copying. We also saw that these induction heads overlap with task-specific important attention heads to varying degrees, indicating that induction heads are capable of more sophisticated forms of in-context learning and reinforcing arguments \cite{olsson2022context} about their generality. Overall, our in-context learning-centric observations complement recent work \cite{hoffmann2022training} in indicating that large language models may be under-trained and motivate several interesting directions for future work. While induction heads are formed naturally during self-supervised pre-training in its current form, we believe it may be possible to increase the number and strength of induction heads formed by defining auxiliary pre-training objectives for primitives like prefix matching and copying. More generally, it may also be prudent to investigate and improve (pre-)training regimes to increase the number of important model components to in-context learn-perform a wide variety of downstream tasks. Multi-task instruction-tuning likely belongs to this category and it would be interesting to replicate our study with now increasingly accessible instruction-tuned model variants (such as OPT's instruction meta-learned variant OPT-IML).


\newpage

\bibliography{anthology,custom}
\bibliographystyle{acl2023_natbib}
\appendix
\section{Appendix}

\subsection{Task-Specific Head Importance Scores}
\label{app:head_imp_pertask}

\begin{figure*}[h]
    \centering
    
    \subfloat[\centering HellaSwag]{{\includegraphics[scale=0.2]{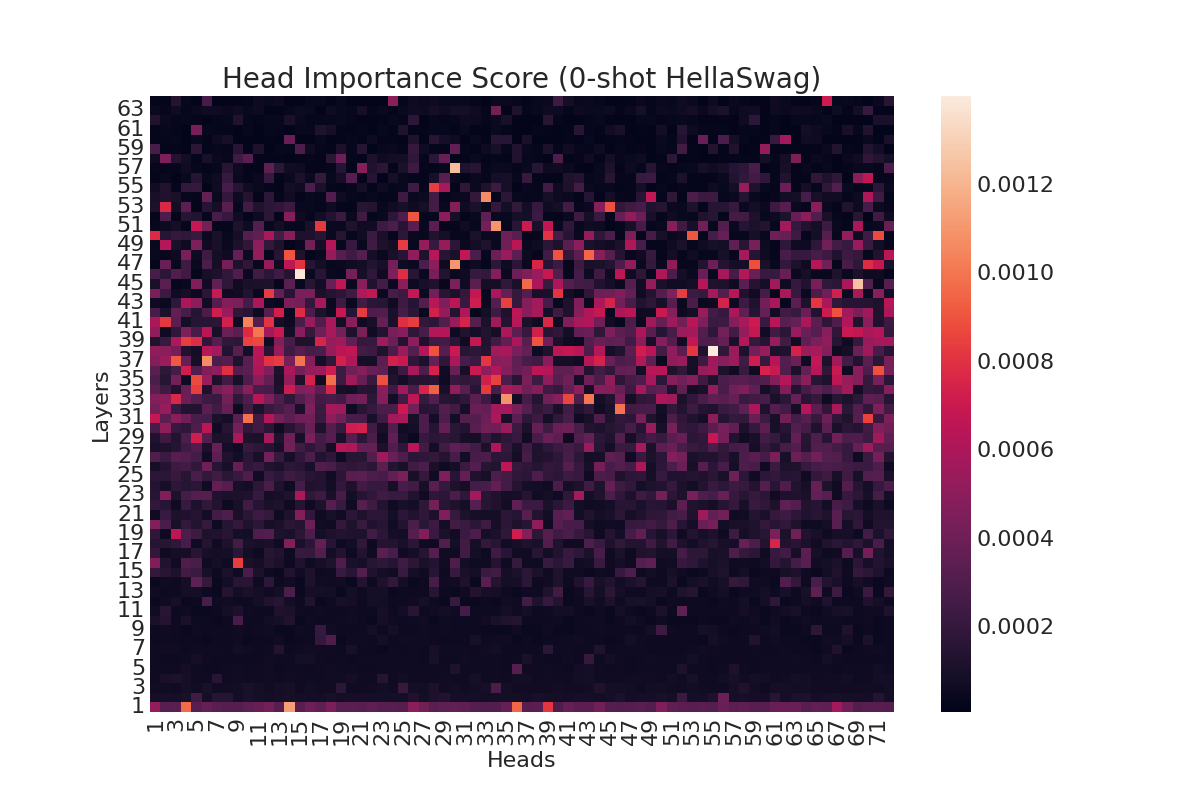}}}
    \subfloat[\centering ARC (Easy)]{{\includegraphics[scale=0.2]{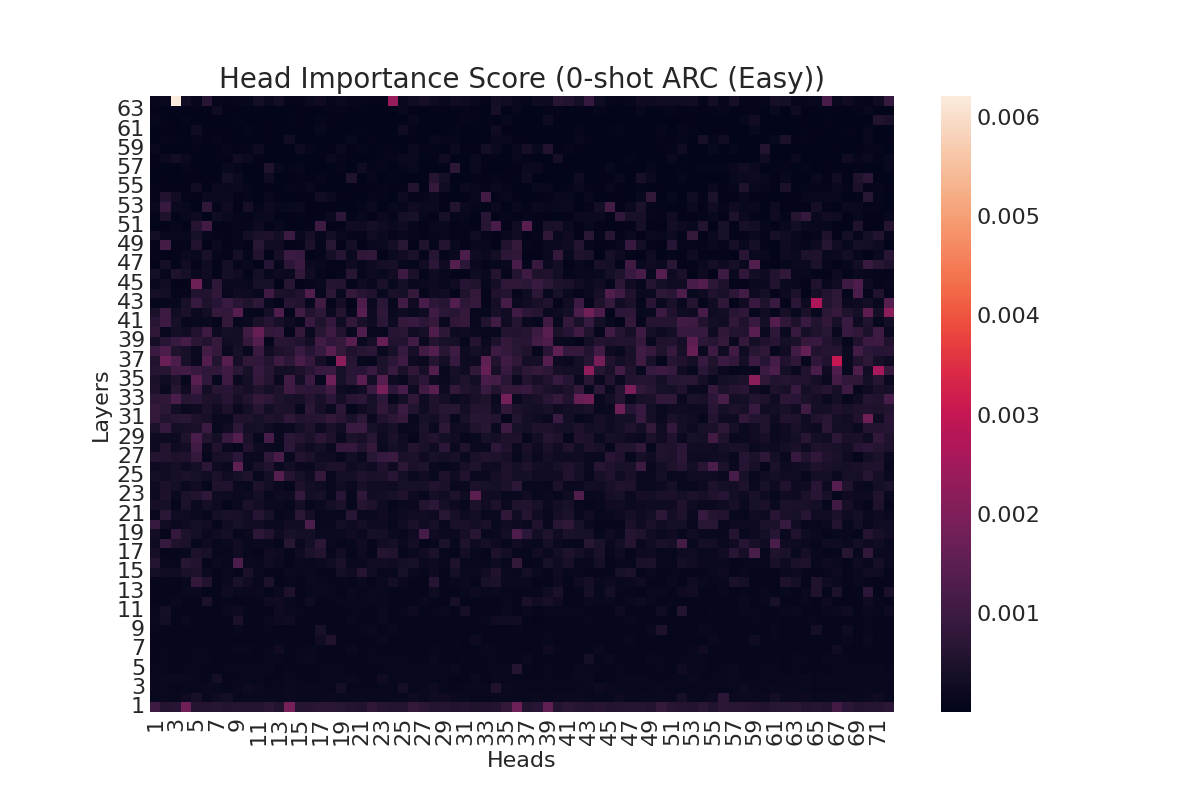}}}
    \subfloat[\centering ARC (Challenge)]{{\includegraphics[scale=0.2]{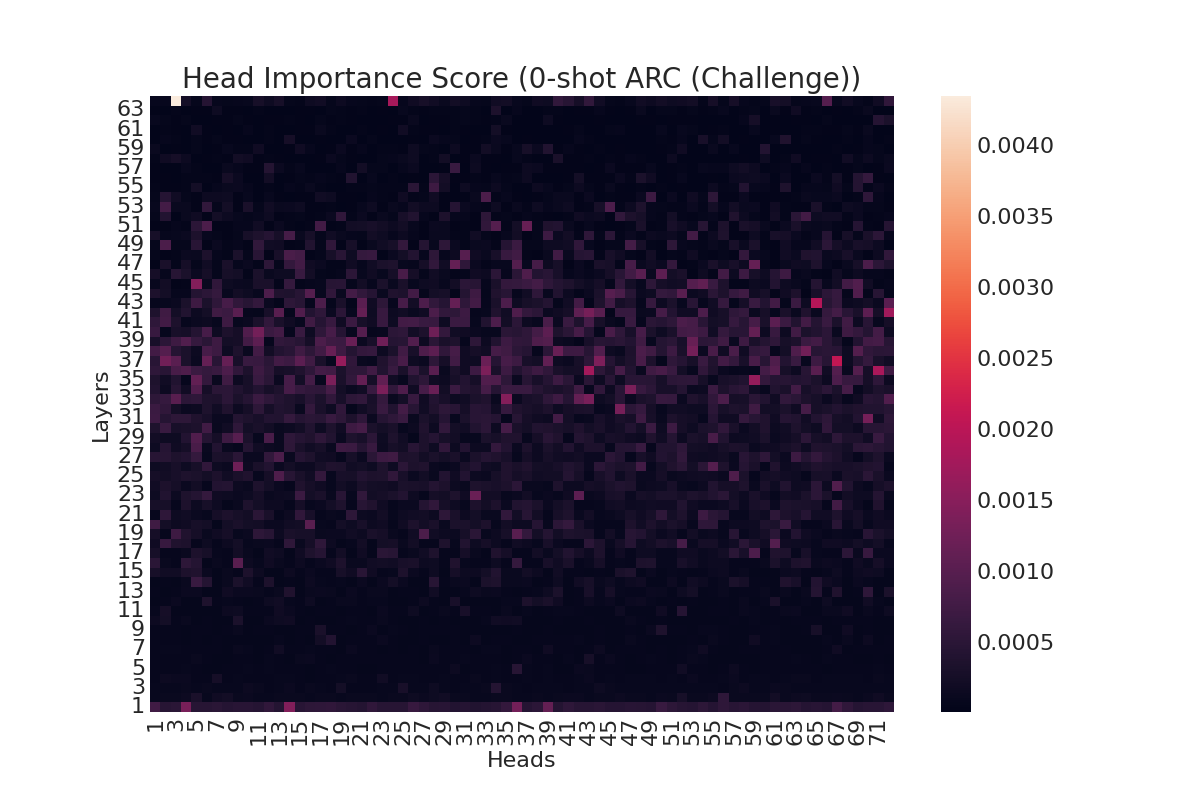}}}
    \qquad
    
     \subfloat[\centering CB]{{\includegraphics[scale=0.2]{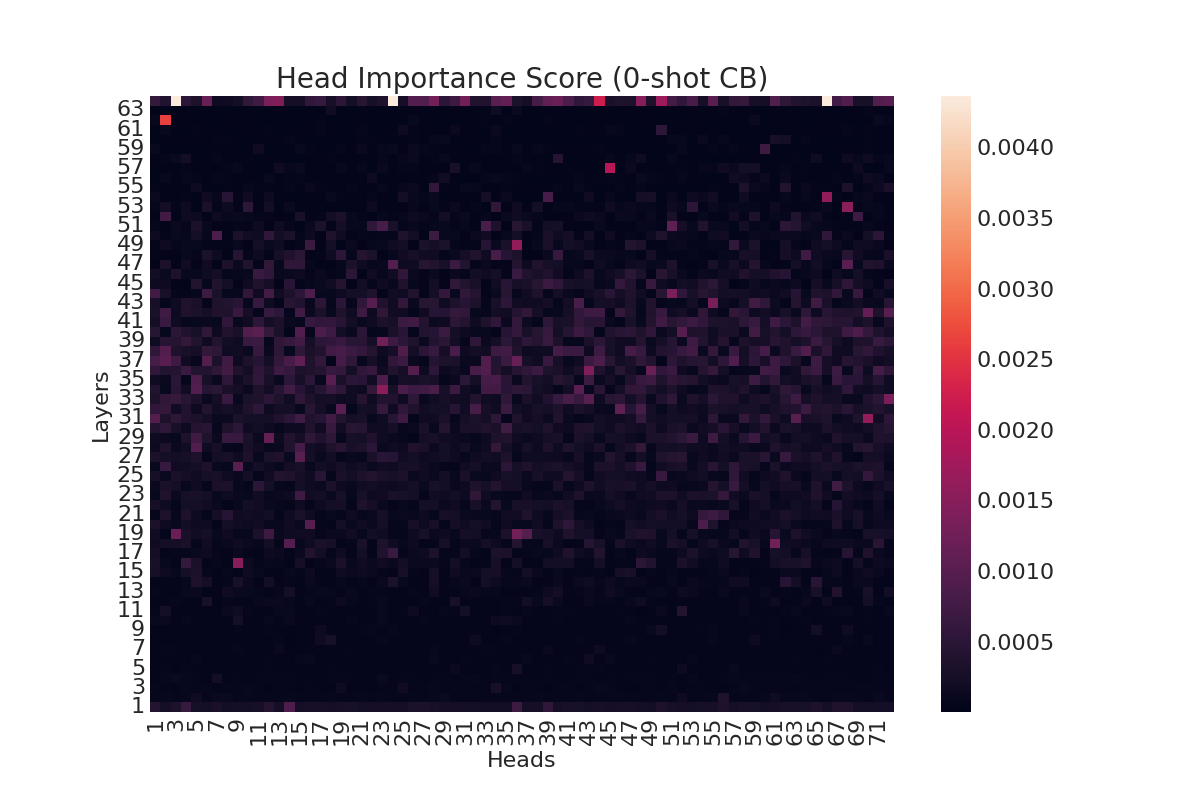}}}
     \subfloat[\centering BoolQ]{{\includegraphics[scale=0.2]{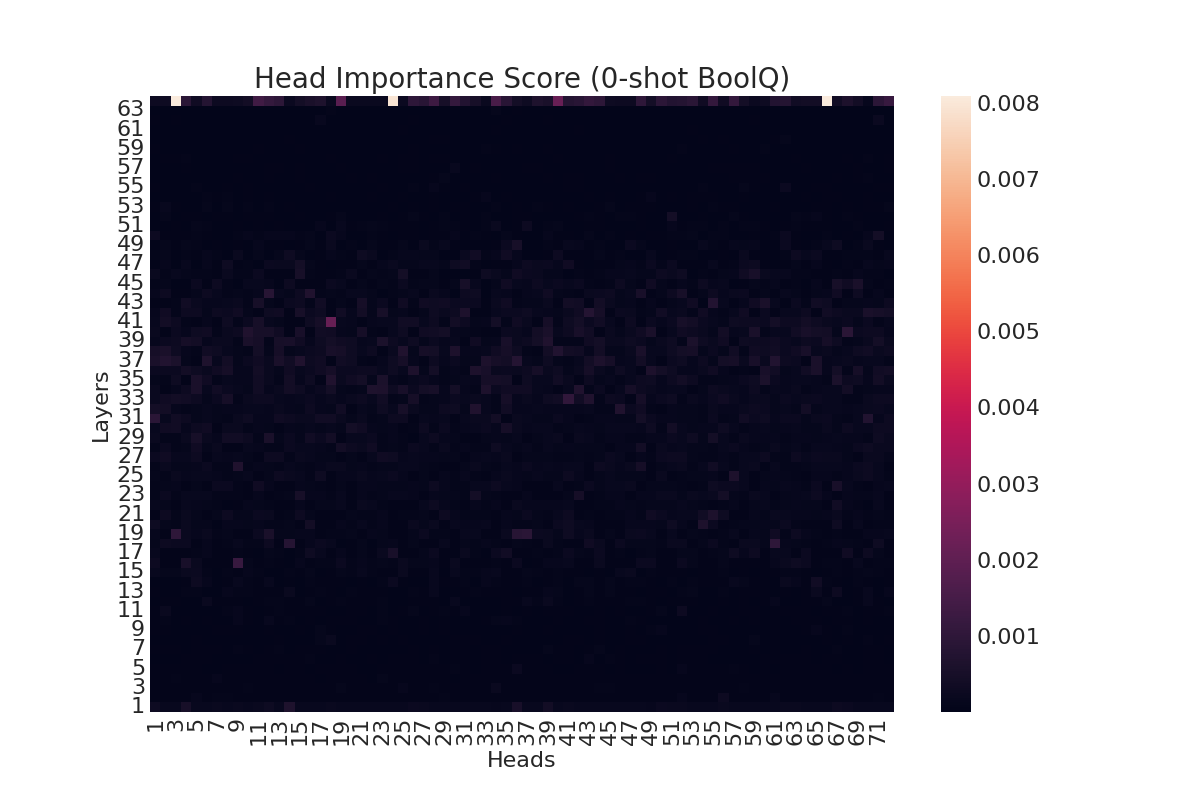}}}
     \subfloat[\centering Winogrande]{{\includegraphics[scale=0.2]{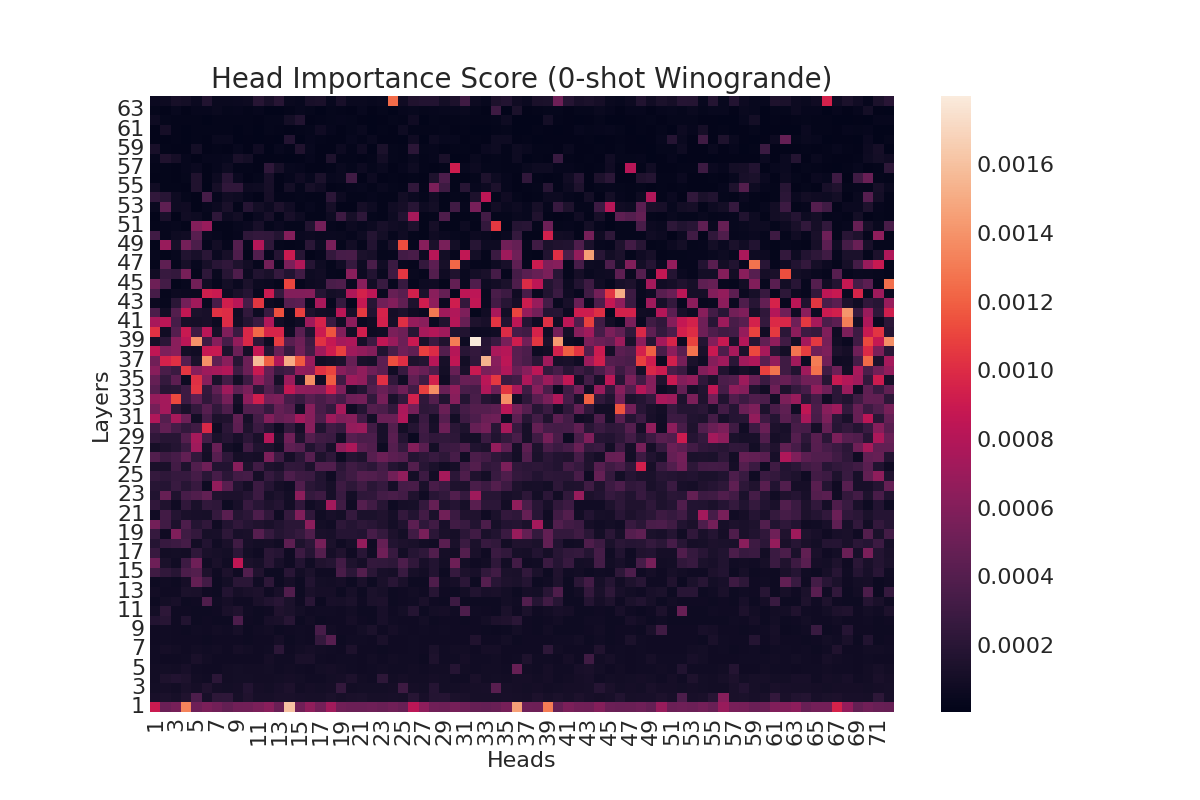}}}
     \qquad
     
     \subfloat[\centering RTE]{{\includegraphics[scale=0.2]{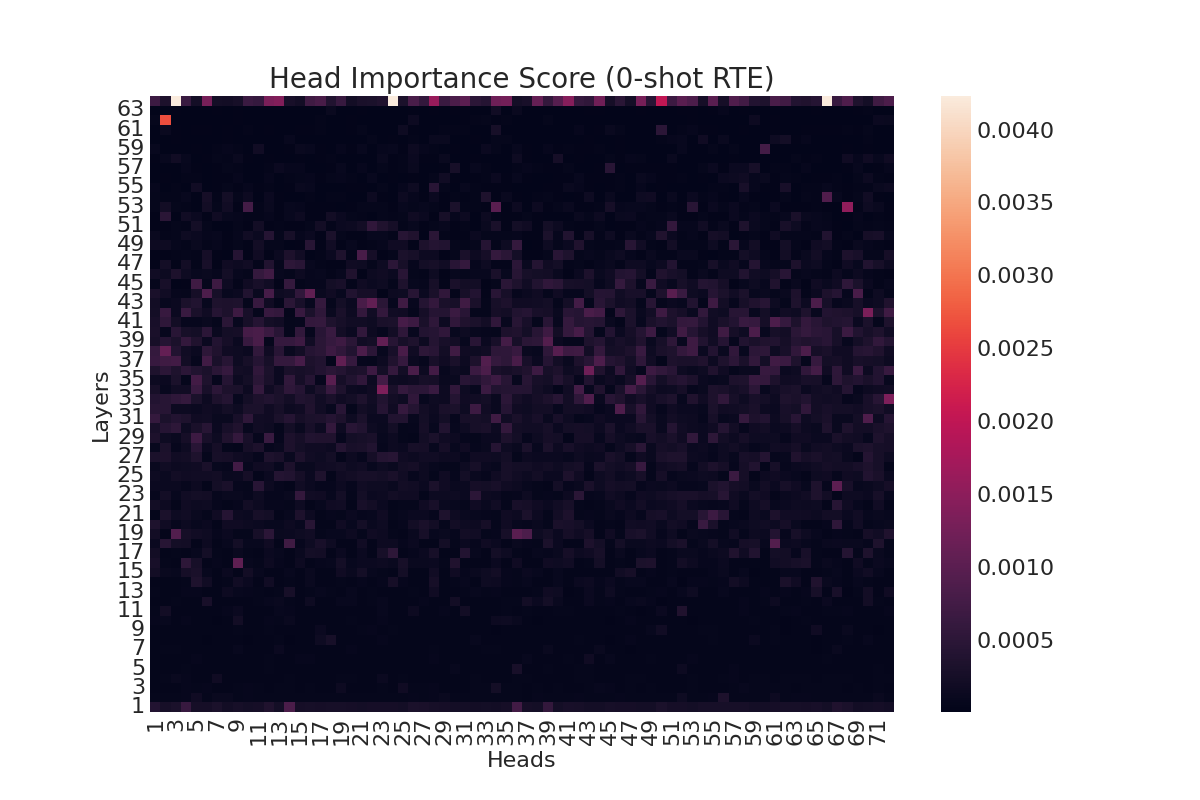}}}
     \subfloat[\centering MultiRC]{{\includegraphics[scale=0.2]{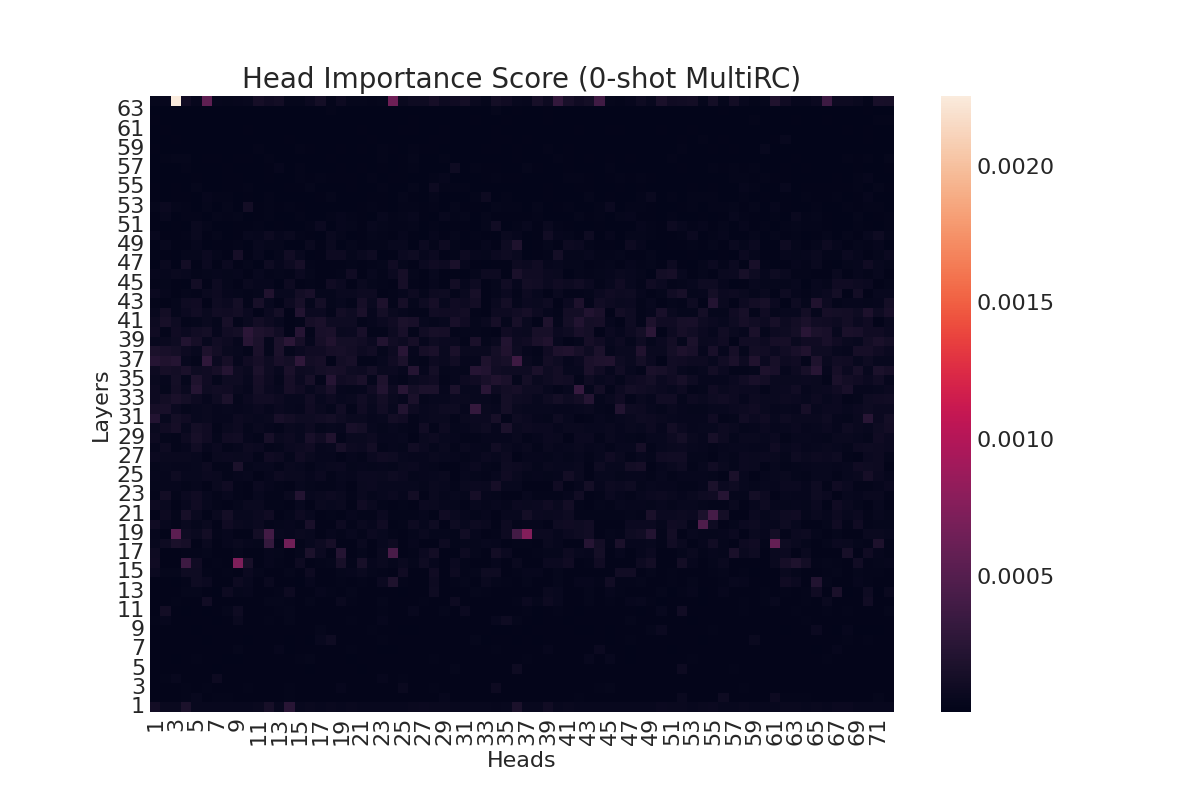}}}
     \subfloat[\centering OpenBookQA]{{\includegraphics[scale=0.2]{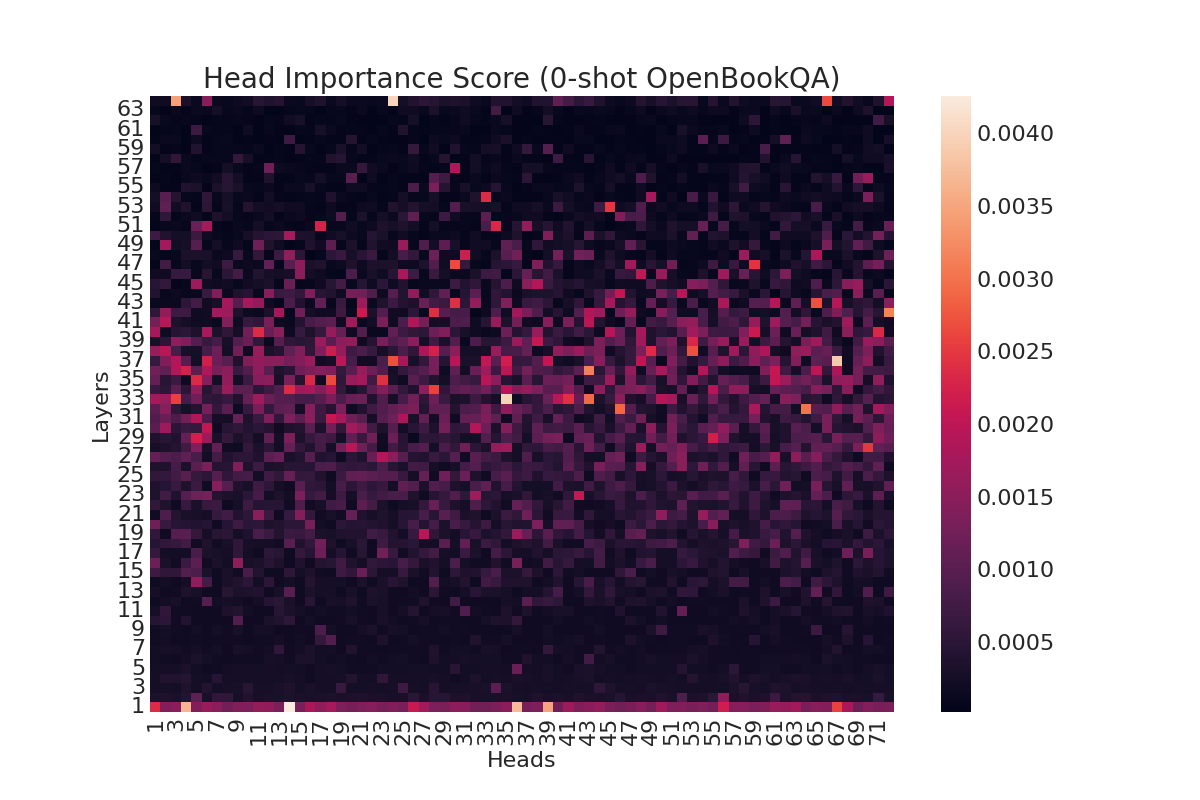}}}
     \qquad
     
     \subfloat[\centering COPA]{{\includegraphics[scale=0.2]{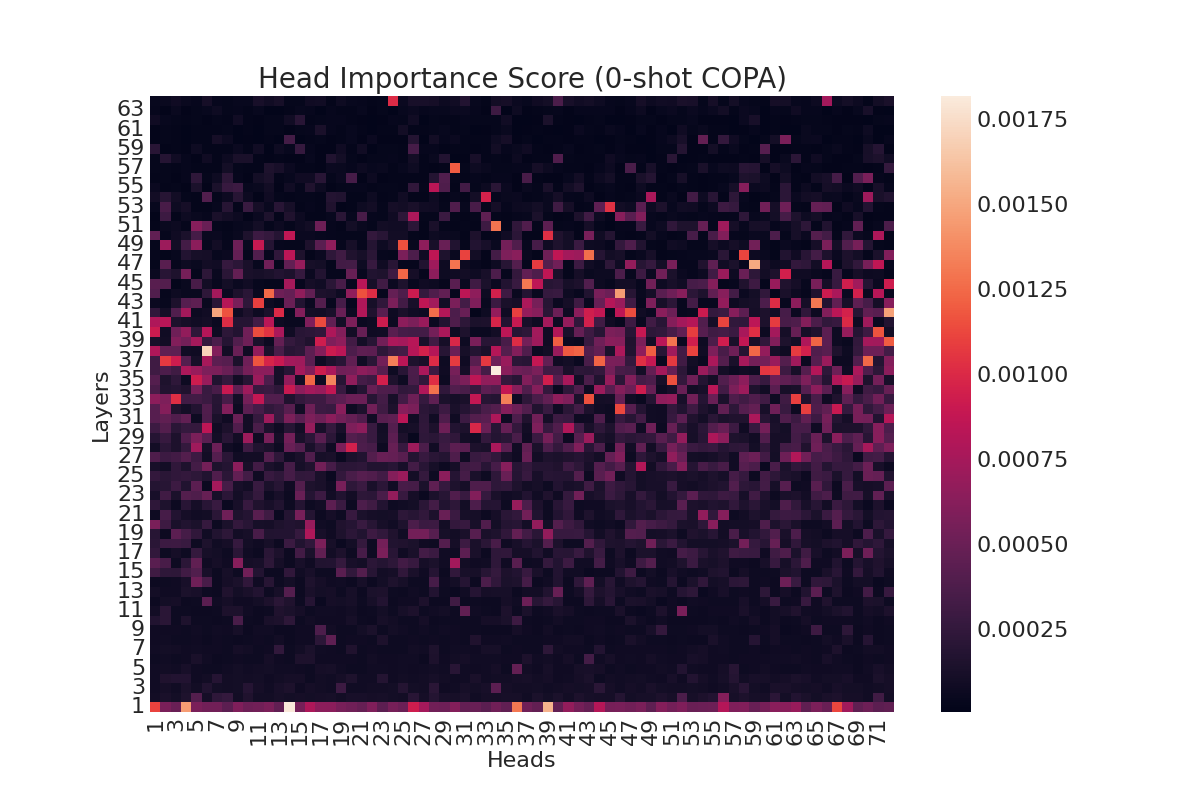}}}
     \subfloat[\centering PIQA]{{\includegraphics[scale=0.2]{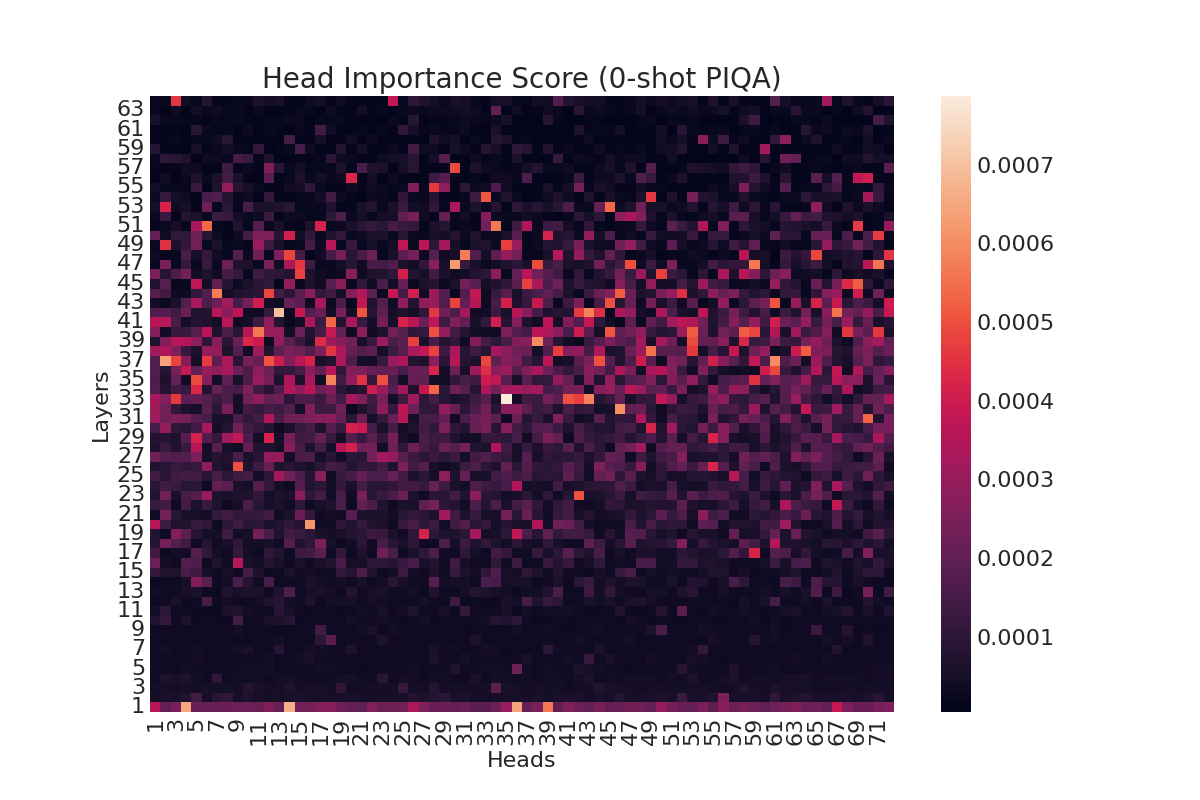}}}
     \subfloat[\centering ReCoRD]{{\includegraphics[scale=0.2]{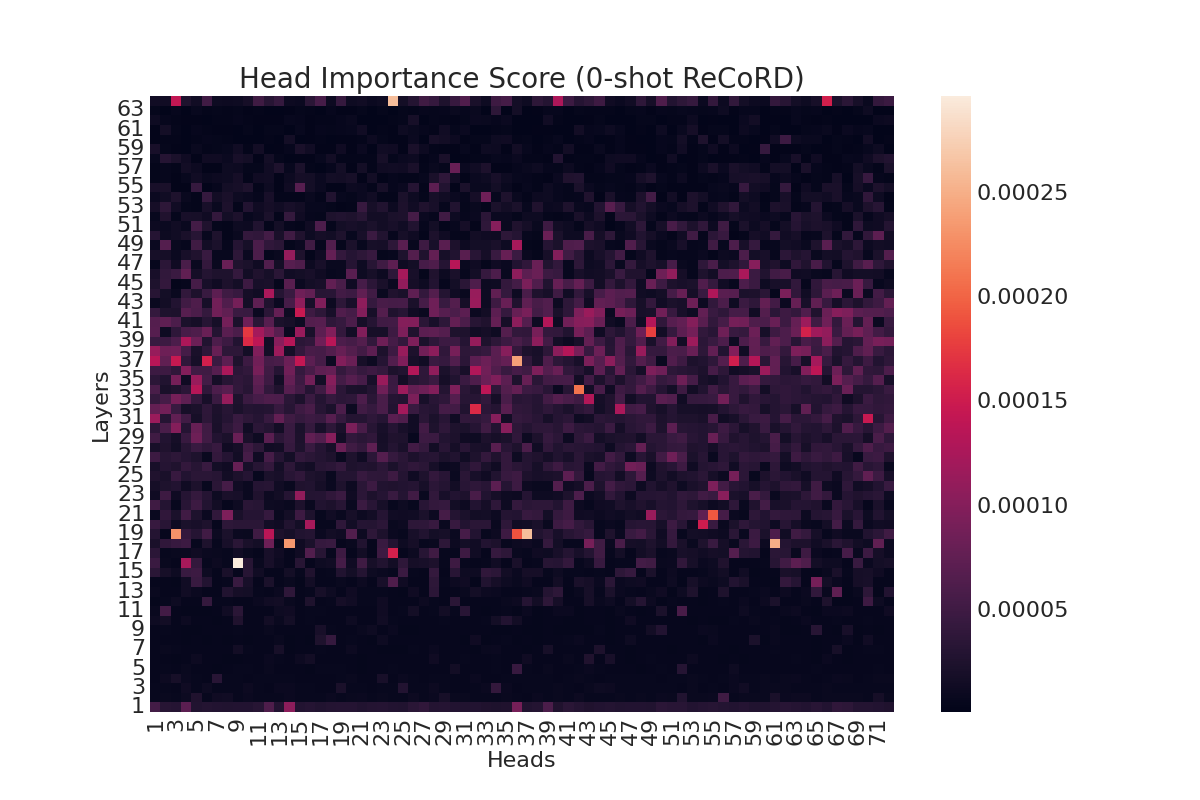}}}
     \qquad
     
     \subfloat[\centering WIC]{{\includegraphics[scale=0.2]{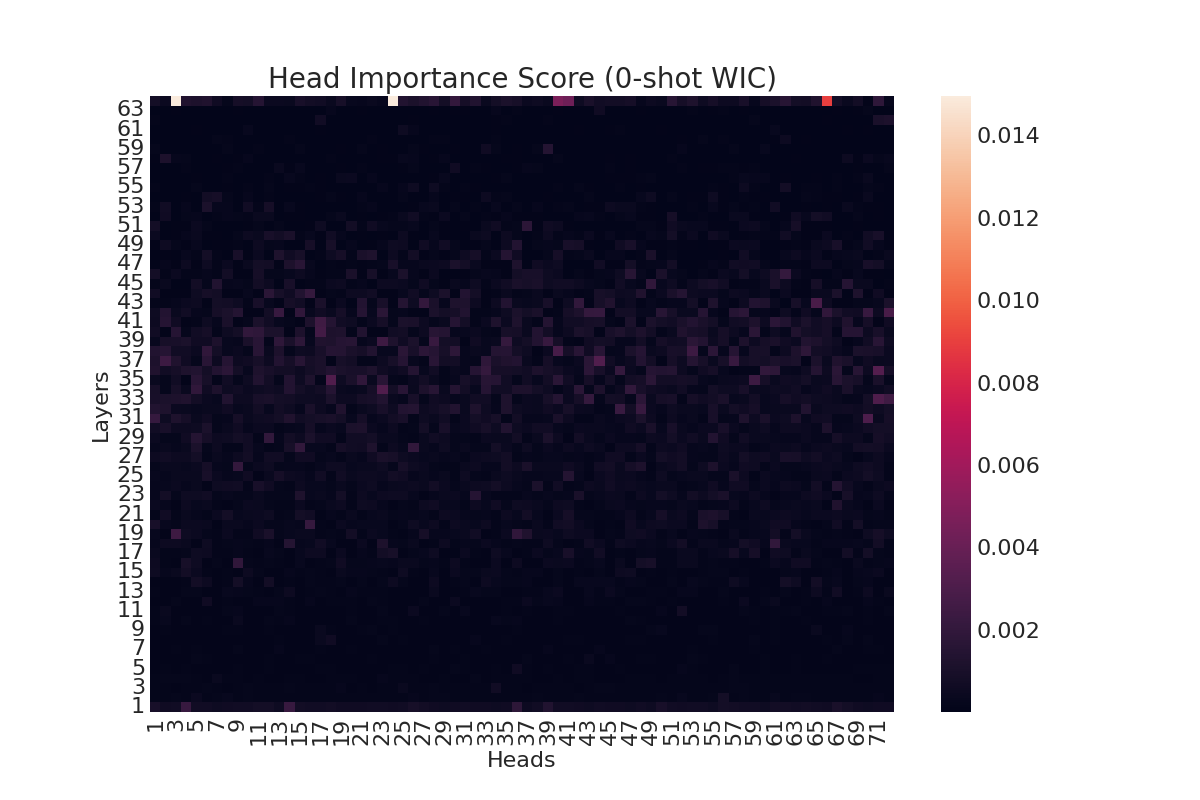}}}
     \subfloat[\centering WSC]{{\includegraphics[scale=0.2]{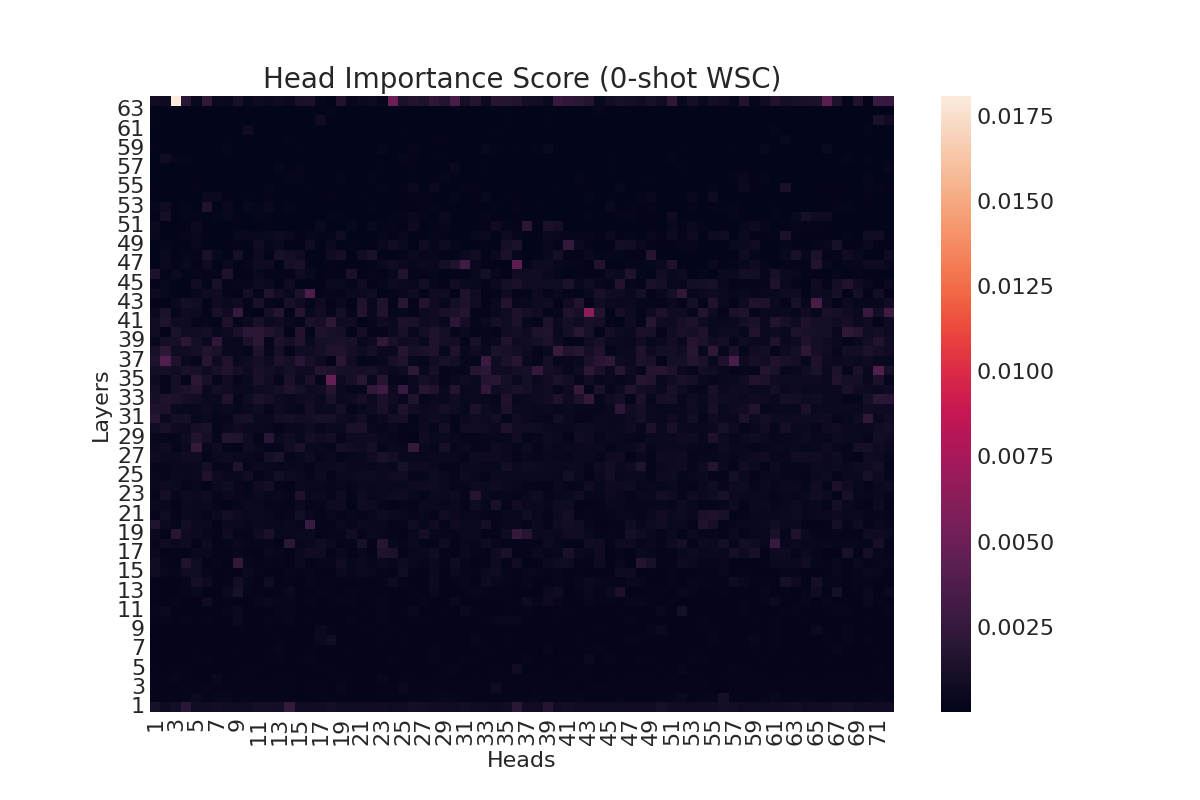}}}
     
    \caption{Attention head importance score heatmaps for zero-shot in-context learning with OPT-66B for each task.}
    \label{app_fig:heatmap-0shot}
\end{figure*}

\begin{figure*}[h]
    \centering
    
    \subfloat[\centering HellaSwag]{{\includegraphics[scale=0.2]{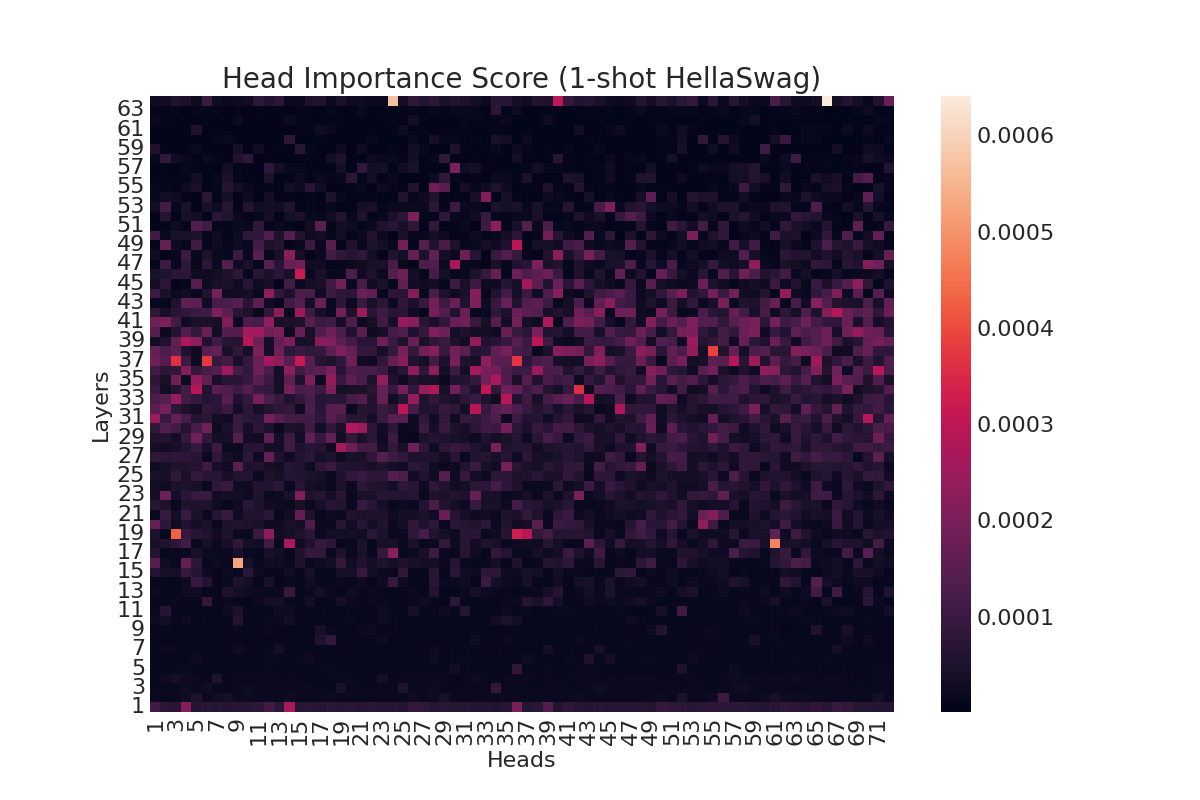}}}
    \subfloat[\centering ARC (Easy)]{{\includegraphics[scale=0.2]{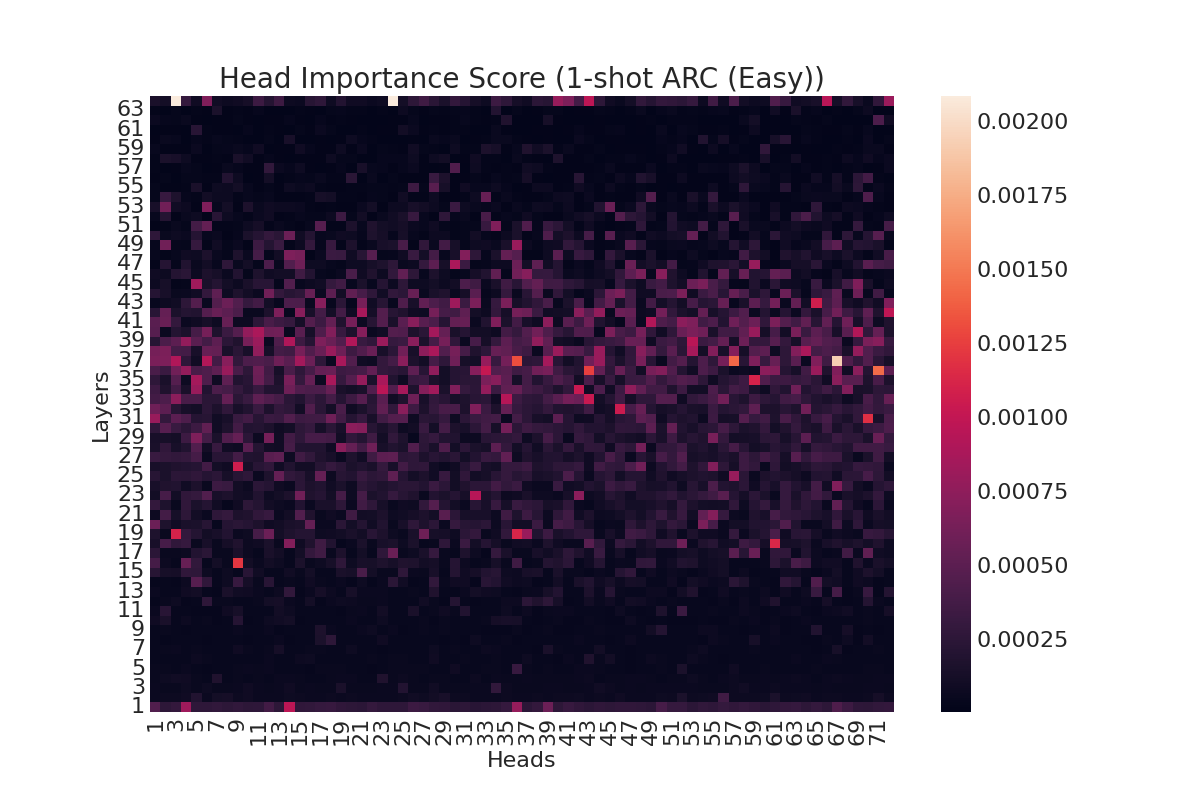}}}
    \subfloat[\centering ARC (Challenge)]{{\includegraphics[scale=0.2]{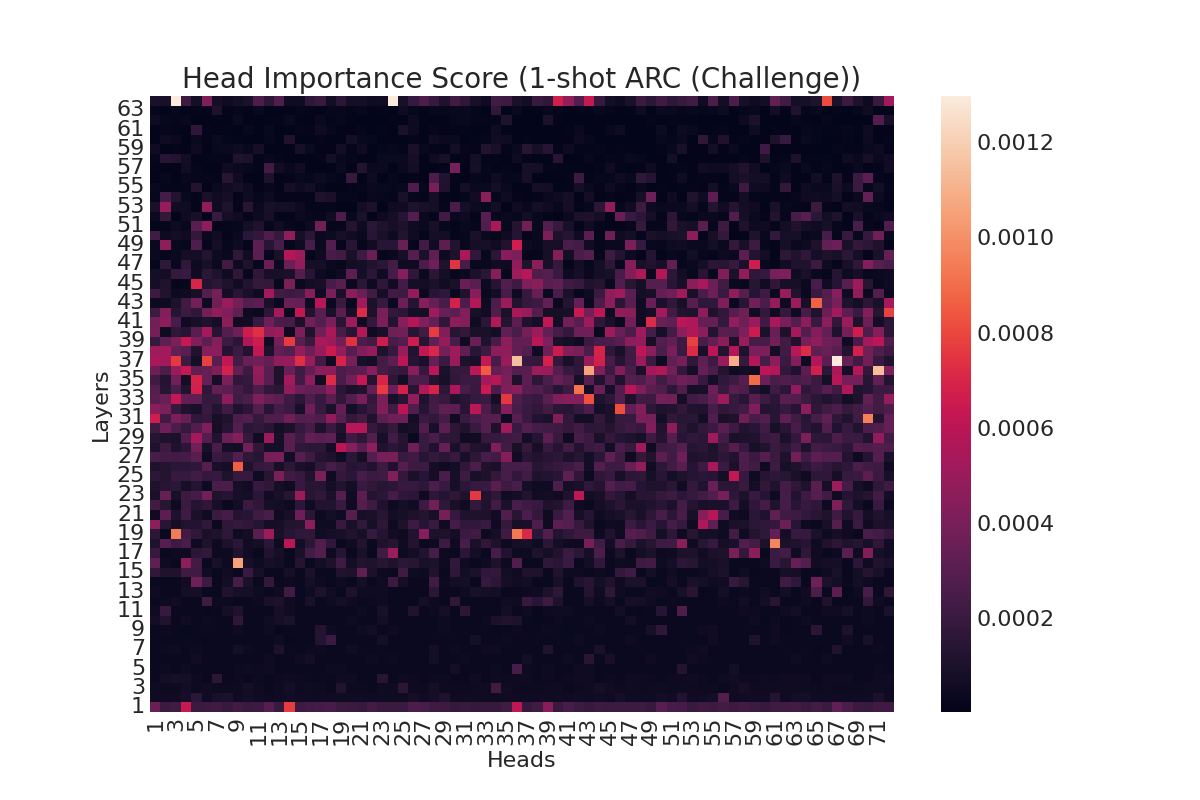}}}
    \qquad
    
     \subfloat[\centering CB]{{\includegraphics[scale=0.2]{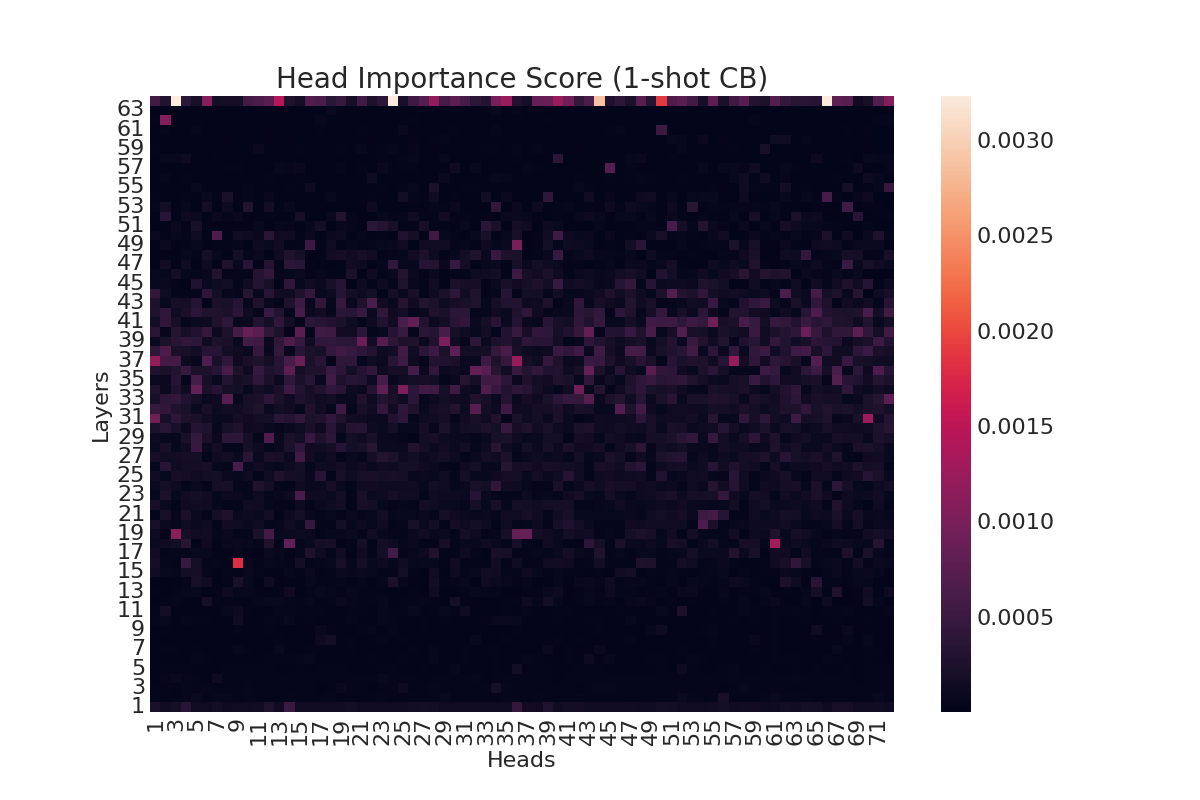}}}
     \subfloat[\centering BoolQ]{{\includegraphics[scale=0.2]{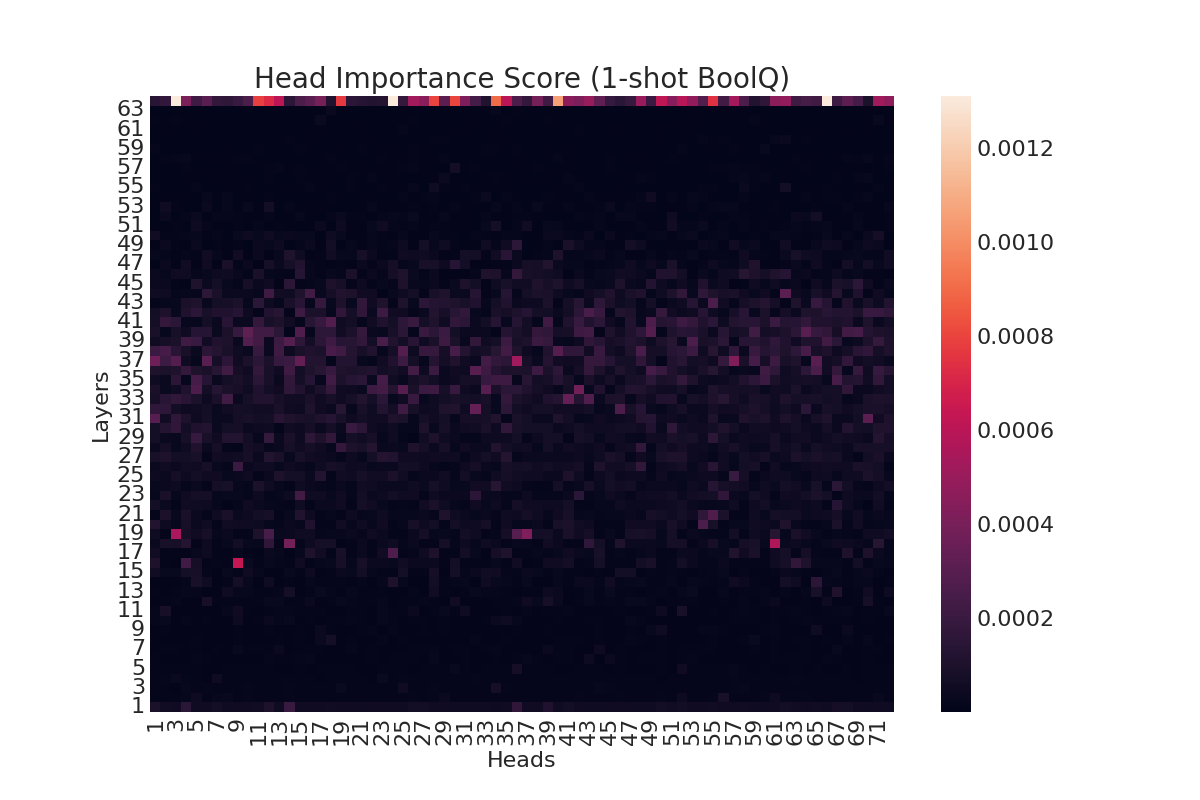}}}
     \subfloat[\centering Winogrande]{{\includegraphics[scale=0.2]{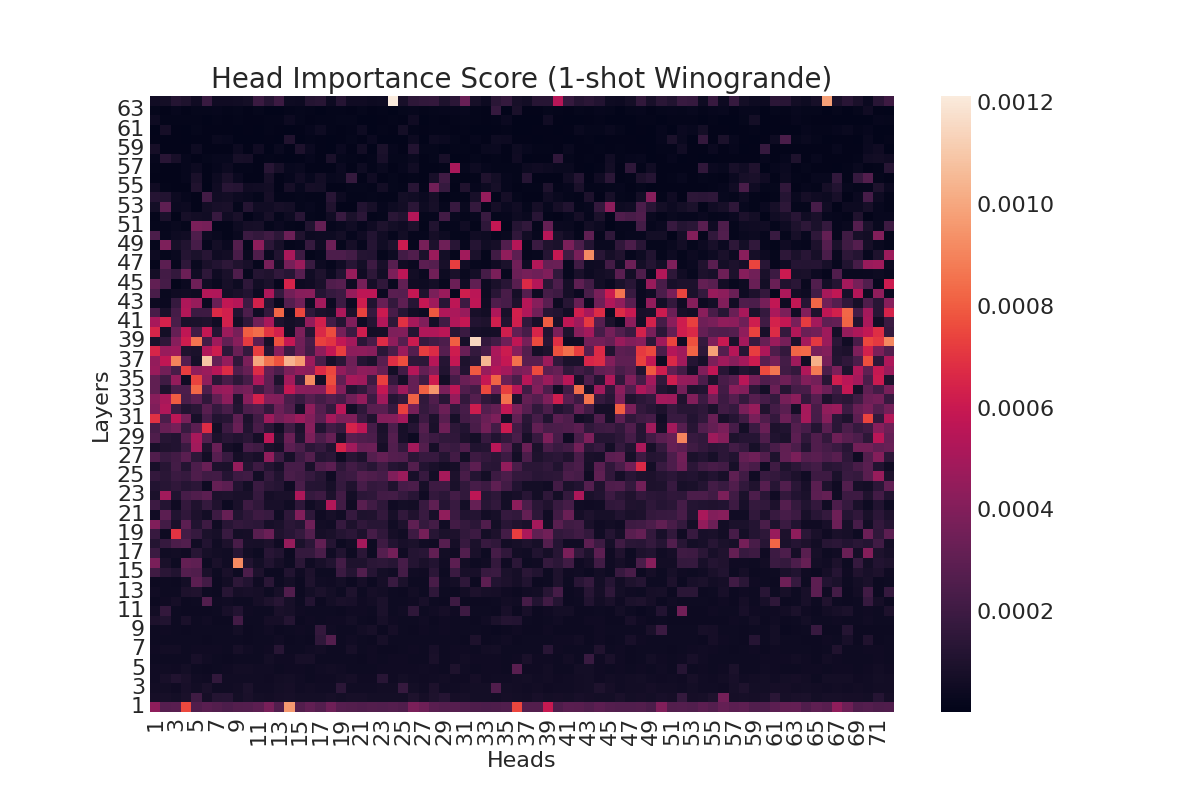}}}
     \qquad
     
     \subfloat[\centering RTE]{{\includegraphics[scale=0.2]{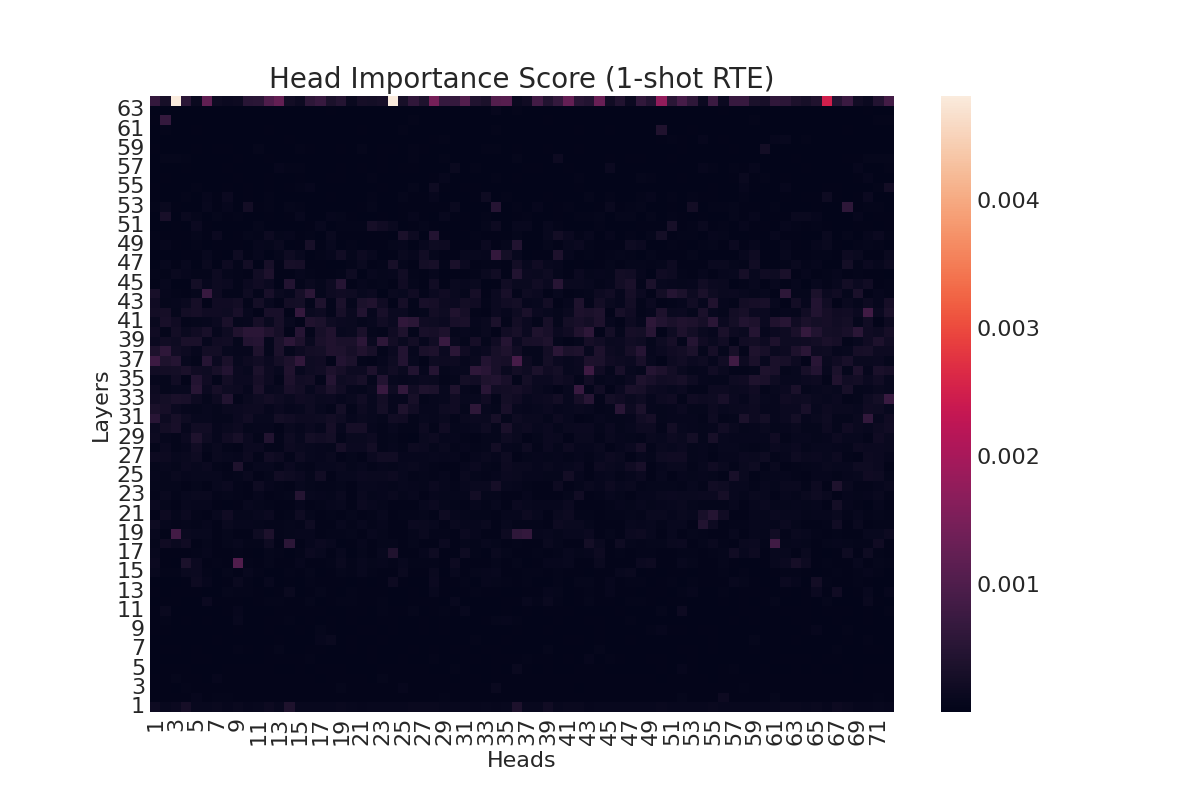}}}
     \subfloat[\centering MultiRC]{{\includegraphics[scale=0.2]{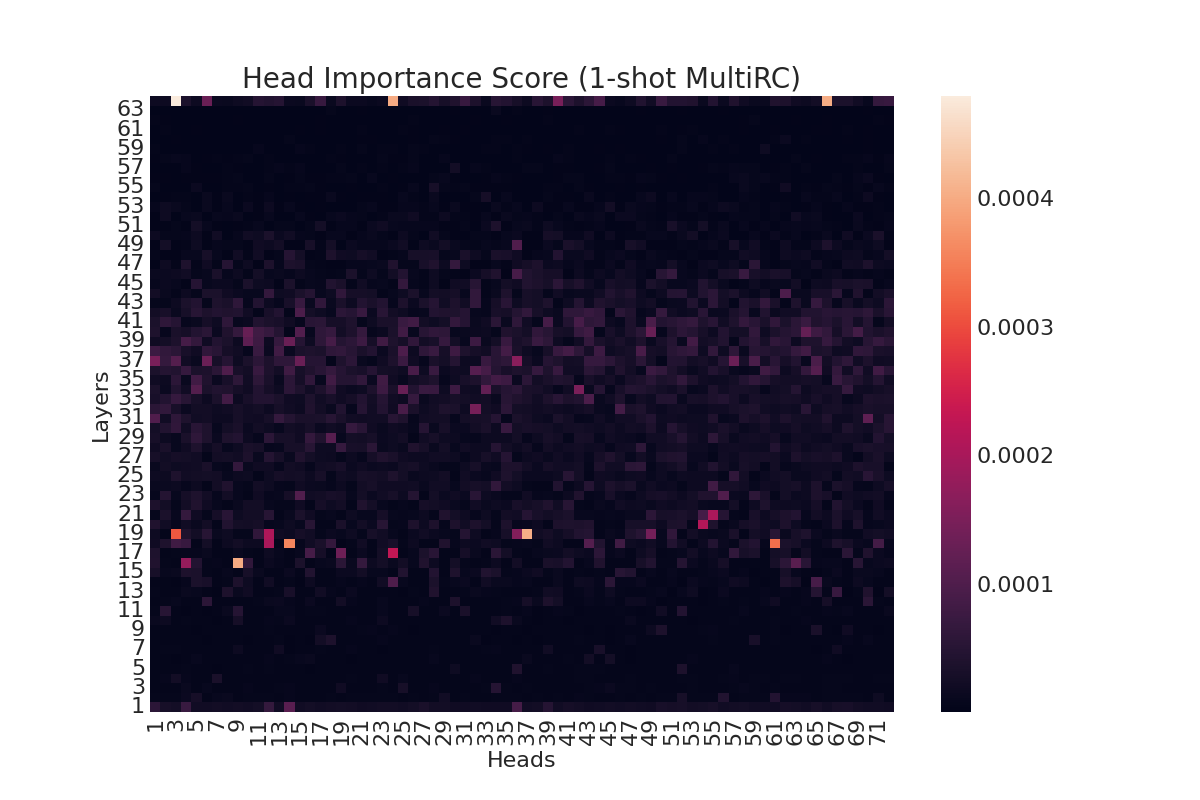}}}
     \subfloat[\centering OpenBookQA]{{\includegraphics[scale=0.2]{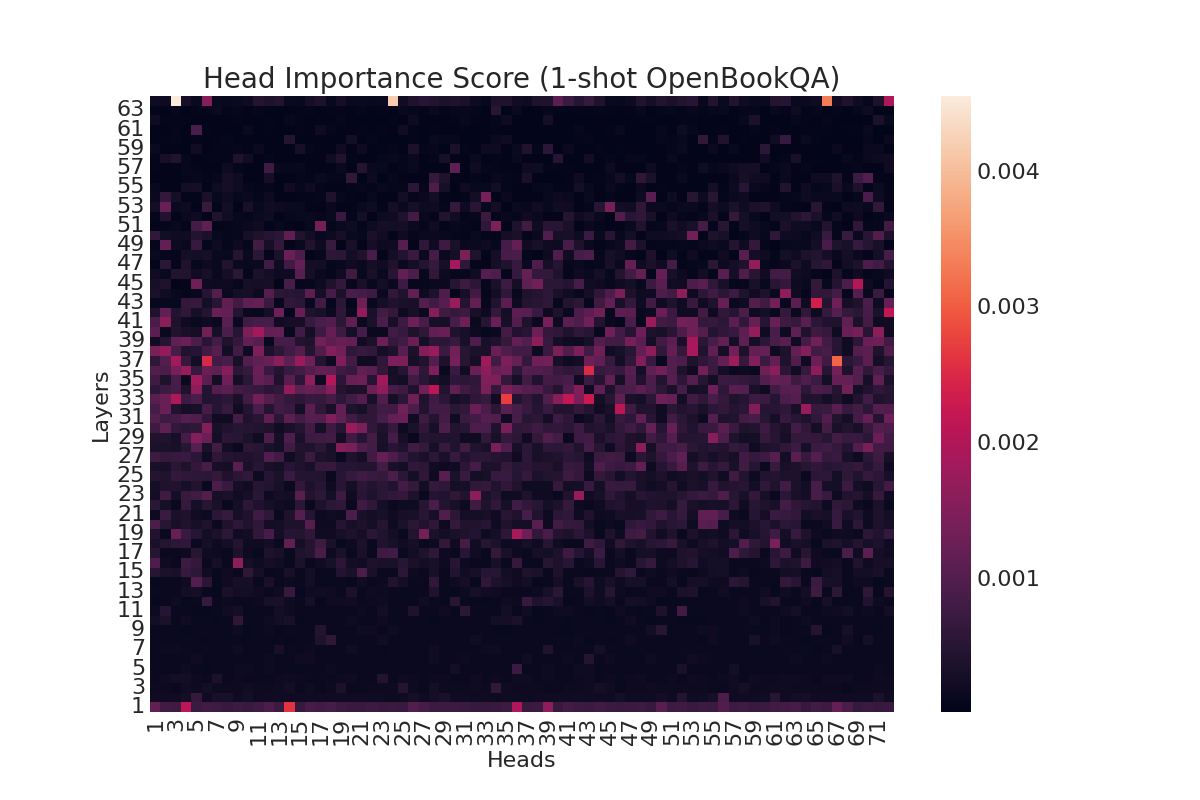}}}
     \qquad
     
     \subfloat[\centering COPA]{{\includegraphics[scale=0.2]{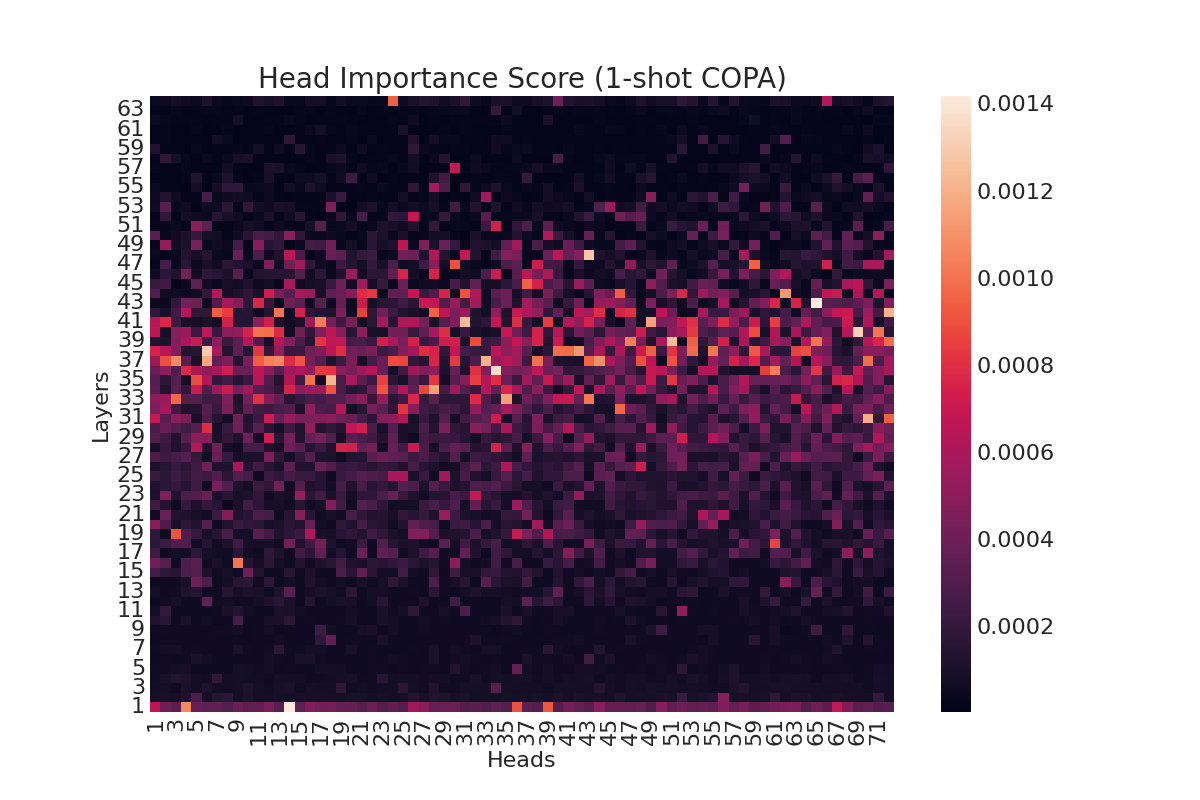}}}
     \subfloat[\centering PIQA]{{\includegraphics[scale=0.2]{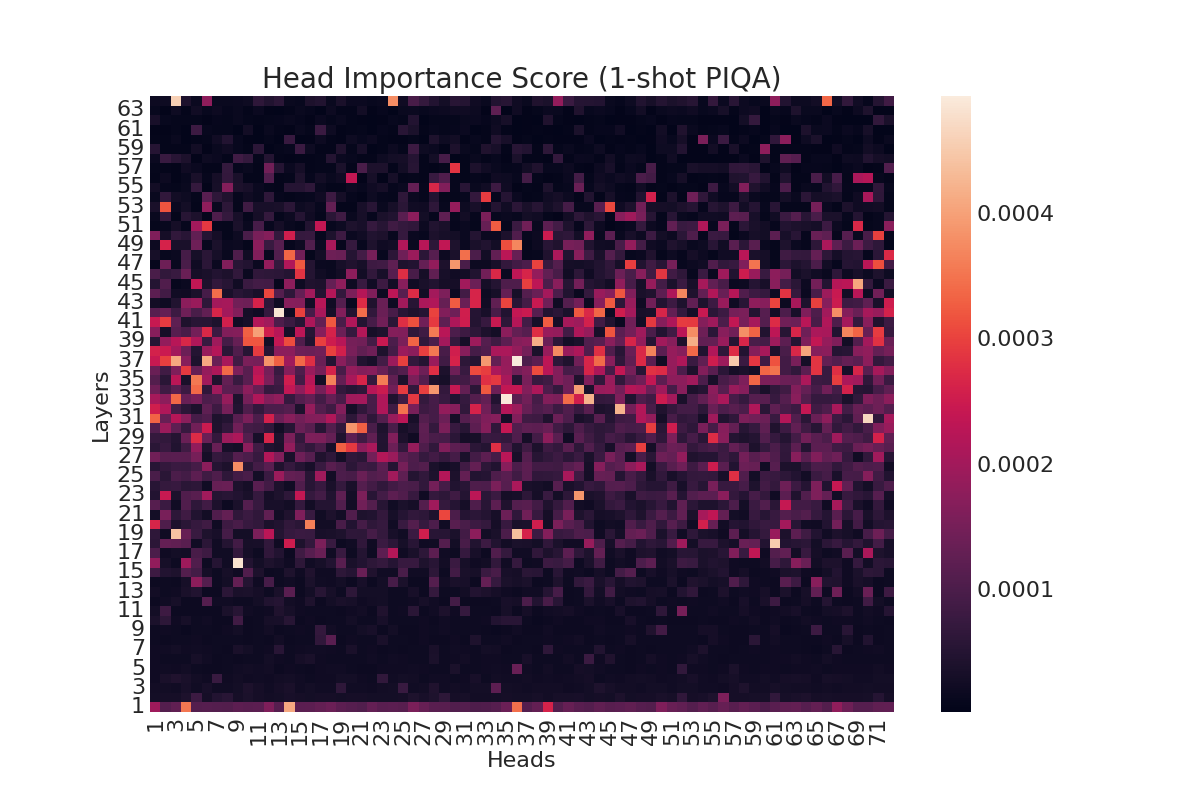}}}
     \subfloat[\centering ReCoRD]{{\includegraphics[scale=0.2]{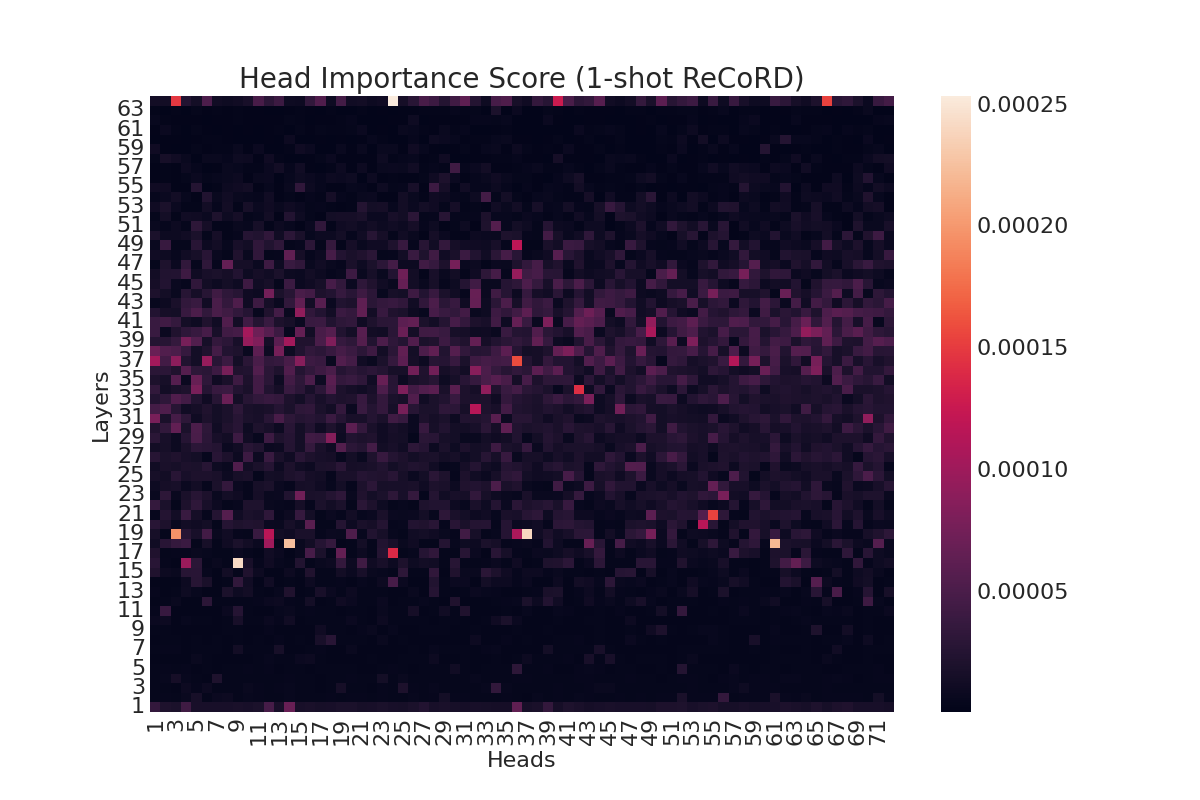}}}
     \qquad
     
     \subfloat[\centering WIC]{{\includegraphics[scale=0.2]{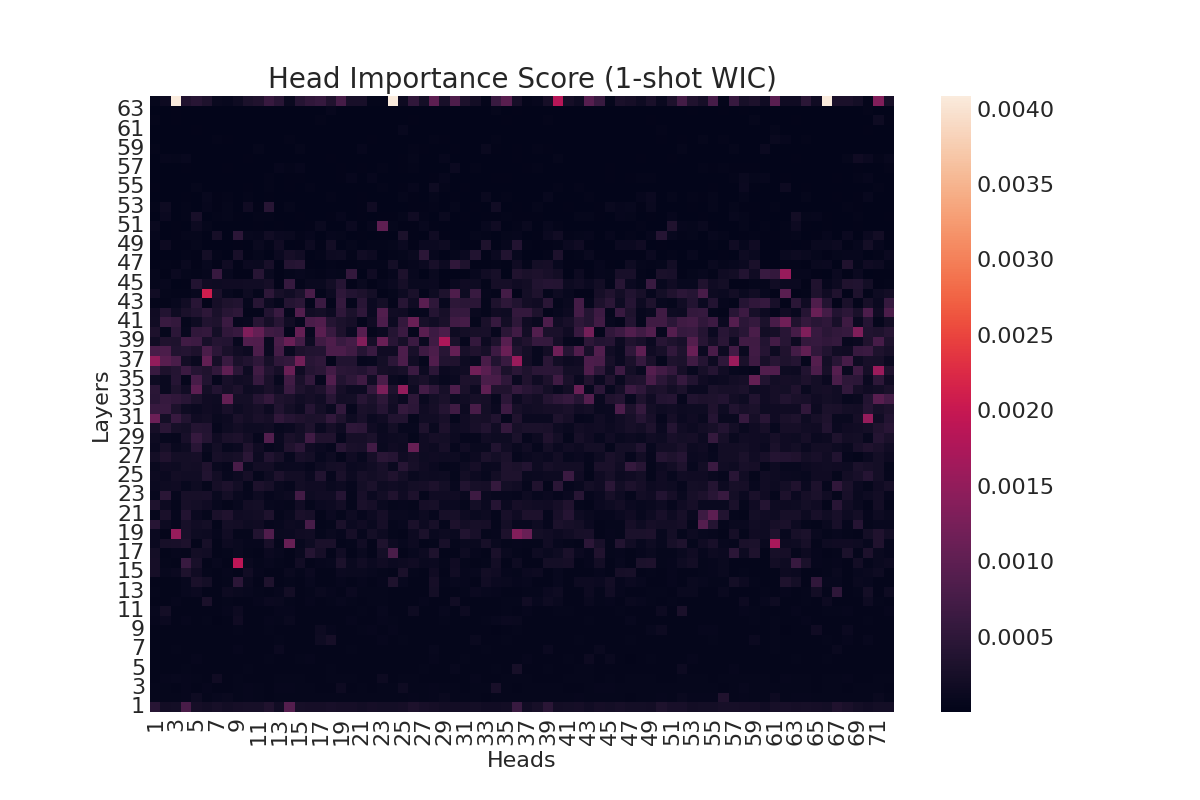}}}
     \subfloat[\centering WSC]{{\includegraphics[scale=0.2]{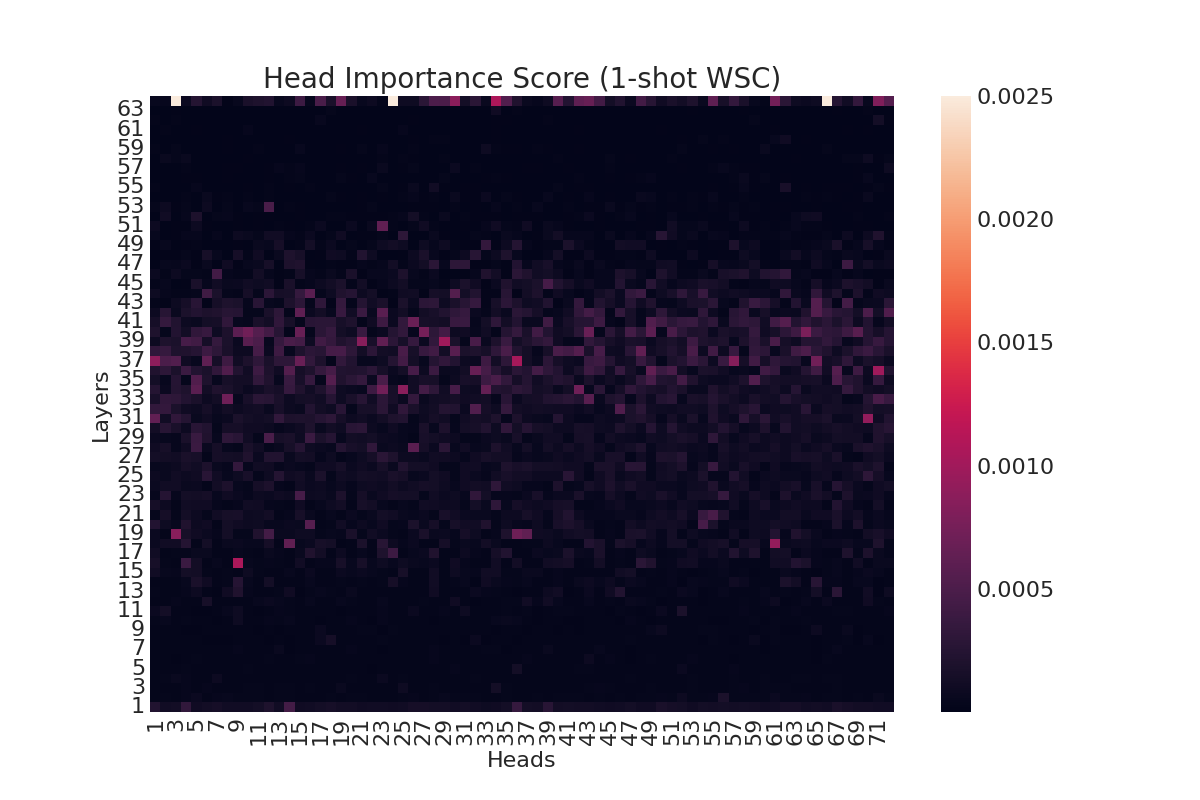}}}
     
    \caption{Attention head importance score heatmaps for one-shot in-context learning with OPT-66B for each task.}
    \label{app_fig:heatmap-1shot}
\end{figure*}

\begin{figure*}[h]
    \centering
    
    \subfloat[\centering HellaSwag]{{\includegraphics[scale=0.2]{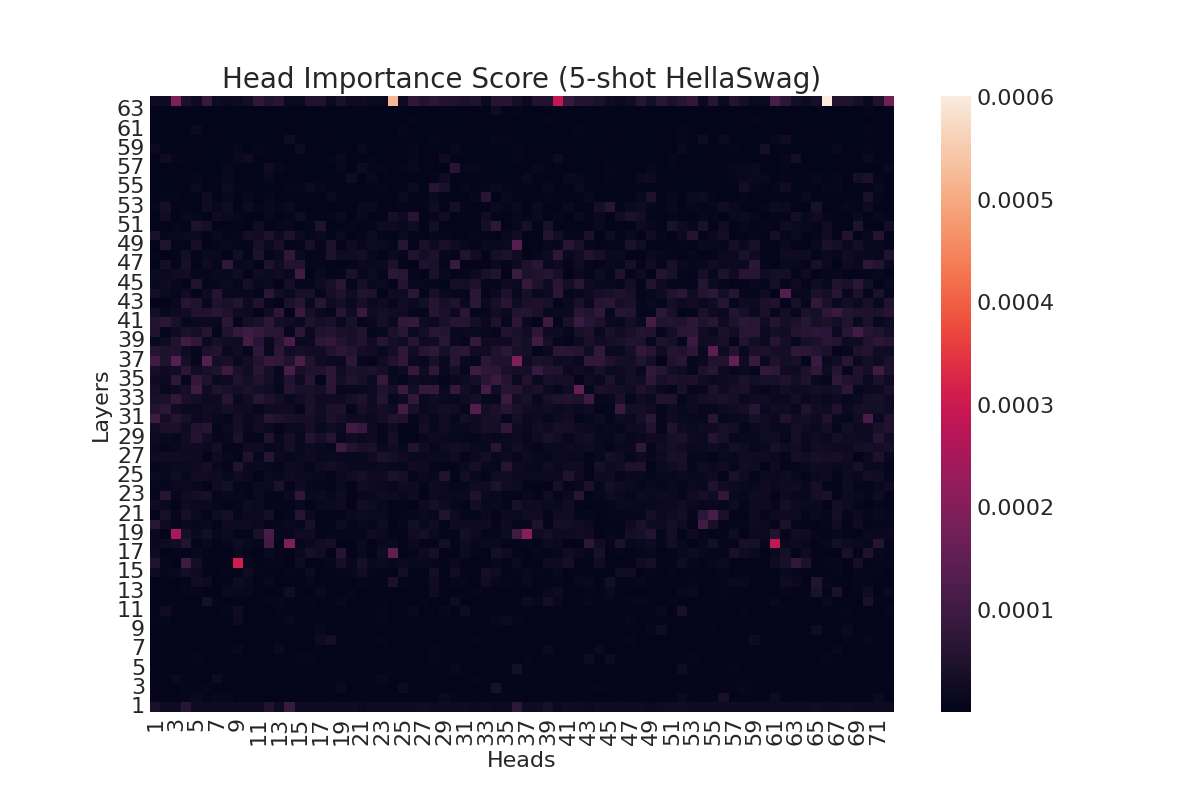}}}
    \subfloat[\centering ARC (Easy)]{{\includegraphics[scale=0.2]{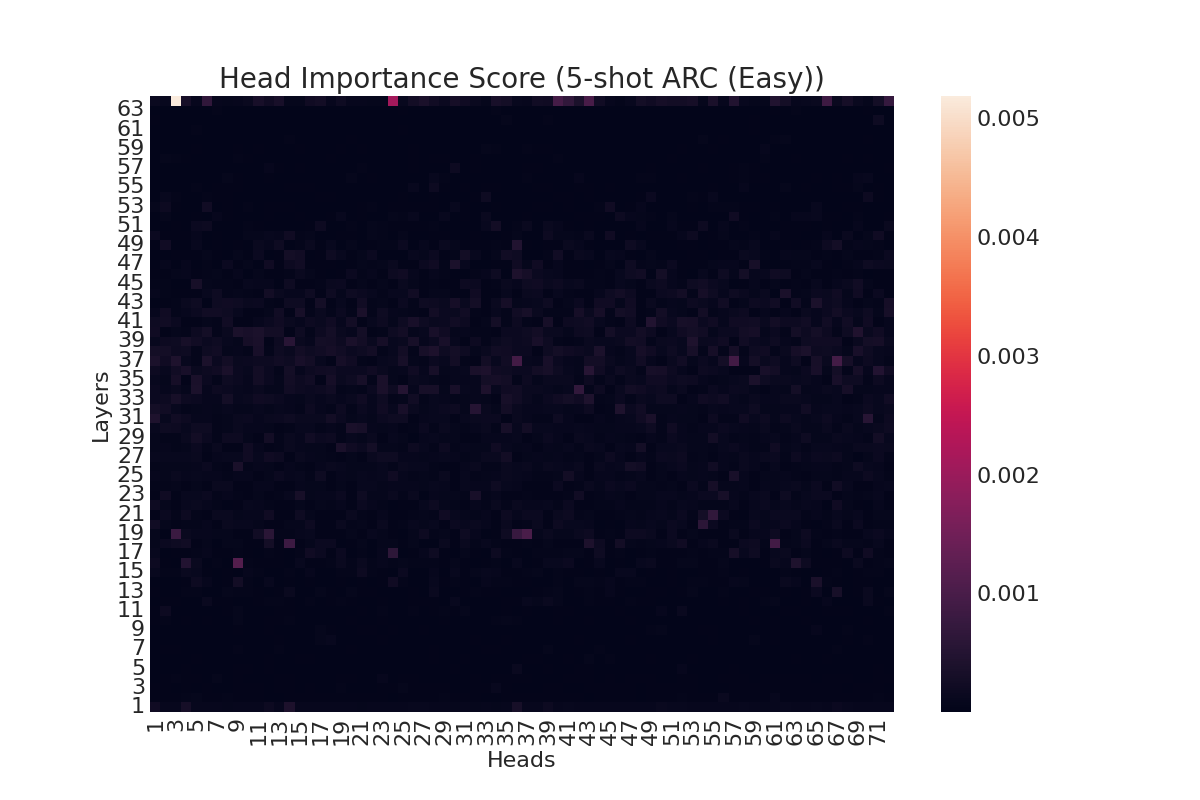}}}
    \subfloat[\centering ARC (Challenge)]{{\includegraphics[scale=0.2]{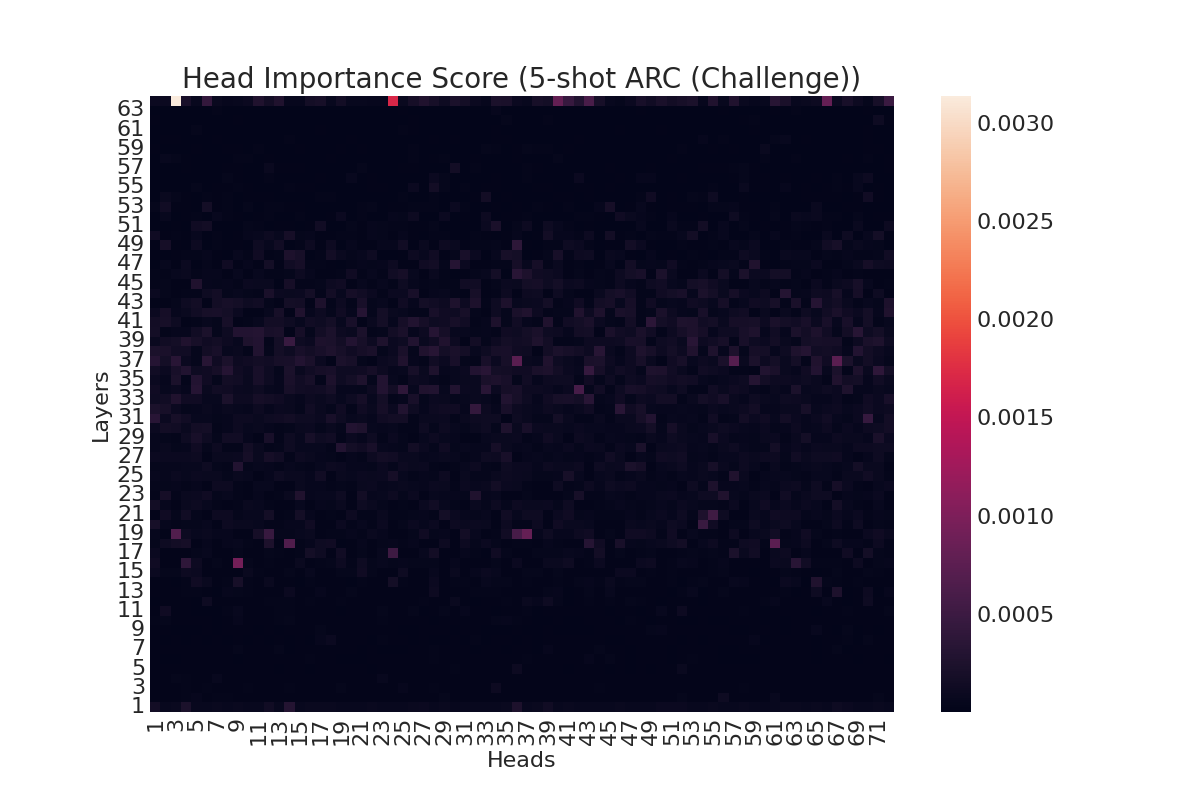}}}
    \qquad
    
     \subfloat[\centering CB]{{\includegraphics[scale=0.2]{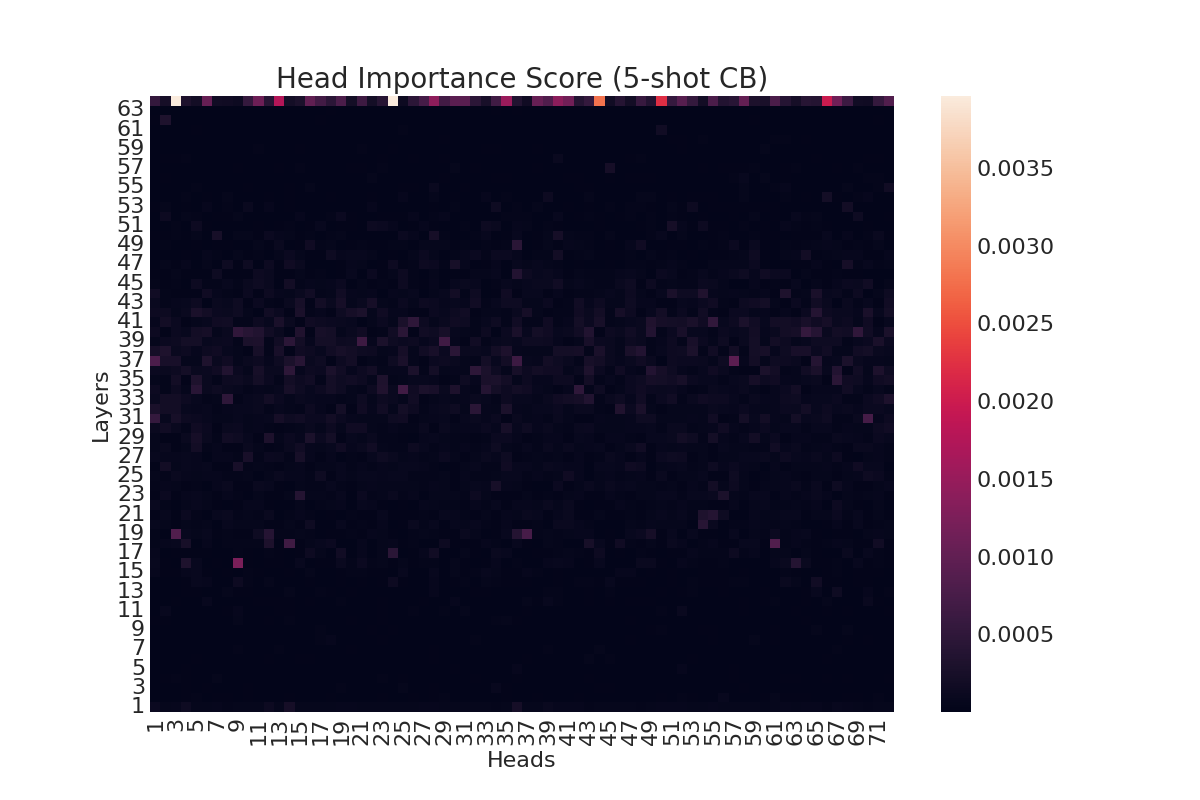}}}
     \subfloat[\centering BoolQ]{{\includegraphics[scale=0.2]{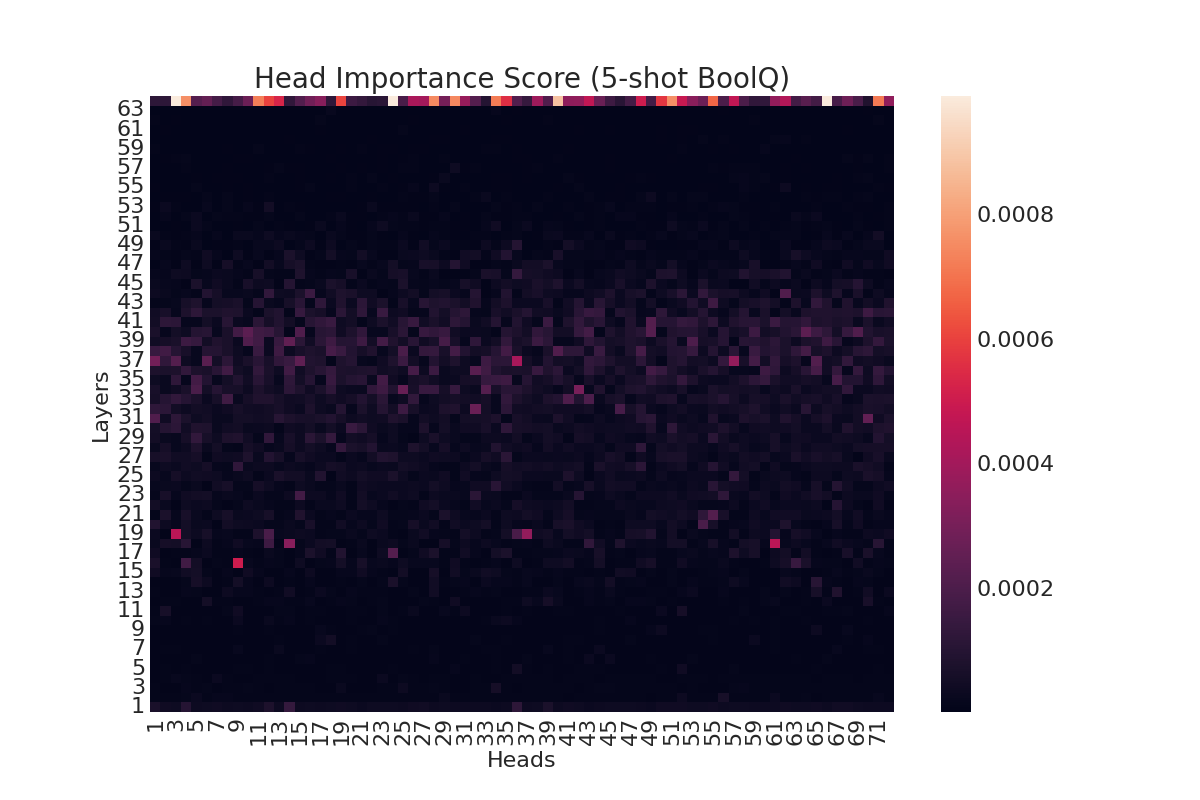}}}
     \subfloat[\centering Winogrande]{{\includegraphics[scale=0.2]{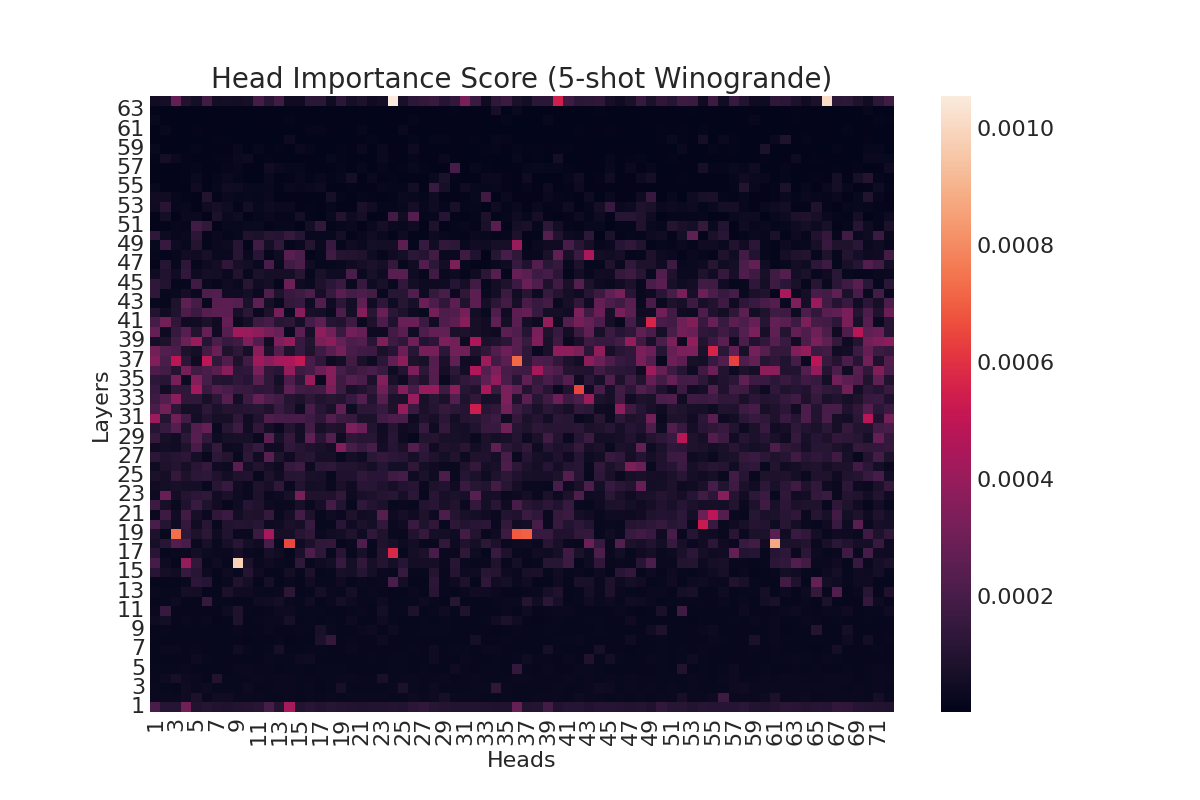}}}
     \qquad
     
     \subfloat[\centering RTE]{{\includegraphics[scale=0.2]{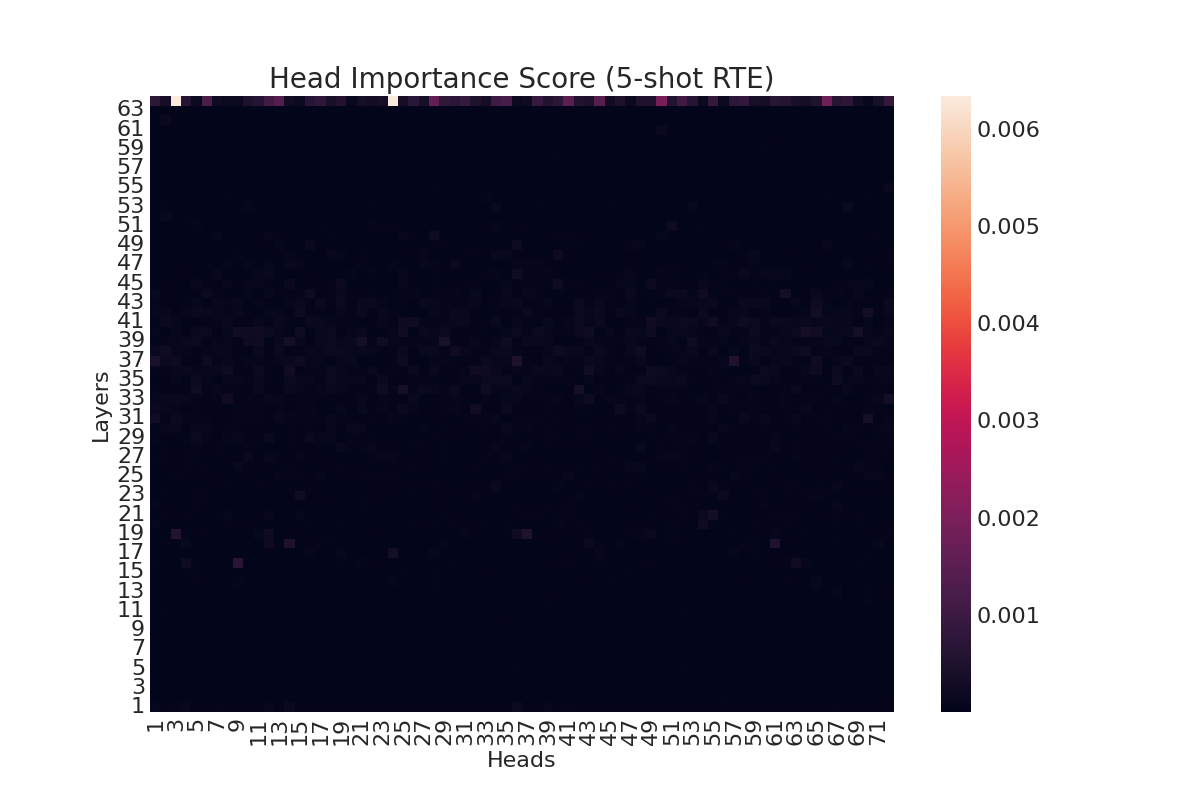}}}
     \subfloat[\centering MultiRC]{{\includegraphics[scale=0.2]{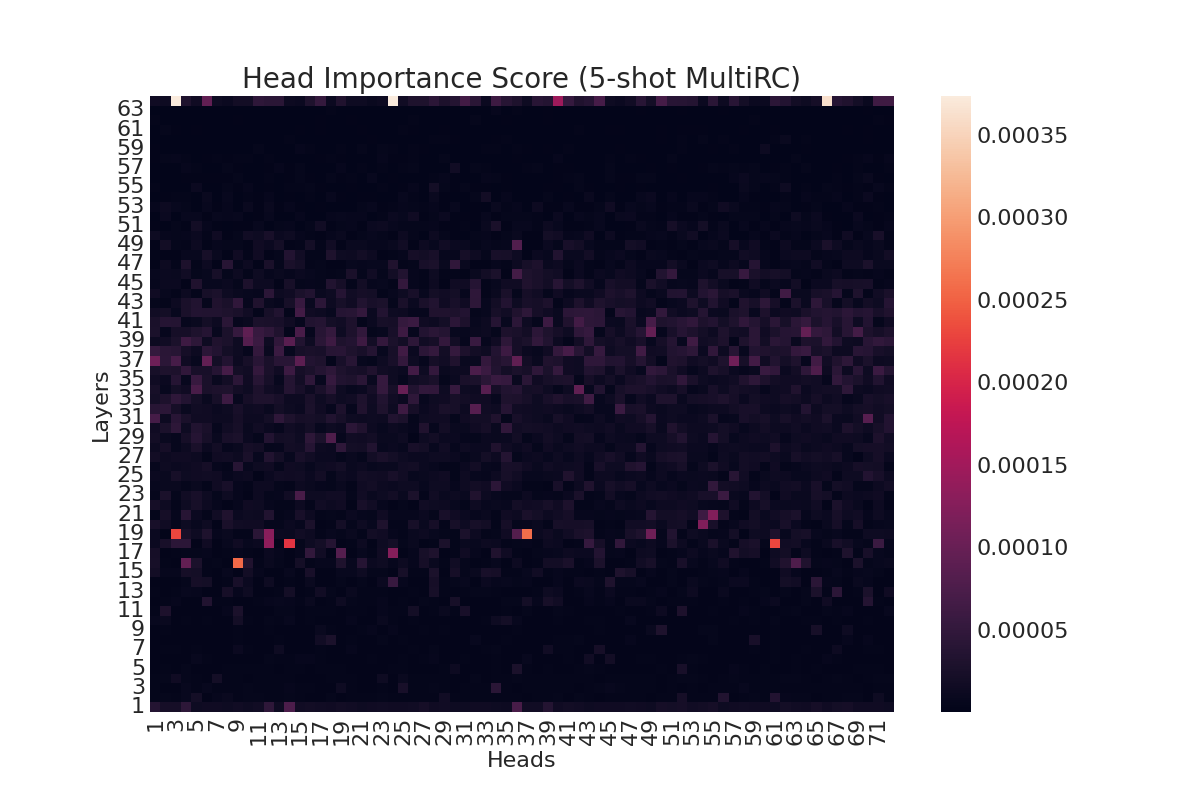}}}
     \subfloat[\centering OpenBookQA]{{\includegraphics[scale=0.2]{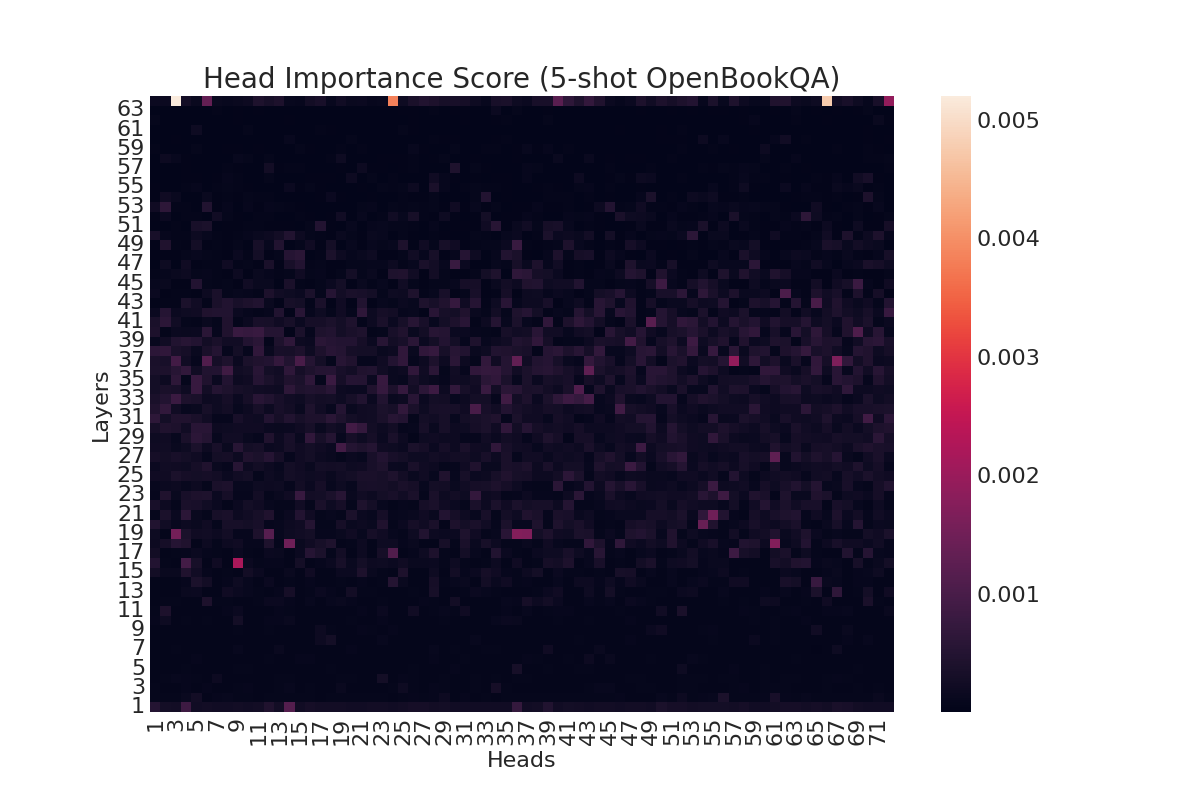}}}
     \qquad
     
     \subfloat[\centering COPA]{{\includegraphics[scale=0.2]{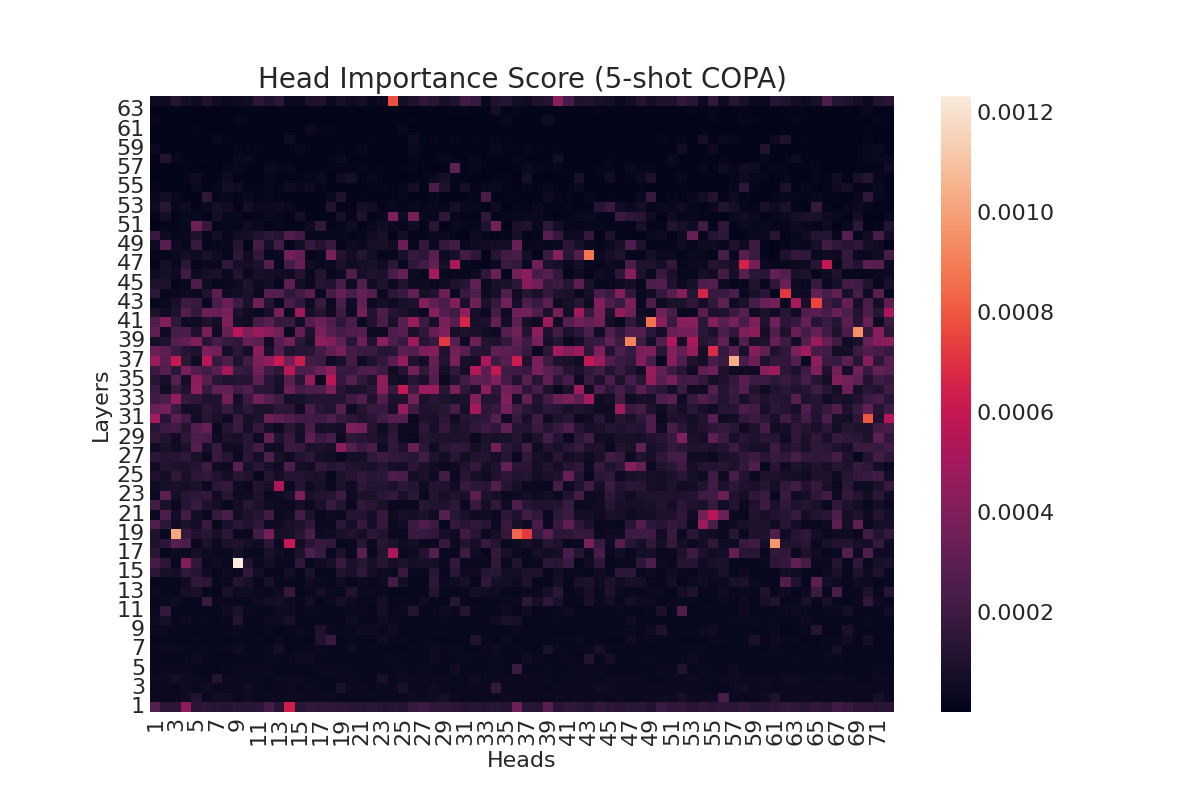}}}
     \subfloat[\centering PIQA]{{\includegraphics[scale=0.2]{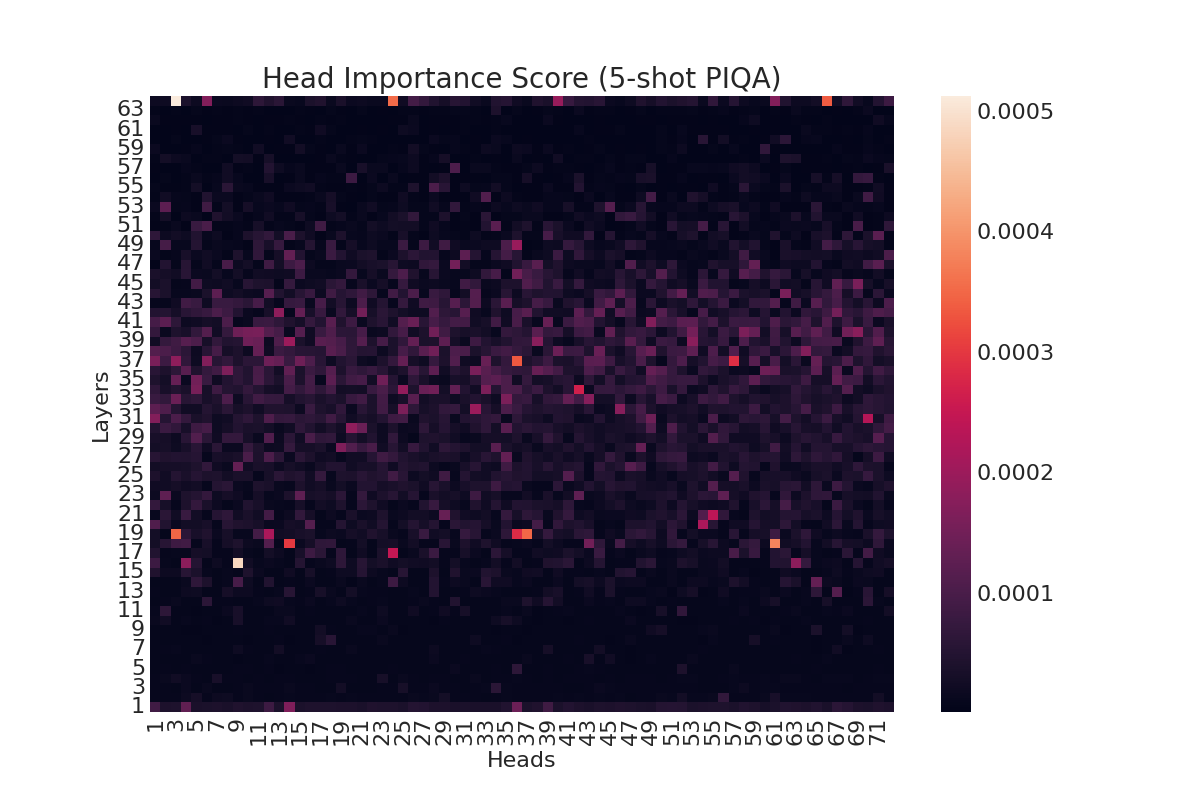}}}
     \subfloat[\centering ReCoRD]{{\includegraphics[scale=0.2]{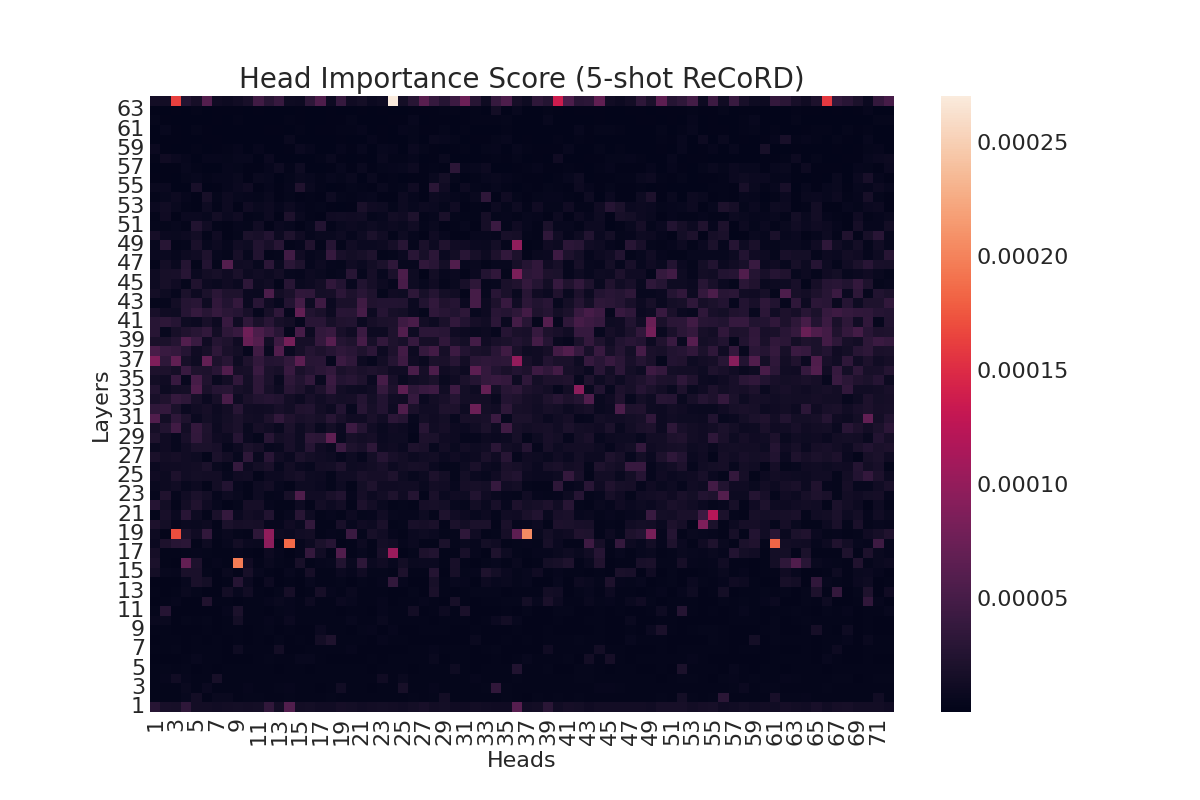}}}
     \qquad
     
     \subfloat[\centering WIC]{{\includegraphics[scale=0.2]{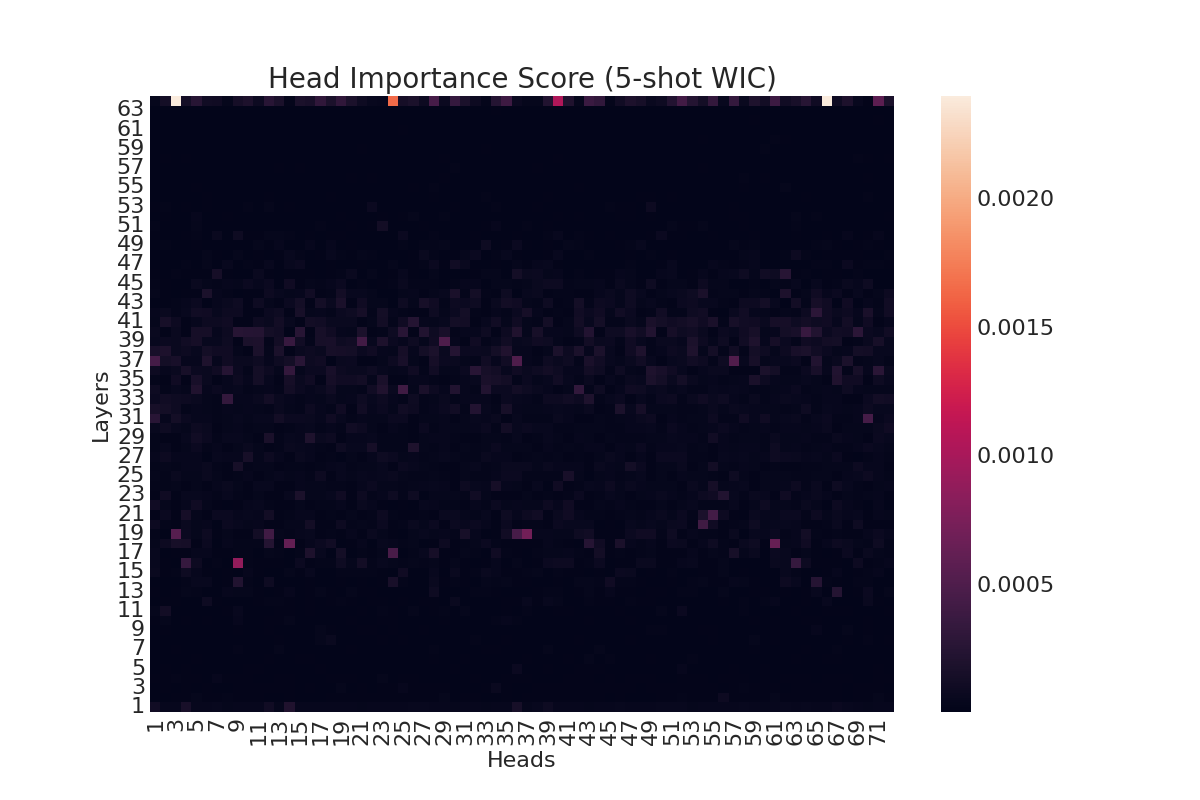}}}
     \subfloat[\centering WSC]{{\includegraphics[scale=0.2]{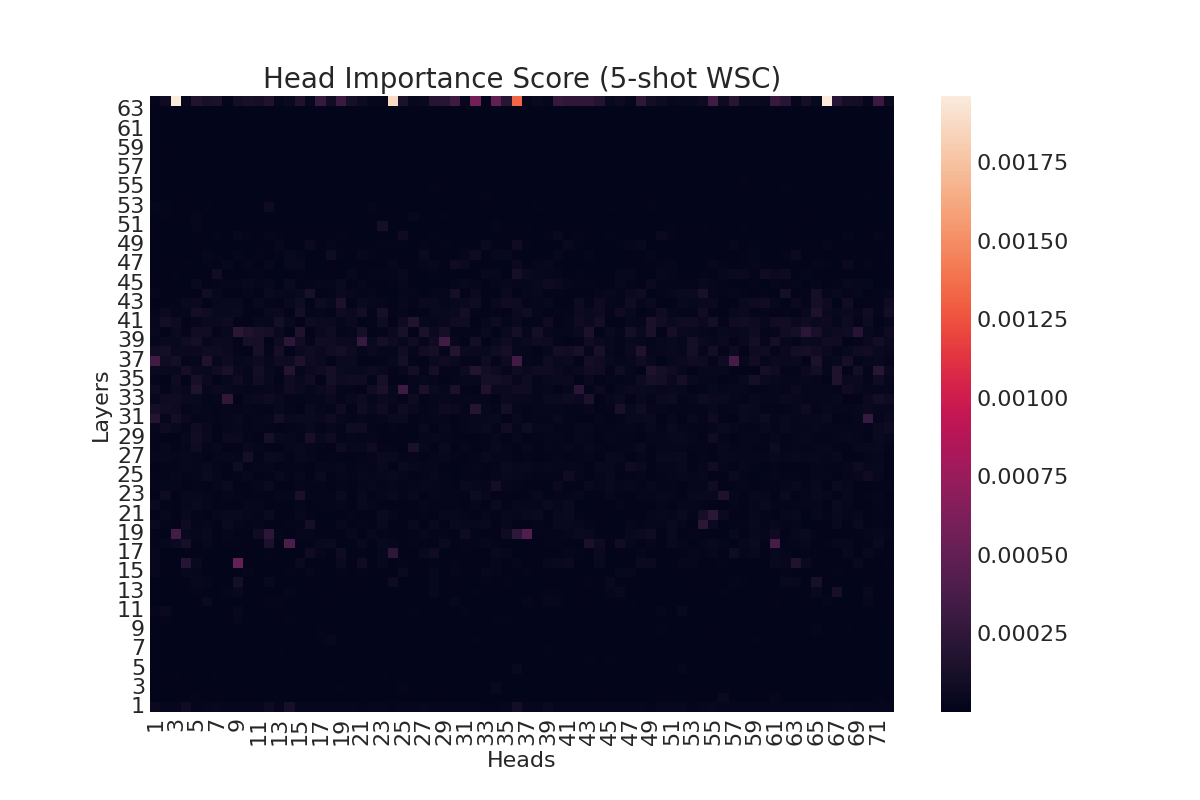}}}
     
    \caption{Attention head importance score heatmaps for five-shot in-context learning with OPT-66B for each task.}
    \label{app_fig:heatmap-5shot}
\end{figure*}

Figures \ref{app_fig:heatmap-0shot}, \ref{app_fig:heatmap-1shot} and \ref{app_fig:heatmap-5shot} depict the attention head importance scores for each task in the zero-shot, one-shot and five-shot settings respectively.

\begin{figure*}[h]
    \centering
    \includegraphics[width=\linewidth]{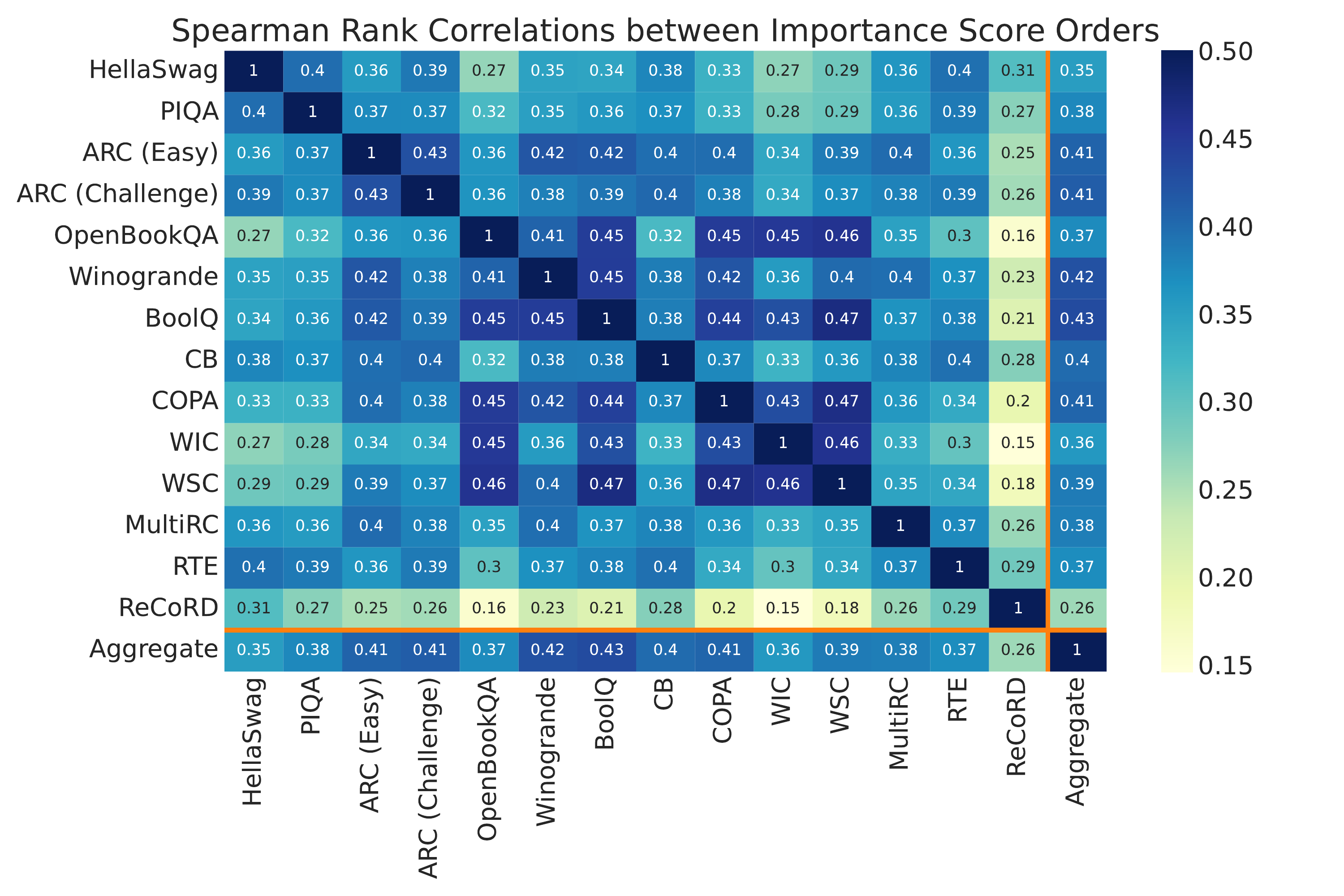}
    \caption{Spearman's rank correlation coefficients between the attention head importance rankings for various tasks in the five-shot setting. All $p$-values $< 0.01$.}
    \label{appen_fig:spearman}
\end{figure*}

\subsection{Cross-Task Analysis: Spearman's Rank Correlation}
\label{appendix:spearman rank}

Figure \ref{appen_fig:spearman} depicts the Spearman's rank correlation coefficients (SRCC) between the attention head importance rankings for every pair of tasks in the five-shot setting. It also depicts the SRCC between the aggregate ranking and the ranking for each constituent task.

\subsection{Cross-Task Analysis: Generalization Trends}
\label{appen:cross_task}

\begin{figure*}
    \centering
    \subfloat[\centering\label{appen_fig:cross_task_a_1shot} COPA]{{\includegraphics[width=0.36\linewidth]{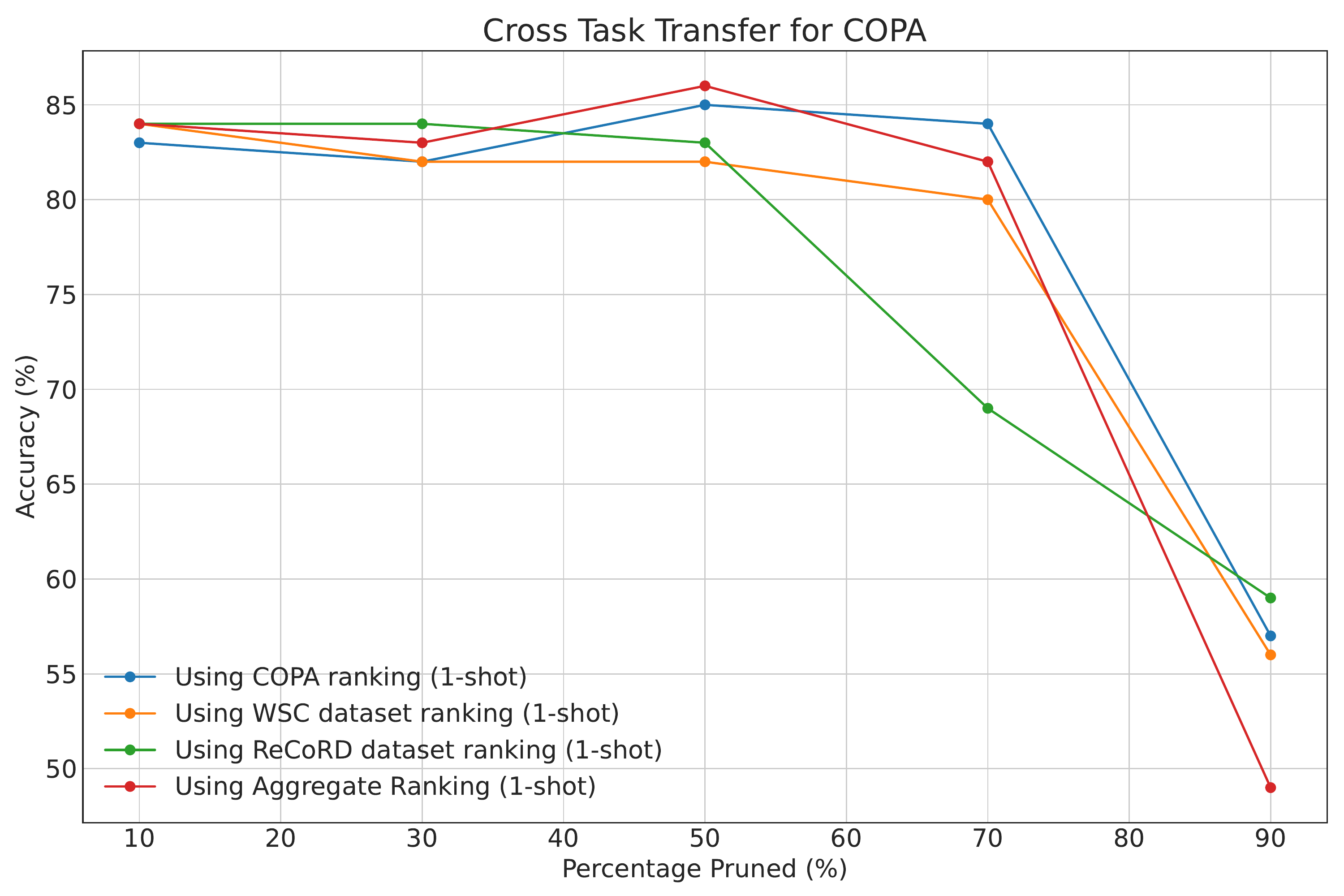}}}
     \subfloat[\centering\label{appen_fig:cross_task_b_1shot} Winogrande]{{\includegraphics[width=0.36\linewidth]{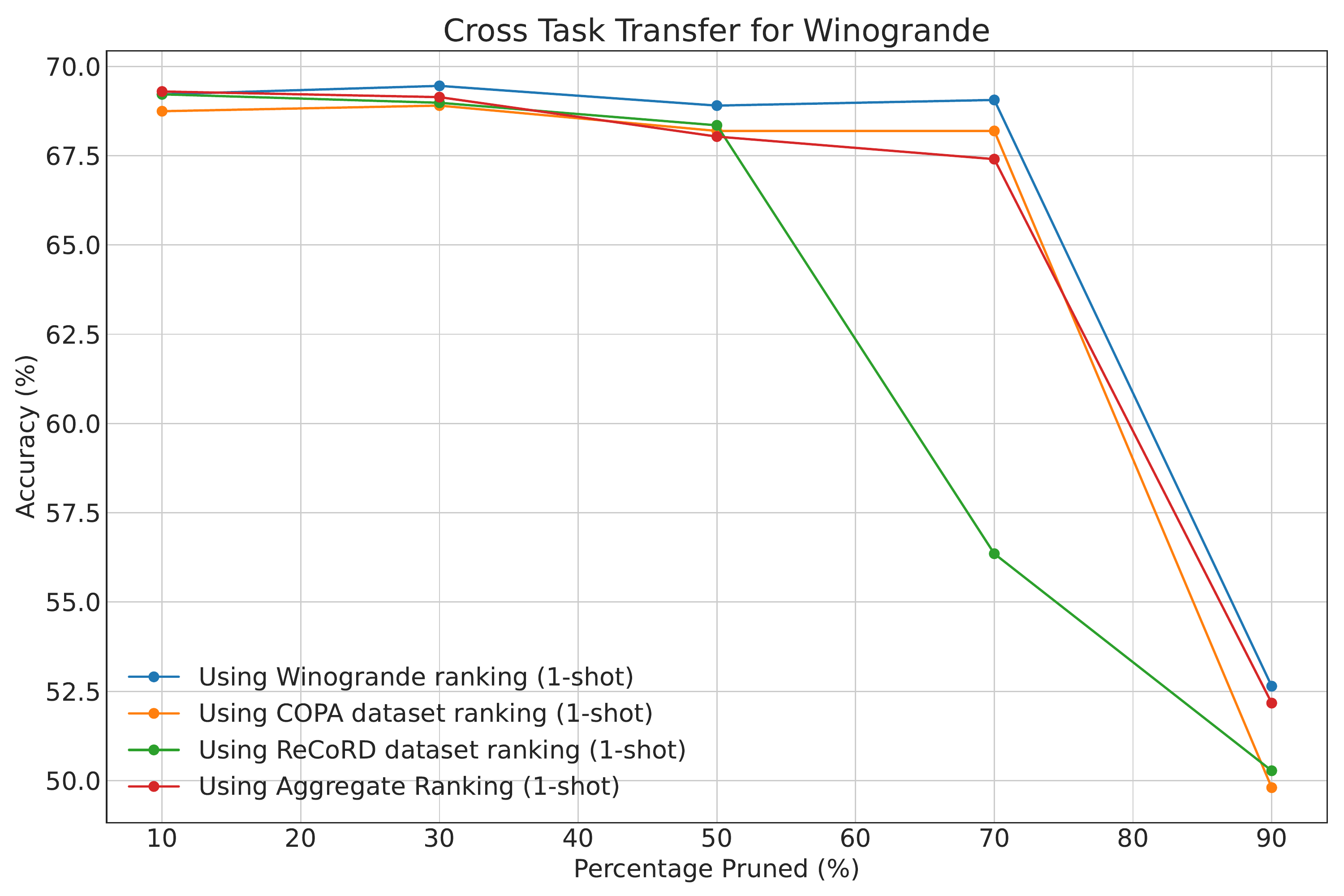}}}
     \subfloat[\centering\label{appen_fig:cross_task_c_1shot} ReCoRD]{{\includegraphics[width=0.36\linewidth]{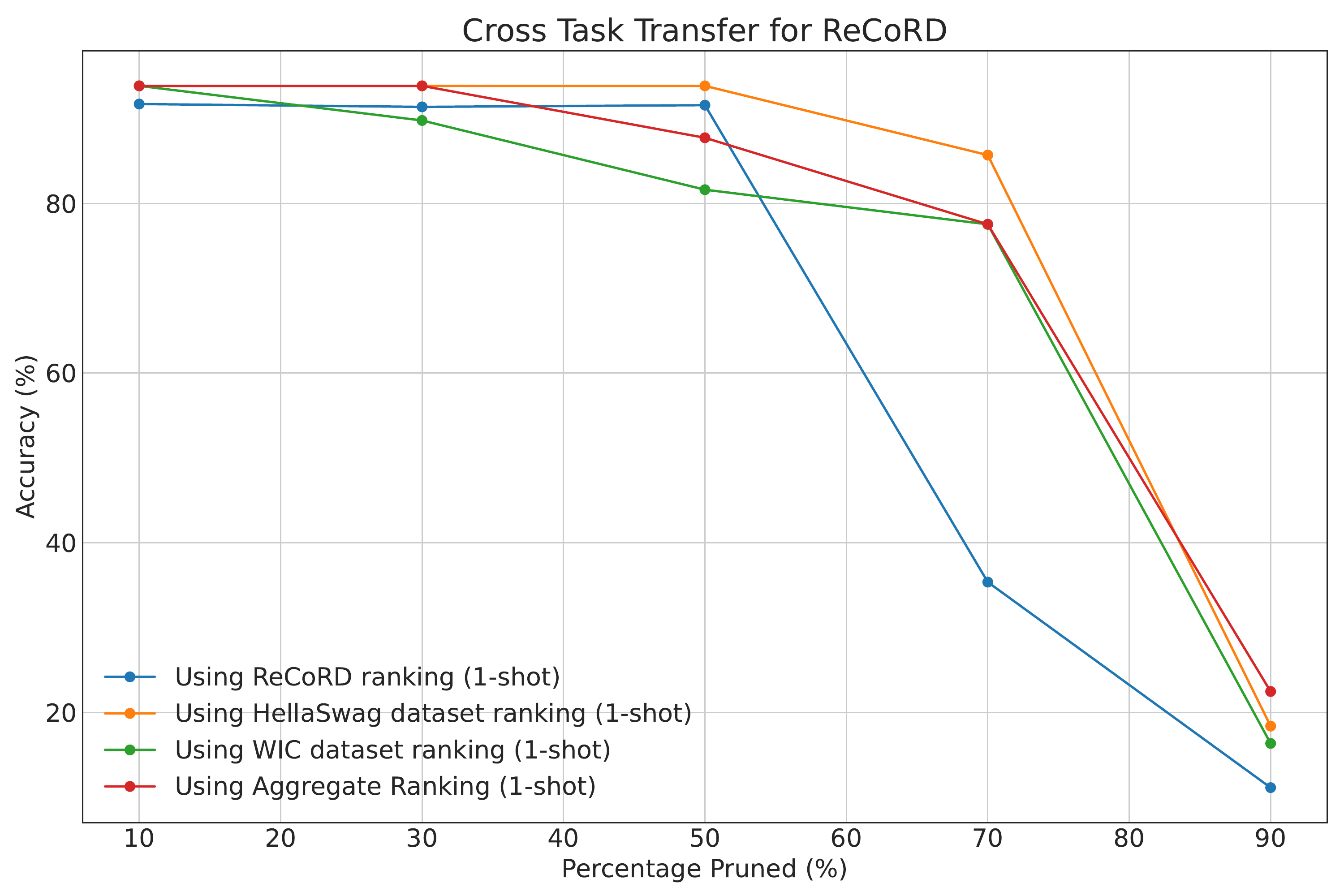}}}
     \qquad
      \subfloat[\centering\label{appen_fig:cross_task_d_1shot} MathQA]{{\includegraphics[width=0.4\linewidth]{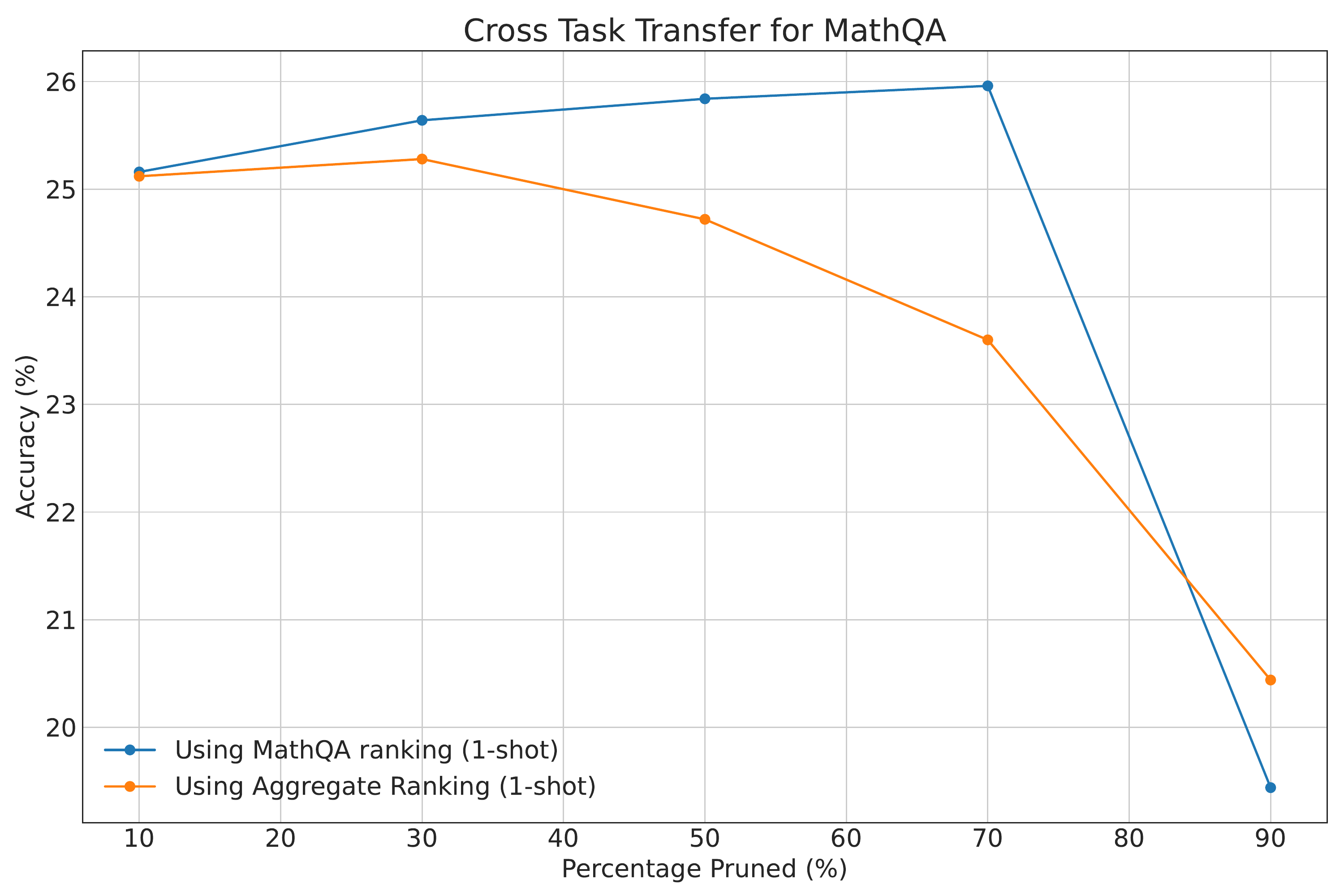}}}
    \subfloat[\centering\label{appen_fig:cross_task_e_1shot} LAMBADA]{{\includegraphics[width=0.4\linewidth]{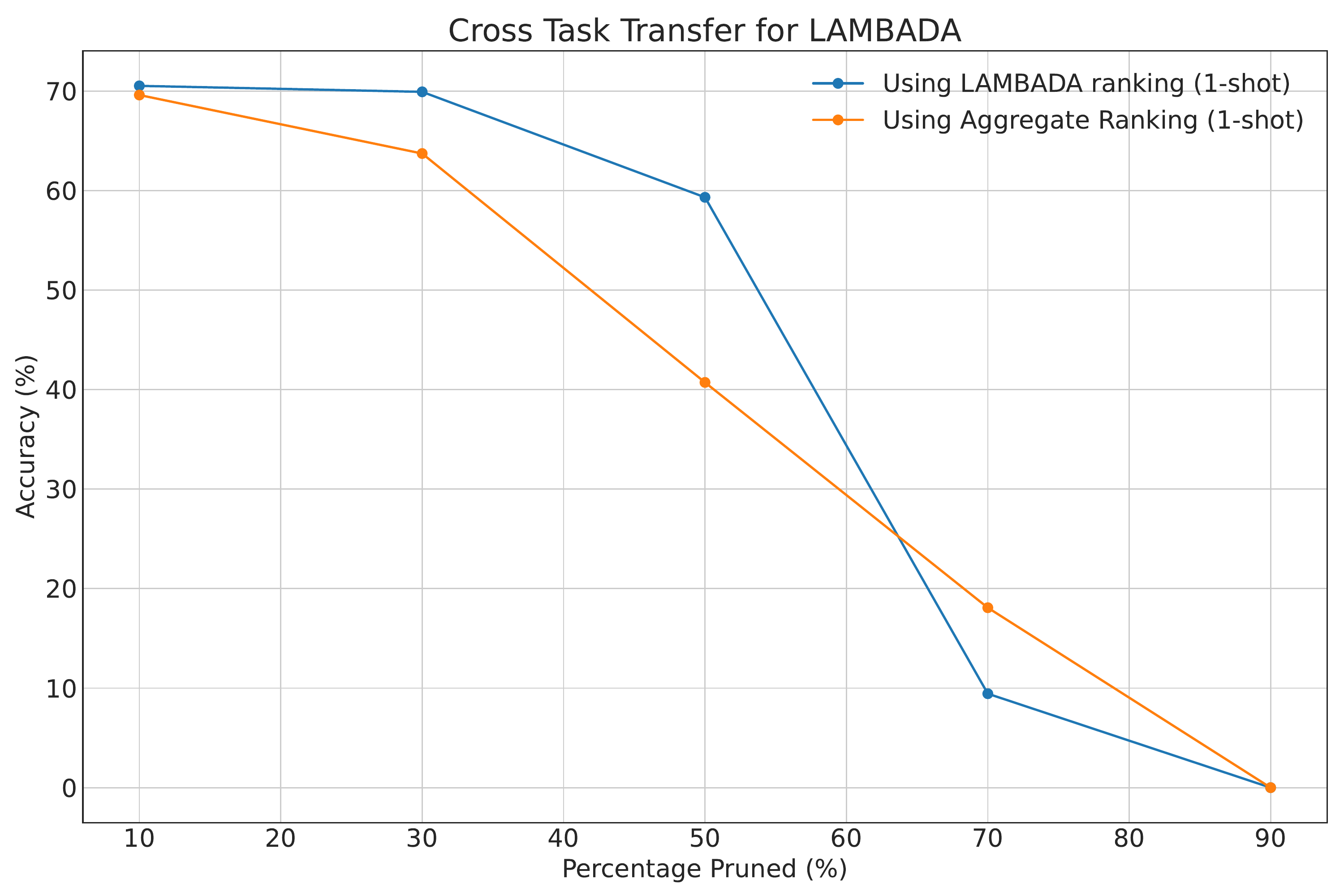}}}
    \caption{Cross-task transfer of attention head importance rankings as measured by impact of pruning on accuracy in the one-shot setting.}
    \label{appen_fig:cross_task_1_shot}
\end{figure*}

\begin{figure*}[h]
    \centering
    \subfloat[\centering\label{appen_fig:cross_task_a_5shot} COPA]{{\includegraphics[width=0.36\linewidth]{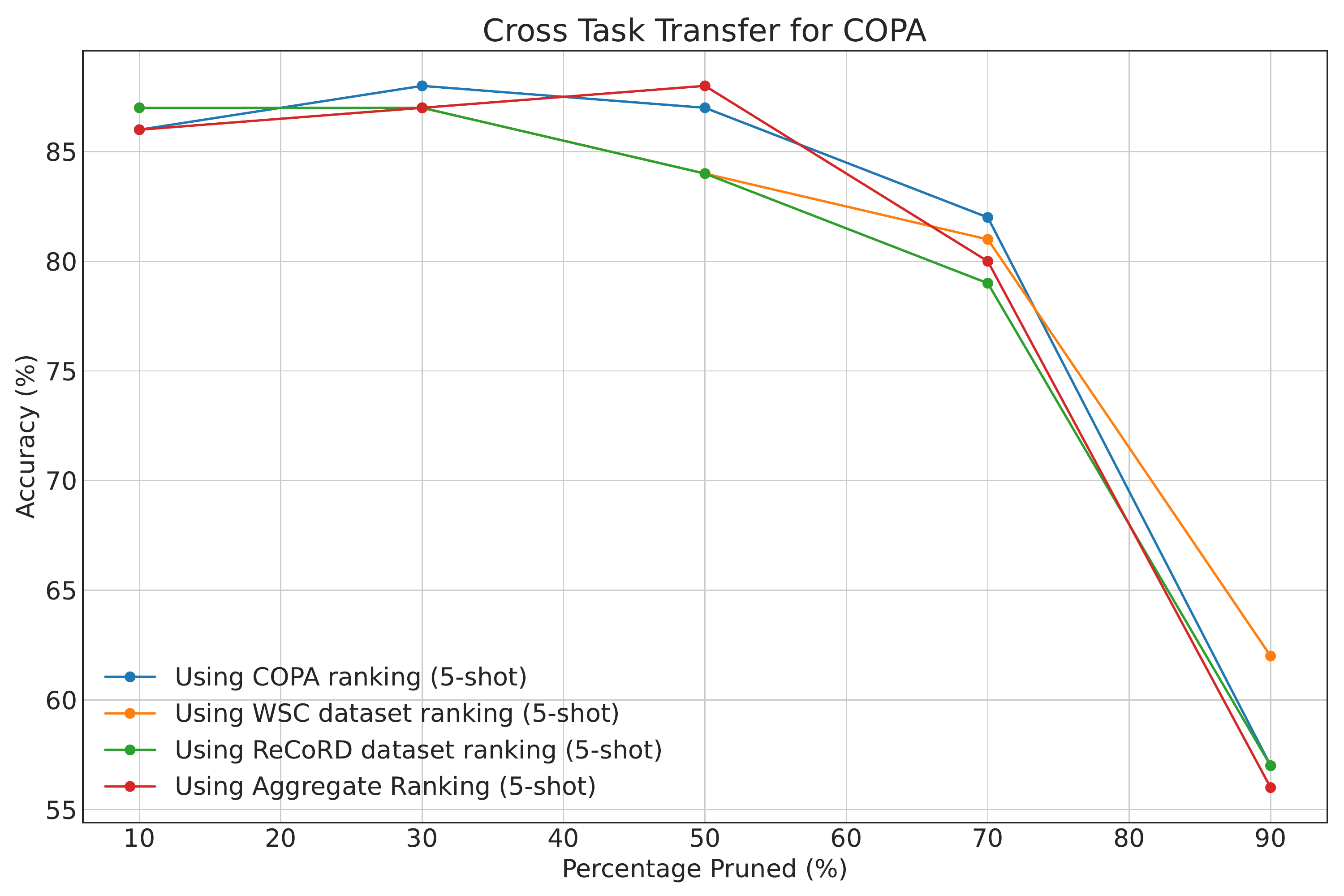}}}
     \subfloat[\centering\label{appen_fig:cross_task_b_5shot} Winogrande]{{\includegraphics[width=0.36\linewidth]{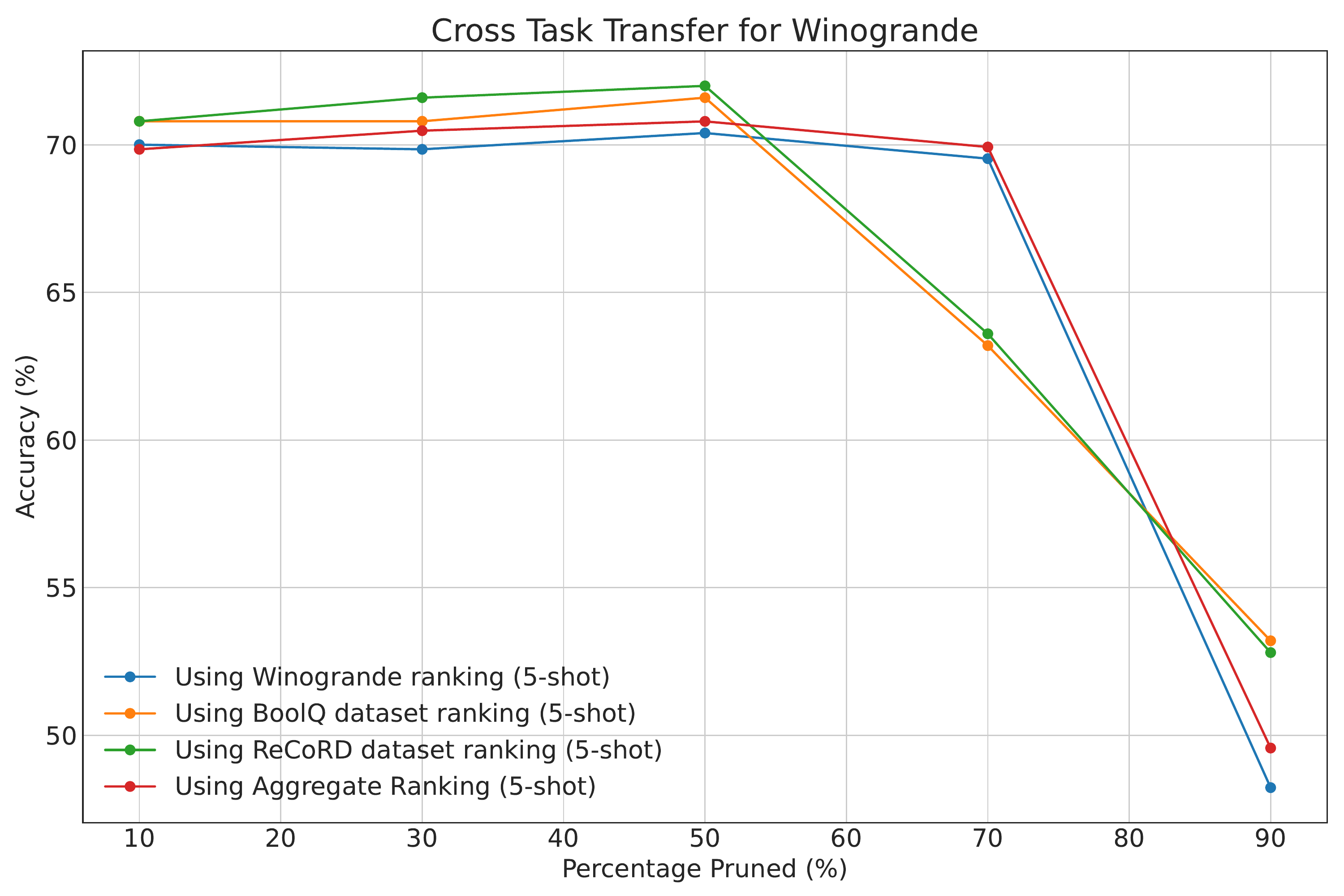}}}
     \subfloat[\centering\label{appen_fig:cross_task_c_5shot} ReCoRD]{{\includegraphics[width=0.36\linewidth]{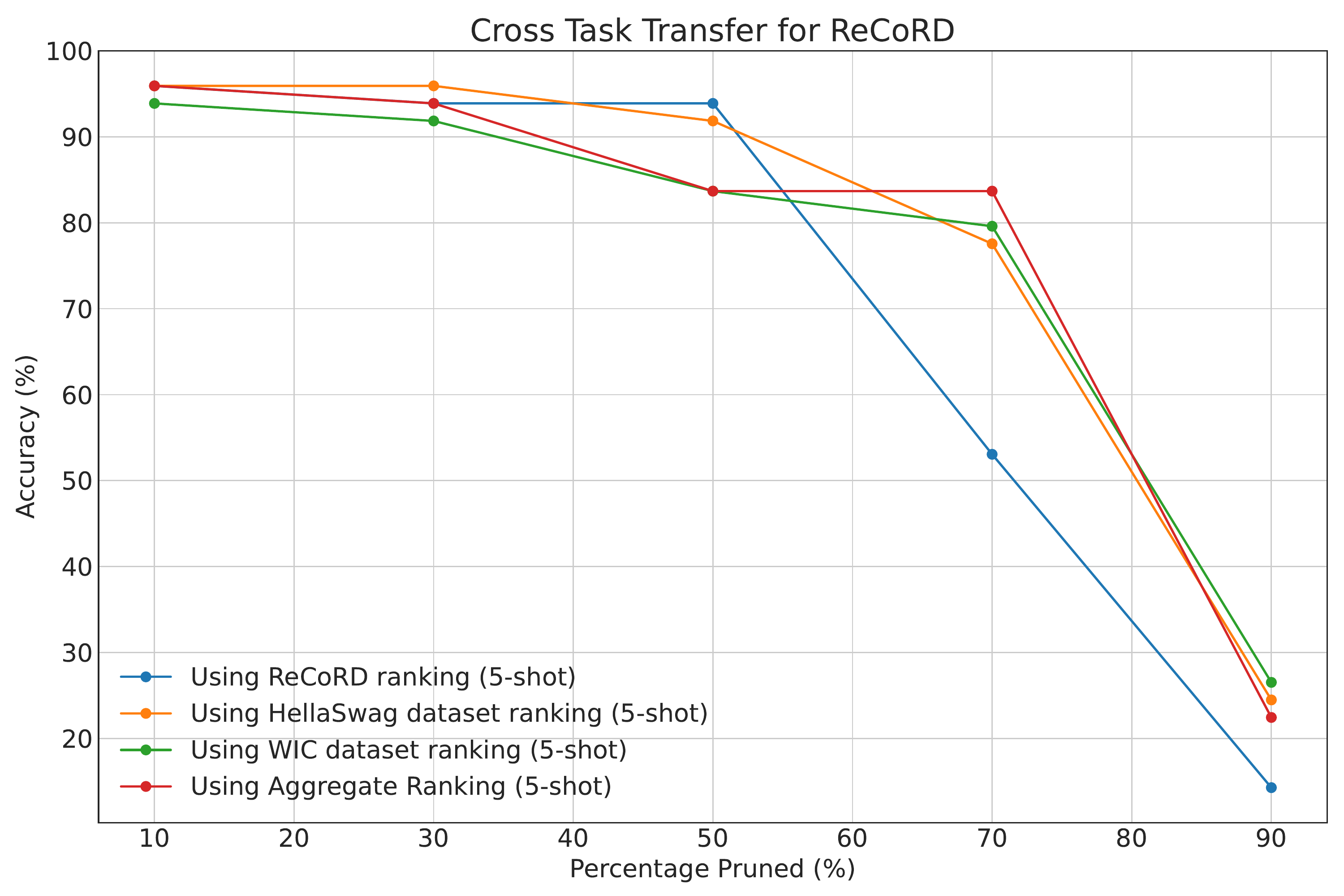}}}
     \qquad
      \subfloat[\centering\label{appen_fig:cross_task_d_5shot} MathQA]{{\includegraphics[width=0.4\linewidth]{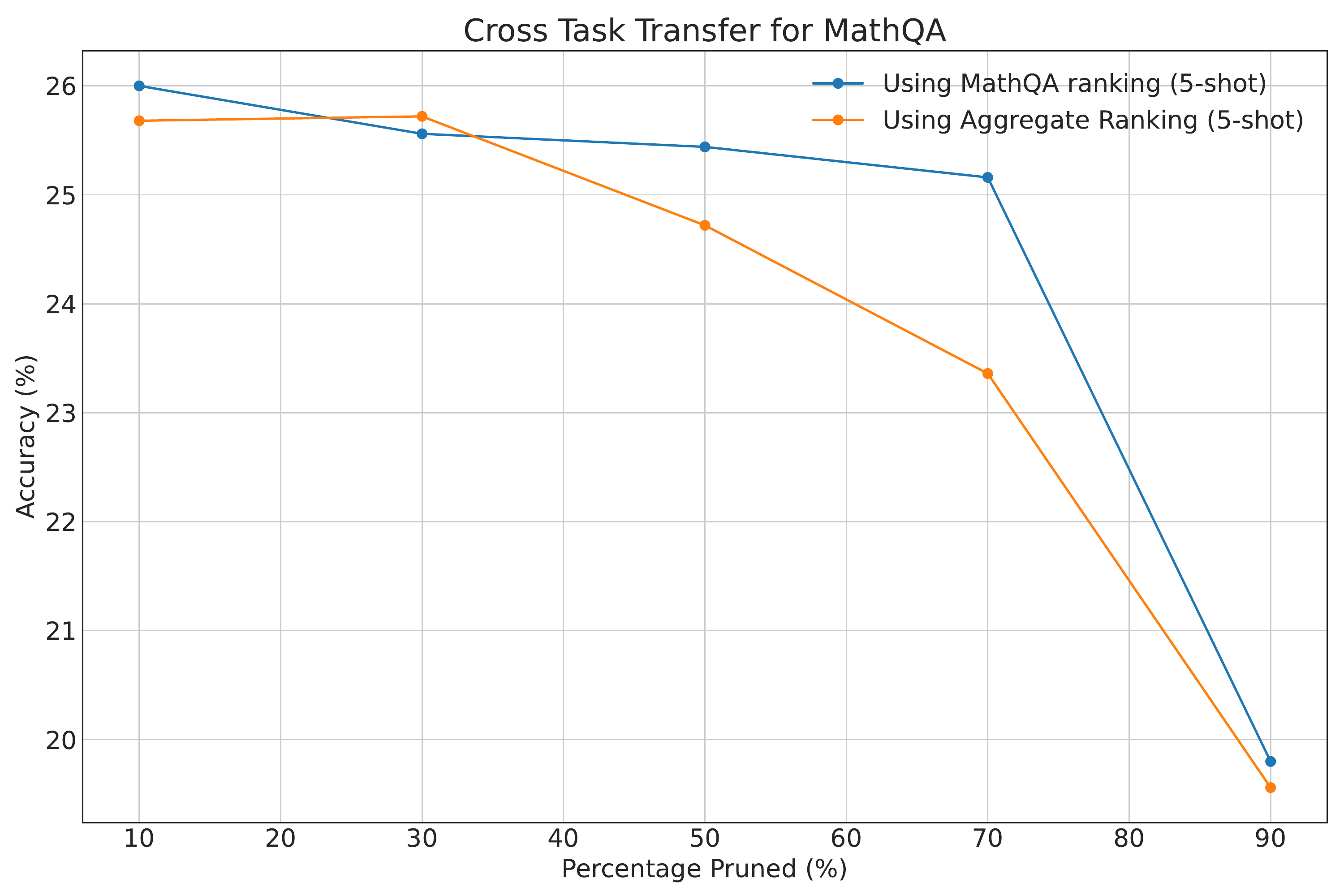}}}
    \subfloat[\centering\label{appen_fig:cross_task_e_5shot} LAMBADA]{{\includegraphics[width=0.4\linewidth]{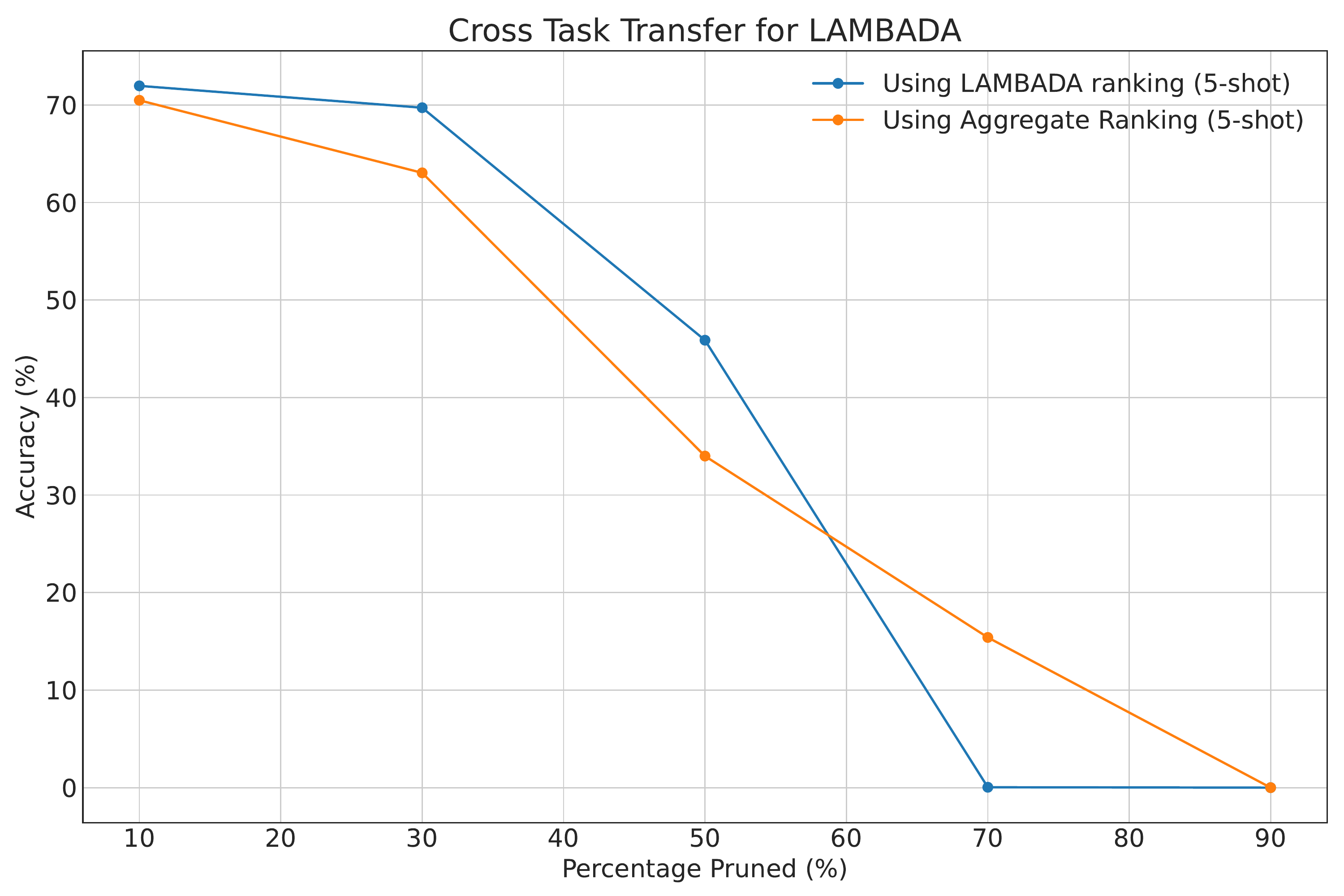}}}
    \caption{Cross-task transfer of attention head importance rankings as measured by impact of pruning on accuracy in the five-shot setting.}
    \label{appen_fig:cross_task_5_shot}
\end{figure*}

Figures \ref{appen_fig:cross_task_1_shot} and \ref{appen_fig:cross_task_5_shot} depict the cross-task head importance ranking generalization plots in the one-shot and five-shot settings.

\subsection{Details of Prefix Matching and Copying Scores}
\label{appen:pm_cs}

Algorithms \ref{alg:prefix_matching} and \ref{alg:copying_score} contain pseudo-code to compute prefix matching and copying scores respectively for each attention head in OPT-66B. We follow the approach described by \citet{olsson2022context}, but instead of computing scores using 10 sequences with fixed length of 25, we compute these scores using 100 sequences with varying lengths to account for OPT-66B's large maximum sequence length. As in \citet{olsson2022context}, we exclude a small fraction of the most and least common tokens from the model's vocabulary and randomly sample tokens for these sequences to strip out the effects of pre-training corpora memorization from our scores and inductive behavior analyses.

For prefix matching, the high-level approach is the following: take a random sequence, repeat it 4 times, perform a forward pass and then for each head, compute the attention pattern and take the average of all attention pattern entries attending from a given token back to tokens that succeeded the same token in earlier repeats.

For copying, the high-level approach is the following: take a random sequence, directly feed the sequence through each head and compute the contribution of the head to the output logits, and then measure how much the head increased the logit of the maximally attended to token over increasing the logits of other attendable tokens at each time-step. Unlike \citet{olsson2022context}, we do not scale the raw scores to be in the range of -1 to 1.

\subsection{Importance of Induction Heads to Each Task}
\label{appen:pm_cs_task}
\begin{figure*}[h]
\centering
    \subfloat[\centering\label{app_fig:prefix-matching-0-shot} Zero-shot]{{\includegraphics[width=0.36\linewidth]{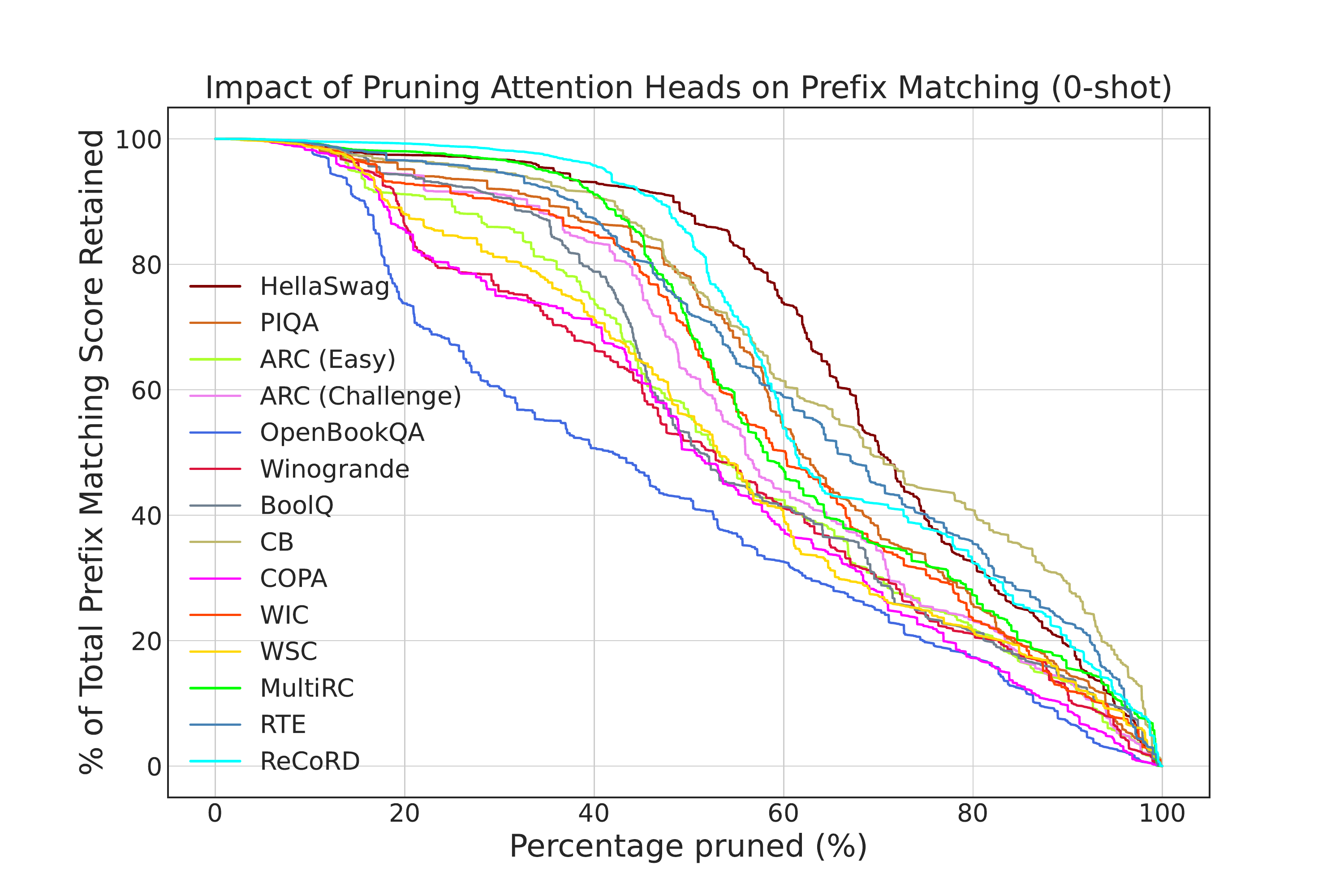}}}
    \subfloat[\centering\label{app_fig:prefix-matching-1-shot} One-shot]{{\includegraphics[width=0.36\linewidth]{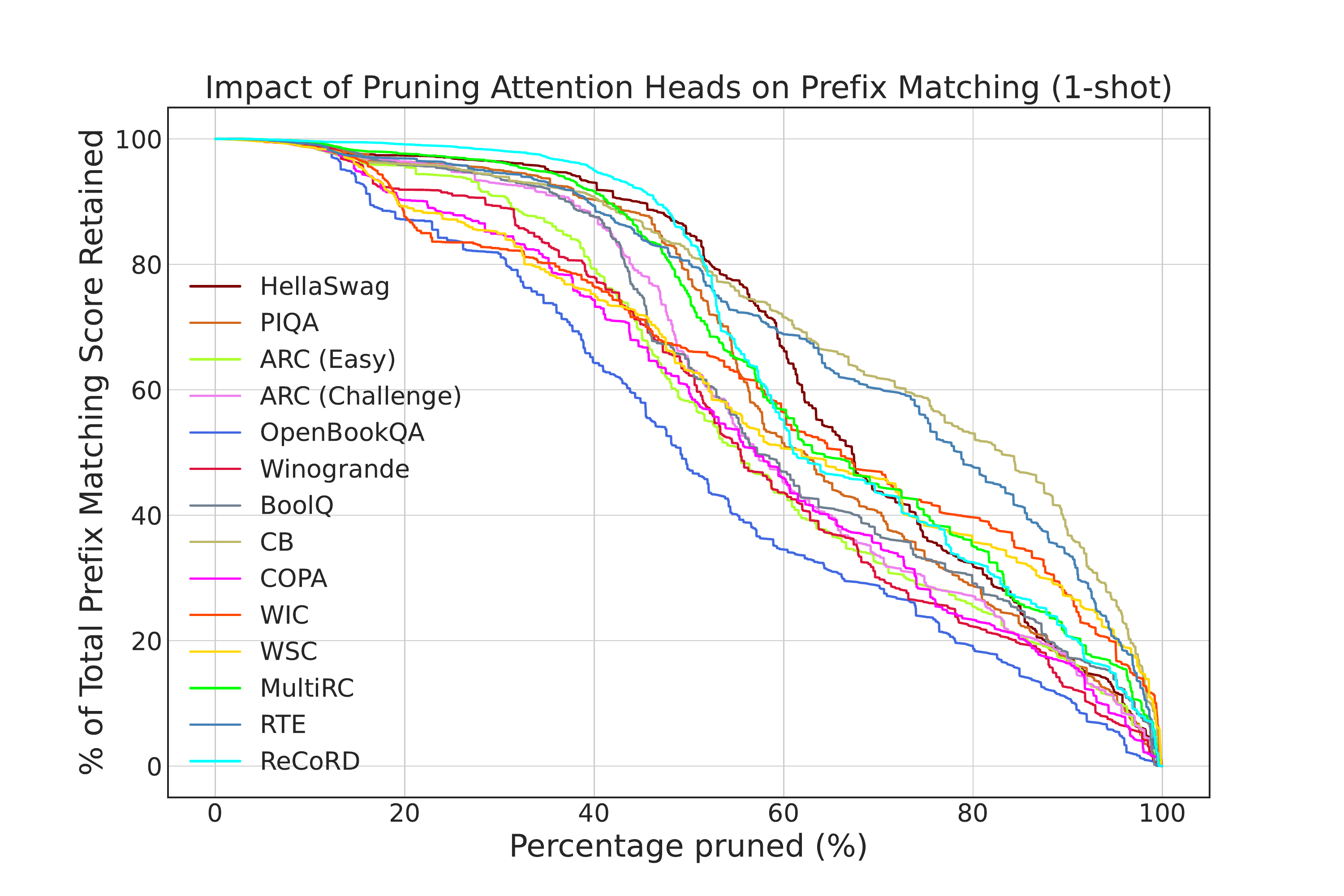}}}
    \subfloat[\centering\label{app_fig:prefix-matching-5-shot} Five-shot]{{\includegraphics[width=0.36\linewidth]{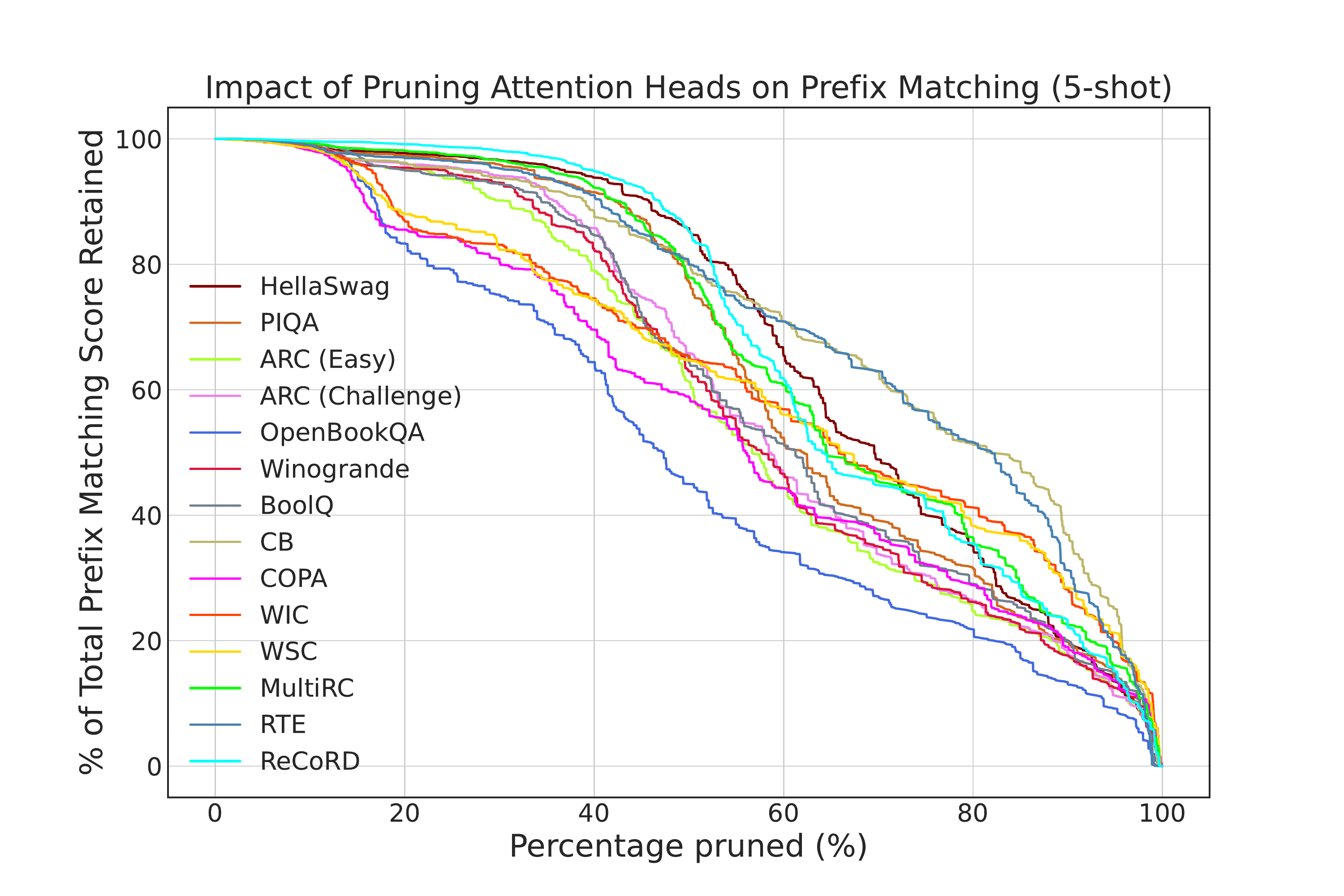}}}
\caption{Total prefix matching capacity retained as a function of percentage of attention heads pruned, where heads are pruned based on task-specific importance score rankings in the order of least important first.}
    \label{app_fig:prefix_matching_pruning}
\end{figure*}

\begin{figure*}[h]
\centering
    \subfloat[\centering\label{app_fig:copying-0-shot} Zero-shot]{{\includegraphics[width=0.36\linewidth]{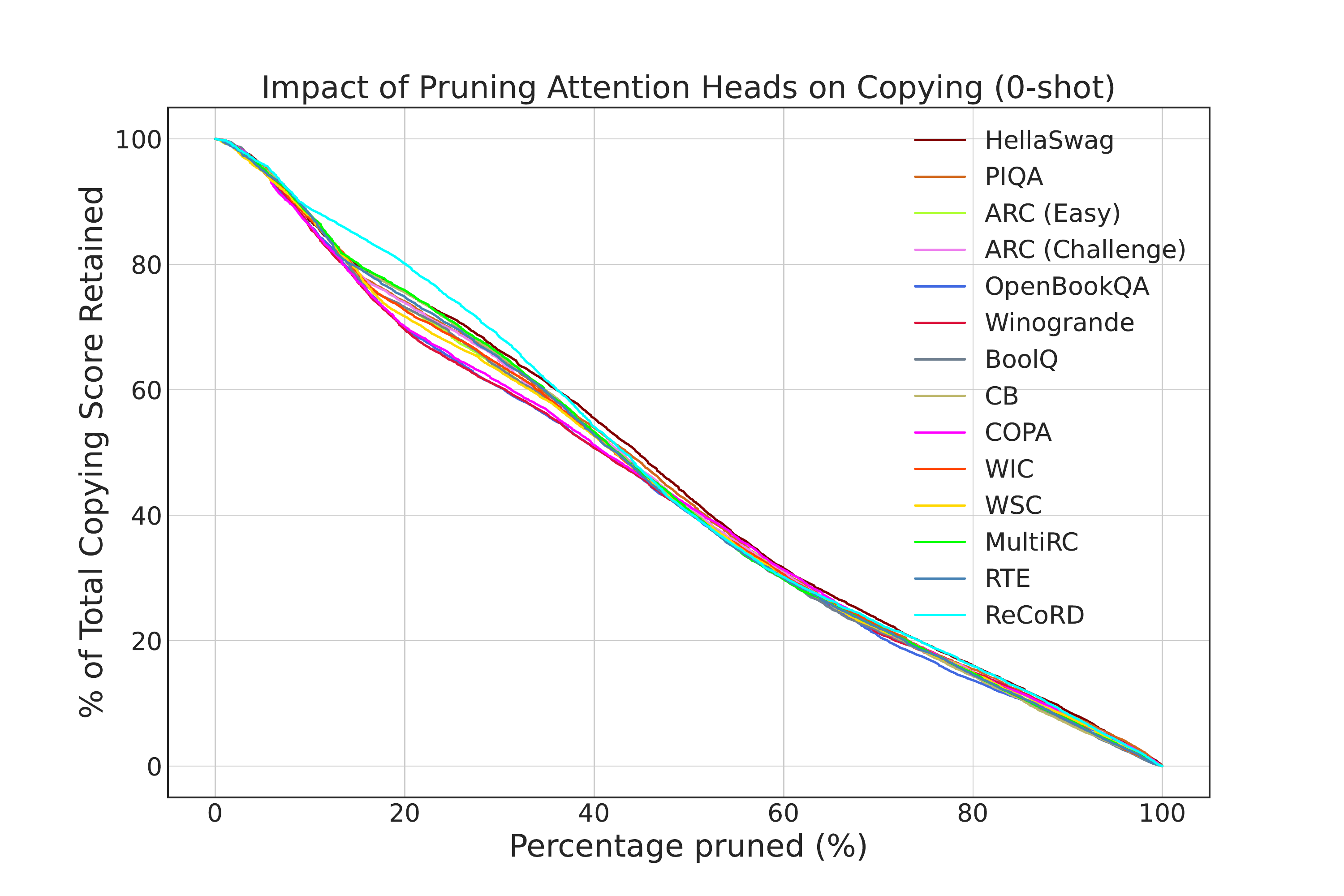}}}
    \subfloat[\centering\label{app_fig:copying-1-shot} One-shot]{{\includegraphics[width=0.36\linewidth]{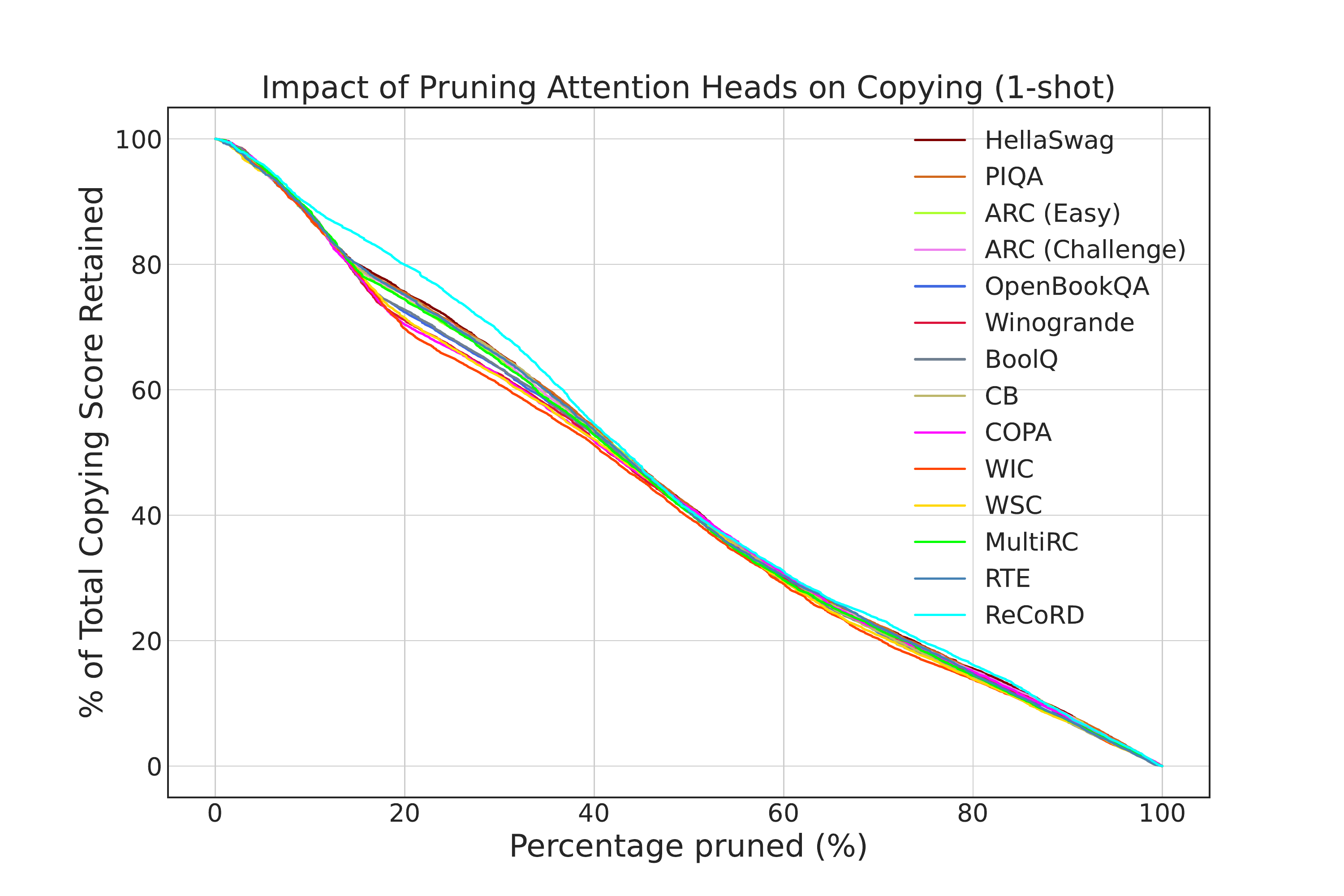}}}
    \subfloat[\centering\label{app_fig:copying-5-shot} Five-shot]{{\includegraphics[width=0.36\linewidth]{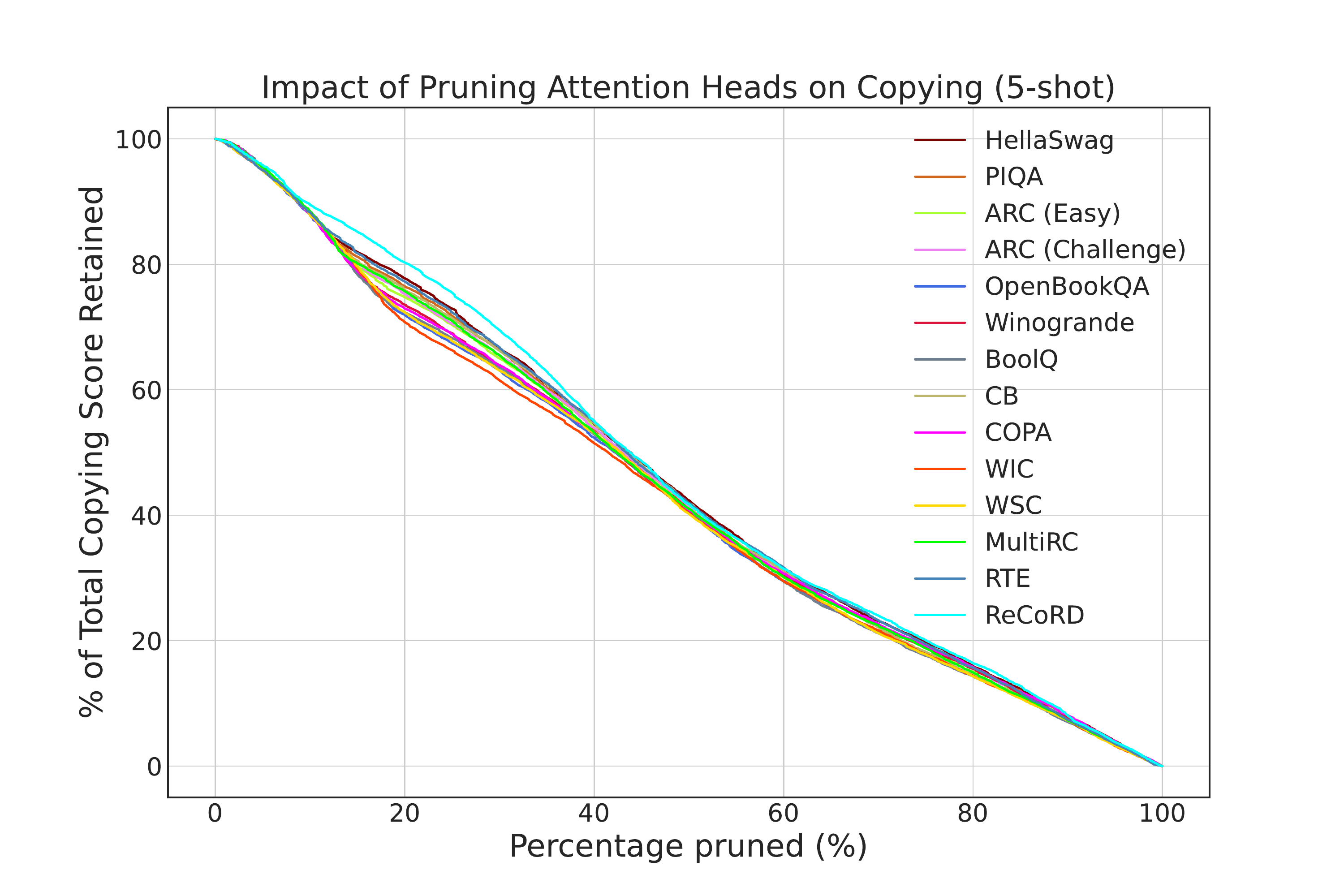}}}
\caption{Total copying capacity retained as a function of percentage of attention heads pruned, where heads are pruned based on task-specific importance score rankings in the order of least important first.}
    \label{app_fig:copying_pruning}
\end{figure*}

Figures \ref{app_fig:prefix_matching_pruning} and \ref{app_fig:copying_pruning} showcase the importance of induction heads to each task via measuring the percentage of the total prefix matching and copying capacities retained as a function of percentage of attention heads pruned, where heads are pruned based on each task's head importance ranking for each in-context learning setting (zero-shot, one-shot and five-shot) in the order of least important first. A small initial slope of decline implies that unimportant heads also have low prefix matching or copying scores while a steep initial slope of decline implies unimportant heads also have high prefix matching or copying scores. We observe differences in the slopes of decline across different tasks, with tasks like HellaSwag and ReCoRD (which have high accuracies in Figure \ref{exp_fig:ip_head}) having smaller initial slopes than a task like OpenBookQA (which has relatively lower accuracy in Figure \ref{exp_fig:ip_head}). When seen in conjunction, these plots not only point to the generality of induction heads to more sophisticated behaviors associated with in-context learning but also indicate that some tasks rely on induction heads more than others.

\begin{algorithm*}
\caption{Prefix Matching Scores for Attention Heads }\label{alg:prefix_matching}
\begin{algorithmic}
\State \textbf{Arguments}: $\text{Model }\mathcal{M}$
\\
\State $model \gets \text{Pretrained}(\mathcal{M})$
\State $layers, heads \gets model.\text{num\_layers}, model.\text{num\_heads\_per\_layer}$ 
\State $ranked\_vocab\_list \gets model.\text{tokenizer}.\text{vocab}$ \Comment{term-frequency based vocabulary list of model}
\State $exclude\_vocab\_size \gets 0.04 \times \text{len}(ranked\_vocab\_list)$ \Comment{remove 4\% most \& least common tokens}
\State $ranked\_vocab\_list \gets ranked\_vocab\_list[exclude\_vocab\_size:-exclude\_vocab\_size]$
\State $prefix\_matching \gets [\hspace{0.1cm}]$
\For{$seed \text{ in } \{1\cdots 100\}$}
    \State $L \gets 2 \times seed + 23$ \Comment{ensure $4L \in [100, 892]$}
    \State$X \gets random.choice(ranked\_vocab\_list, \text{size}=L, \text{seed} = seed, \text{replace} = False)$ \Comment{$L$ length random sequence with all unique tokens}
    \State$X \gets \text{repeat}(X, 4)$ \Comment{Repeat it four times}
    \State $Y \gets model.\text{forward}(X)$ \Comment{Forward pass the repeated sequence}
    \State $score \gets$ zeros($layers$, $heads$) \Comment{Zero matrix of shape $layers \times heads$}
        \For{$layer$ in $layers$}
            \For{$head$ in $heads$}
                \State $att \gets model.\text{get\_attention}(layer, head)$ \Comment{Shape: $4L \times 4L$}
                \For{$token$ in $\{L+1 \cdots 4L\}$} \Comment{Repetition starts from token $L+1$}
                    \State $att\_token \gets att[token]$ \Comment{Shape: $4L$}
                    \For{every $prev\_token$ == $token$} \Comment{Look at the previous repetitions of the token}
                        \State $prefix\_score = att\_token[prev\_token + 1]$\Comment{Attention given to token whose prefix is current token}
                        \State $score[layer][head] \gets score[layer][head] + prefix\_score$ 
                    \EndFor
                \EndFor
                \State $score[layer][head] \gets score[layer][head]/3L$ \Comment{Normalizing by length of for loop}
            \EndFor
        \EndFor
    \State $prefix\_matching.\text{append}(score)$ \Comment{Prefix matching scores via one randomly generated example}
\EndFor
\State $prefix\_matching \gets \text{average}(prefix\_matching)$ \Comment{Attention head-wise average over all examples}\\
\Return{$prefix\_matching$}
\end{algorithmic}
\end{algorithm*}

\begin{algorithm*}
\caption{Copying Scores for Attention Heads}\label{alg:copying_score}
\begin{algorithmic}
\State \textbf{Arguments}: $\text{Model }\mathcal{M}$
\State \textbf{Definitions}: $\text{Dimension per Head }D, \text{Vocabulary Size }V$ 
\\
\State $model \gets \text{Pretrained}(\mathcal{M})$
\State $layers, heads \gets model.\text{num\_layers}, model.\text{num\_heads\_per\_layer}$
\State $ranked\_vocab\_list \gets model.\text{tokenizer}.\text{vocab}$ \Comment{term-frequency based vocabulary list of model}
\State $exclude\_vocab\_size \gets 0.04 \times \text{len}(ranked\_vocab\_list)$ \Comment{remove 4\% most \& least common tokens}
\State $ranked\_vocab\_list \gets ranked\_vocab\_list[exclude\_vocab\_size:-exclude\_vocab\_size]$
\State $copying\_score \gets [\hspace{0.1cm}]$
\For{$seed \text{ in } \{1\cdots 100\}$}
    \State $L \gets 4 \times (2 \times seed + 23)$ \Comment{$L \in [100, 892]$}
    \State$X \gets random.choice(ranked\_vocab\_list, \text{size}=L, \text{seed} = seed, \text{replace}=False)$ \Comment{$L$ length random sequence with all unique tokens}
    \State $score \gets$ zeros($layers$, $heads$) \Comment{Zero matrix of shape $layers \times heads$}
        \For{$layer$ in $layers$}
            \For{$head$ in $heads$}
                \State $attn\_layer\_head \gets model.\text{get\_attention\_head}(layer,head)$ 
                \State $out \gets attn\_layer\_head(X)$ \Comment{Shape: $L \times D$}
                \State $attention \gets model.\text{get\_attention}(layer, head)$ \Comment{Shape: $L \times L$}
                \State $logits \gets model.\text{hidden\_to\_vocab}(out)$ \Comment{Shape: $L \times V$}
                \State $logits \gets \text{softmax}(logits, dim = 1)$ 
                \For{$token$ in $\{1 \cdots L\}$} 
                    \State $max\_ind \gets \text{argmax}(attention[token])$ \Comment{Index of the token being max attended to}
                    \State $attendable\_input \gets X[1:token]$ \Comment{Attendable input tokens}
                    \State $attendable\_logits \gets logits[token][attendable\_input]$ \Comment{Logits of attendable tokens}
                    \State $mean\_of\_logits \gets \text{average}(attendable\_logits)$
                    \State $raised\_logits \gets attendable\_logits - mean\_of\_logits$
                    \State $relu\_raised\_logits \gets \text{ReLU}(raised\_logits)$ \Comment{Computing raise in logit values}
                    \State $relu\_raised\_logit\_max\_ind \gets relu\_raised\_logits[X[max\_ind]]$
                    \State $temp\_score \gets relu\_raised\_logit\_max\_ind / \text{sum}(relu\_raised\_logits)$ 
                    \State $score[layer][head]\gets score[layer][head]+ temp\_score$
                \EndFor
                \State $score[layer][head] \gets score[layer][head]/L$ \Comment{Normalizing by length of for loop}
            \EndFor
        \EndFor
    \State $copying\_score.\text{append}(score)$ \Comment{Copying scores via one randomly generated example}
\EndFor
\State $copying\_score \gets \text{average}(copying\_score)$ \Comment{Attention head-wise average over all examples}\\
\Return{$copying\_score$}
\end{algorithmic}
\end{algorithm*}

\end{document}